\documentclass{tmlr}
\usepackage{anyfontsize}
\usepackage{pifont}
\usepackage{dsfont}
\usepackage{adjustbox}

\newtcolorbox{AIbox}[2][]{aibox,title=#2,#1}
\usepackage{sidecap}

\usepackage{lipsum} 

\definecolor{midnightgreen}{rgb}{0.0, 0.29, 0.33}
\definecolor{deepgreen}{HTML}{055c29}
\definecolor{deeppurple}{HTML}{7030a0}
\definecolor{deepblue}{HTML}{171d91}
\definecolor{brown}{HTML}{843c0c}
\definecolor{shadered}{HTML}{ffe5e5}
\definecolor{shadegreen}{HTML}{e5f7ed}
\definecolor{msftBlack}{RGB}{0,0,0}
\definecolor{lightred}{RGB}{255,163,163}
\definecolor{deepred}{RGB}{153,0,0}

\definecolor{lightblue}{rgb}{0.22,0.45,0.70}%
\definecolor{queryrewritter}{HTML}{91B1DA}
\definecolor{resultsummarizer}{HTML}{5680A8}
\definecolor{shadecolor}{rgb}{0.92,0.92,0.92}
\definecolor{LightGray}{HTML}{F0F1F2} 
\definecolor{barblue}{RGB}{90,120,180}
\definecolor{barorange}{RGB}{225,124,5}

\begin{document}

\title{\textbf{AlphaApollo:} A System for Deep Agentic Reasoning}

\author{
{\small
Zhanke Zhou\textsuperscript{1$\dagger$},
Chentao Cao\textsuperscript{1*},
Xiao Feng\textsuperscript{1*},
Xuan Li\textsuperscript{1*},
Zongze Li\textsuperscript{1*},
Xiangyu Lu\textsuperscript{1*},
Jiangchao Yao\textsuperscript{3*},
\quad \quad
Weikai Huang\textsuperscript{1},
Tian Cheng\textsuperscript{1},
Jianghangfan Zhang\textsuperscript{1},
Tangyu Jiang\textsuperscript{1},
Linrui Xu\textsuperscript{1},
Yiming Zheng\textsuperscript{1},
\quad \quad \quad \quad
Brando Miranda\textsuperscript{4},
Tongliang Liu\textsuperscript{5,2},
Sanmi Koyejo\textsuperscript{4},
Masashi Sugiyama\textsuperscript{2,6},
Bo Han\textsuperscript{1,2}
}
\\
\textsuperscript{1}TMLR Group, Department of Computer Science, Hong Kong Baptist University;
\textsuperscript{2}RIKEN AIP;
\\
\textsuperscript{3}Cooperative Medianet Innovation Center, Shanghai Jiao Tong University;
\\
\textsuperscript{4}Stanford University; 
\textsuperscript{5}Sydney AI Centre, The University of Sydney;
\textsuperscript{6}The University of Tokyo
\\
\textsuperscript{$\dagger$}Team lead;
\textsuperscript{*}Equal contribution, listed in alphabetical order
}

\date{}

\definecolor{mygray}{gray}{.92}
\newcommand{\gray}{\cellcolor{mygray}}
\definecolor{baselinecolor}{rgb}{1, 1, 1}
\newcommand{\baseline}{\cellcolor{baselinecolor}}
\definecolor{ourmethodcolor}{rgb}{0.94, 0.97, 1.0}
\newcommand{\ours}{\cellcolor{ourmethodcolor}}
\newcommand{\quot}[1]{``#1''}

\maketitle

\thispagestyle{firstpage}
\pagestyle{mynormal}

\begin{abstract}
We present AlphaApollo, an agentic reasoning \emph{system} that targets two bottlenecks in foundation-model reasoning: (1) limited reasoning capacity for complex, long-horizon problem solving and (2) unreliable test-time evolution without trustworthy verification.
AlphaApollo orchestrates models and tools via three components: 
(i) \emph{multi-turn agentic reasoning}, which formalizes model-environment interaction with structured tool calls and responses; 
(ii) \emph{multi-turn agentic learning}, which applies turn-level reinforcement learning to optimize tool-use reasoning while decoupling actions from tool responses for stable training; and 
(iii) \emph{multi-round agentic evolution}, which refines solutions through a propose--judge--update loop with tool-assisted verifications and long-horizon memory.
Across seven math reasoning benchmarks and multiple model scales, AlphaApollo improves performance through reliable tool use (\(>\!85\%\) tool-call success), substantial gains from multi-turn RL (Avg@32: Qwen2.5-1.5B-Instruct \(1.07\%\!\rightarrow\!9.64\%\), Qwen2.5-7B-Instruct \(8.77\%\!\rightarrow\!20.35\%\)), and improvements from evolution (e.g., Qwen2.5-3B-Instruct \(5.27\%\!\rightarrow\!7.70\%\), Qwen2.5-14B-Instruct \(16.53\%\!\rightarrow\!21.08\%\)).
This project is still ongoing.
\footnote{Project website: \url{https://alphaapollo.org/}.}
We welcome feedback from the community and will frequently update the 
\href{https://github.com/tmlr-group/AlphaApollo}{source code}
and \href{https://arxiv.org/abs/2510.06261}{technical report}.
\end{abstract}


\section{Introduction}
\label{sec: intro}

Foundation models (FMs) increasingly power diverse applications via explicit \textit{reasoning}, decomposing complex tasks into manageable steps.
Yet single-model reasoning remains insufficient for frontier problems or real-world tasks.
As of December 2025, GPT-5 achieves $25.3\%$ and Gemini 2.5 Pro $21.6\%$ on Humanity’s Last Exam \citep{phan2025humanity}, and $9.9\%$ and $4.9\%$ on ARC-AGI-2 \citep{chollet2025arc}. 
Beyond math and code, limited compute and domain knowledge restrict performance in biology, chemistry, and healthcare, undermining reliability for scientific discovery.

Two bottlenecks limit FM reasoning are \textit{(i) model-intrinsic capacity} to generate candidate solutions and \textit{(ii) test-time evolution} to refine them.
First, prompting and post-training are often used to enhance model capacity, but they largely depend on the base model’s priors, making it unclear whether observed gains reflect \textit{emergent} capabilities or mere elicitation from pre-training (e.g., step-by-step reasoning and self-reflection).
Core reasoning skills (e.g., exact calculus and symbolic manipulation) remain constrained by next-token prediction~\citep{yang2024number, wang2025hierarchical}.
Second, without ground-truth verification, test-time evolution often relies on the model’s own judgments, which can be subjective and unreliable~\citep{gao2025survey}.
Finally, scalable parallel evolution remains under-explored, limiting efficiency and multi-model coordination, while long-horizon evolution cannot be achieved without an effective mechanism of memory.

\begin{figure}[t!]
  \centering
  \includegraphics[width=1.0\linewidth]{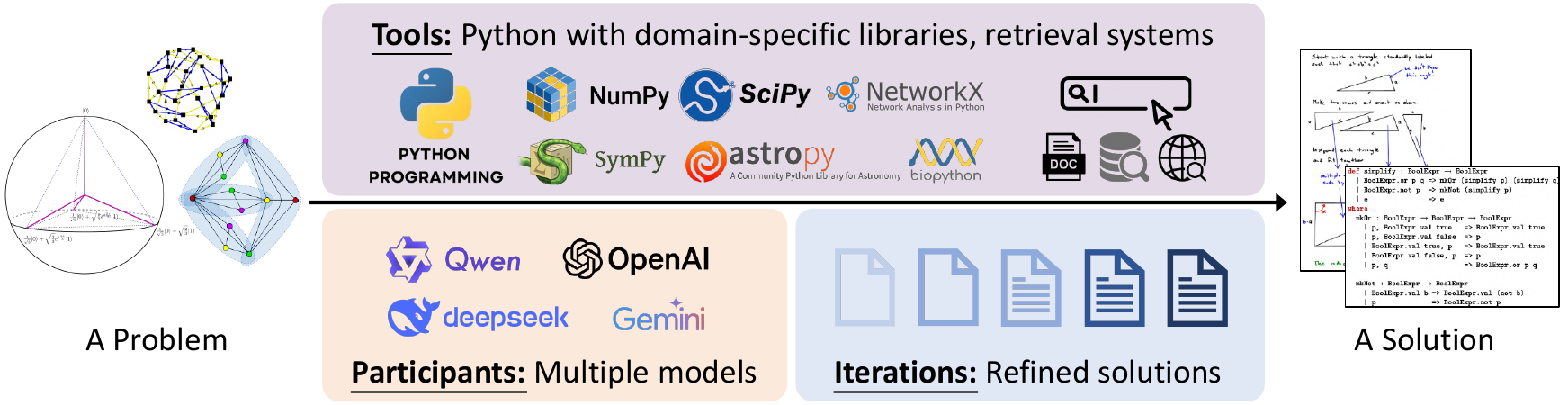}
  \vspace{-10pt}
  \caption{An overview of the AlphaApollo System for problem-solving with foundation models.}
  \label{fig: framework}
  \vspace{-8pt}
\end{figure}

This work introduces \textit{AlphaApollo}, an agentic reasoning \textit{system} designed to overcome these bottlenecks. 
Its design principle is to \textit{orchestrate} models and tools into a \textit{self-evolving system} for deep agentic reasoning (as in Fig. \ref{fig: framework}).
AlphaApollo adopts a strategy of setting clear goals, concentrating expertise and resources, and coordinating systematic collaboration under organizational support, thereby enabling it to tackle complex problems. 
Specifically, AlphaApollo centers on three features to transcend the capacity limits of a single model:
\begin{itemize}[leftmargin=*]
\item
\textbf{Multi-turn agentic reasoning.}
AlphaApollo solves tasks through iterative model-environment interaction: the model produces a structured action (tool call or answer), the environment executes tools and returns feedback, and the accumulated history serves as dynamic memory.

\item
\textbf{Multi-turn agentic learning.}
AlphaApollo post-trains at the turn level, optimizing the model’s \emph{actions} (reasoning and tool invocation) while decoupling them from tool responses, which stabilizes RL/SFT and improves tool-use decisions (what to call, what to query, when to stop).

\item
\textbf{Multi-round agentic evolution.}
At test time, AlphaApollo iteratively refined solutions with a propose-judge-update loop, using tool-assisted checking and selective memory to refine candidates over multiple rounds. This evolution can coordinate multiple models in solving one problem.
\end{itemize}

Empirically, we evaluate AlphaApollo on seven mathematical reasoning benchmarks across multiple model scales and settings. 
AlphaApollo achieves consistent gains from agentic reasoning alone (e.g., for Avg@32, Qwen2.5-3B: \(4.66\%\!\rightarrow\!4.72\%\); Qwen2.5-14B: \(10.82\%\!\rightarrow\!13.49\%\)), driven by reliable tool use with over \(85\%\) tool-call success across datasets. 
Multi-turn RL further boosts Avg@32 substantially (Qwen2.5-1.5B: \(1.07\%\!\rightarrow\!9.64\%\); Qwen2.5-3B: \(4.72\%\!\rightarrow\!13.35\%\); Qwen2.5-7B: \(8.77\%\!\rightarrow\!20.35\%\)). 
Finally, test-time agentic evolution delivers additional scalable gains (Qwen2.5-3B: \(4.79\%\!\rightarrow\!6.92\%\); Qwen2.5-7B: \(9.24\%\!\rightarrow\!10.79\%\); Qwen2.5-14B: \(16.53\%\!\rightarrow\!21.08\%\)), demonstrating reliable tool use, effective learning, and iterative self-improvement.

\section{AlphaApollo}
\label{sec: framework}


\begin{figure}[t!]
\centering
\includegraphics[width=0.8\textwidth]{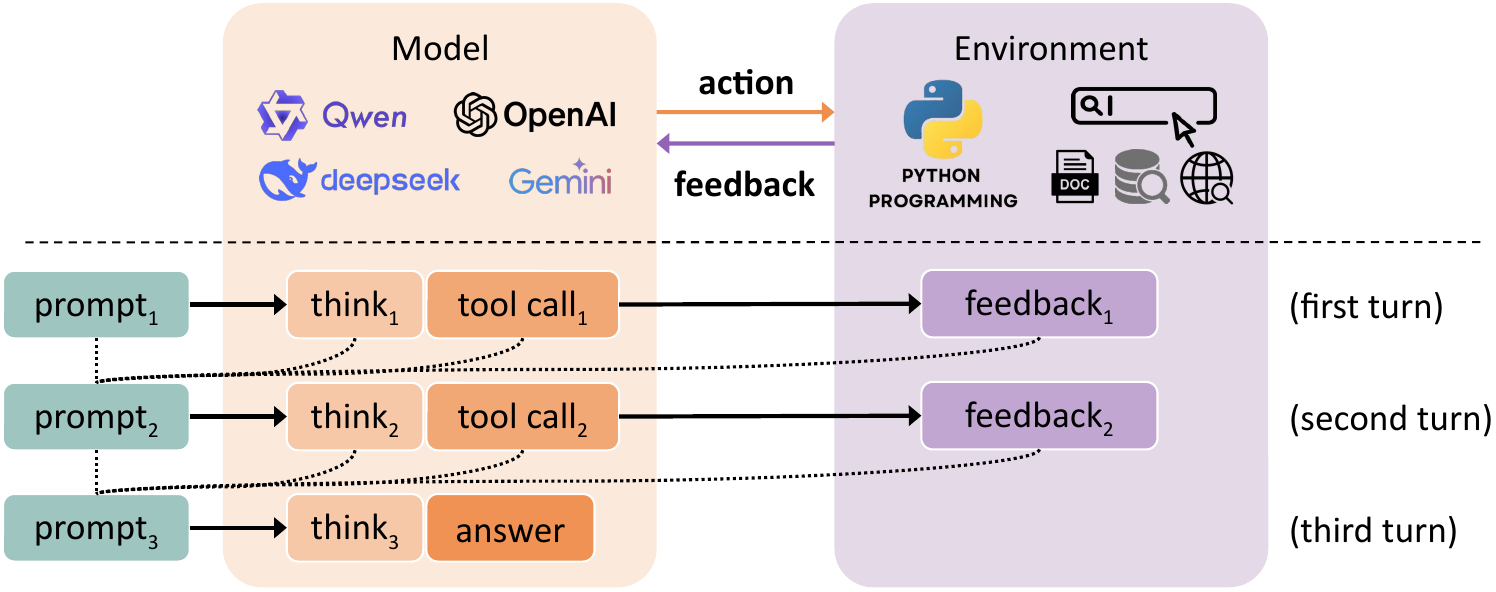}
\caption{
AlphaApollo’s rollout is an iterative model-environment interaction. The model outputs an action, and the environment provides feedback. Each turn’s trajectory is used as the prompt for the next turn, enabling dynamic memory. The final answer is produced when no tool calls are made.
}
\vspace{-.1in}
\label{fig: rollout}
\end{figure}

\subsection{Multi-turn Agentic Reasoning}
\label{ssec: agentic reasoning}

AlphaApollo structures the agentic reasoning as a \textit{multi-turn interaction} between \textit{model} and \textit{environment} (Figure~\ref{fig: rollout}).
In each turn, the model $\pi_\theta$ performs reasoning and invokes a tool call if needed, while the environment captures the tool call, executes the corresponding tool, and provides the tool responses back to the model.
Specifically, a trajectory $\tau_t$ at $t$-th interacttion turn incorporates \textit{prompt} $p_t$ (the model's input), model \textit{output} $o_t$, and environment \textit{feedback} $f_t$, namely, $\tau_t = (p_t, o_t, f_t)$:
\begin{itemize}[leftmargin=*]
\item
The \textit{prompt} $p_t$ incorporates essential information from previous $t-1$ turns.
For example, in a three-turn interaction, we have $p_1$ (the initial prompt), $p_2 = (p_1, o_1, f_1)$, and $p_3 = (p_2, o_2, f_2)$.

\vspace{-.05in}
\item
The \textit{output} $o_t$ consists of the thinking process, followed by either a tool call or the final answer.

\vspace{-.05in}
\item
The \textit{feedback} $f_t$ contains the tool response after executing the tool call.
\end{itemize}
This model-environment interaction terminates in two situations: (1) the model produces a final answer or (2) reaches the predefined maximum number of turns $T_{\max}$.
We denote the number of interacted turns in a reasoning trajectory before termination as $T$ ($T \leq T_{\max}$), and the final-turn trajectory as $\tau_T$.
Details of the model-environment synergies are introduced as follows.

\vspace{-.05in}
\paragraph{Environment side.}
The environment is responsible for (1) hosting tools, (2) parsing the output $o_t$, (3) executing the tool call in $o_t$, and (4) returning the feedback $f_t$ to the model. Specifically:

\vspace{-.05in}
\begin{itemize}[leftmargin=*]
\item
\textbf{Hosting tools.}
Currently, AlphaApollo provides two types of tools: 
\textit{computational tools} to solve precise calculations and \textit{retrieval tools} to retrieve necessary knowledge for problem solving (detailed implementations are introduced in Appendix~\ref{app: discussion_agentic_reasoning}).
These tools are implemented as callable functions for the model, while users can easily plug in custom tools for extension.

\vspace{-.05in}
\item
\textbf{Parsing outputs.} 
The environment parses the model's output $o_t$ to identify two contents:
(1) \textit{tool calls} wrapped within tool-specific tags (e.g., \texttt{<python\_code>} and \texttt{</python\_code>} for computational tools, \texttt{<local\_rag>} and \texttt{</local\_rag>} for retrieval tools) and (2) the \textit{final answer} wrapped within \texttt{<answer>} and \texttt{</answer>} tags.
The interaction terminates if the final answer is extracted.

\vspace{-.05in}
\item
\textbf{Executing tools.}
The system routes each tool call (parsed in the above step) to the corresponding tool hosting in the system.
In AlphaApollo, each trajectory is handled by a separate environment, so tool calls across different trajectories are executed \textit{asynchronously}, enabling efficient parallelism.

\item
\textbf{Returning feedback.} 
The results of tool execution are wrapped within \texttt{<tool\_response>} and \texttt{</tool\_response>} tags and returned to the model as the feedback $f_t$.
\end{itemize}

\vspace{-.05in}
\paragraph{Model side.} 
Given prompt $p_t$, the model $\pi_\theta$ generates output $o_t$ and receives environment feedback $f_t$.
The system updates the memory
and constructs the next-turn prompt $p_{t+1}$.
Specifically:

\vspace{-.05in}
\begin{itemize}[leftmargin=*]
\item 
\textbf{Inference infrastructure.}
AlphaApollo can generate the next token locally or remotely, with a particular model.
The supported \textit{inference backends} include vLLM \citep{kwon2023efficient}, SGLang \citep{zheng2024sglang}, HuggingFace Transformers \citep{Huggingface}, and external APIs \citep{openaiapi}.
AlphaApollo employs Ray~\citep{moritz2018ray}
to parallelize trajectory generation. 
For a batch of questions, the system spawns multiple environments, each handling the interaction for one trajectory, and all environments run concurrently until all trajectories are collected.

\vspace{-.05in}
\item 
\textbf{Output generation.}
The generated output $o_t$ follows a structured format: it begins with a reasoning trajectory where the model articulates its thought process (enclosed within \texttt{<think>} and \texttt{</think>} tags), followed by either a tool call or the final answer. This structured format aligns with the parsing patterns on the environment side, ensuring seamless model-environment interaction.

\vspace{-.05in}
\item 
\textbf{Memory management.}
The prompt in each turn includes three types of information: the \textit{task description} that specifies the user query and output format, \textit{tool specifications} that describe the available tools and their usage formats, and \textit{interaction history} that records previous outputs and feedback from earlier turns if available.
Notably, AlphaApollo supports flexible memory strategies: by default, it concatenates all previous interactions as context, while for long-horizon tasks, it supports long-term memory that selectively retains high-quality trajectories (details in Sec.~\ref{ssec: agentic evolution}).

\end{itemize}

\subsection{Multi-turn Agentic Learning}
\label{ssec: learning}

\begin{figure}[!t]
    \centering
    \includegraphics[width=0.8\textwidth]{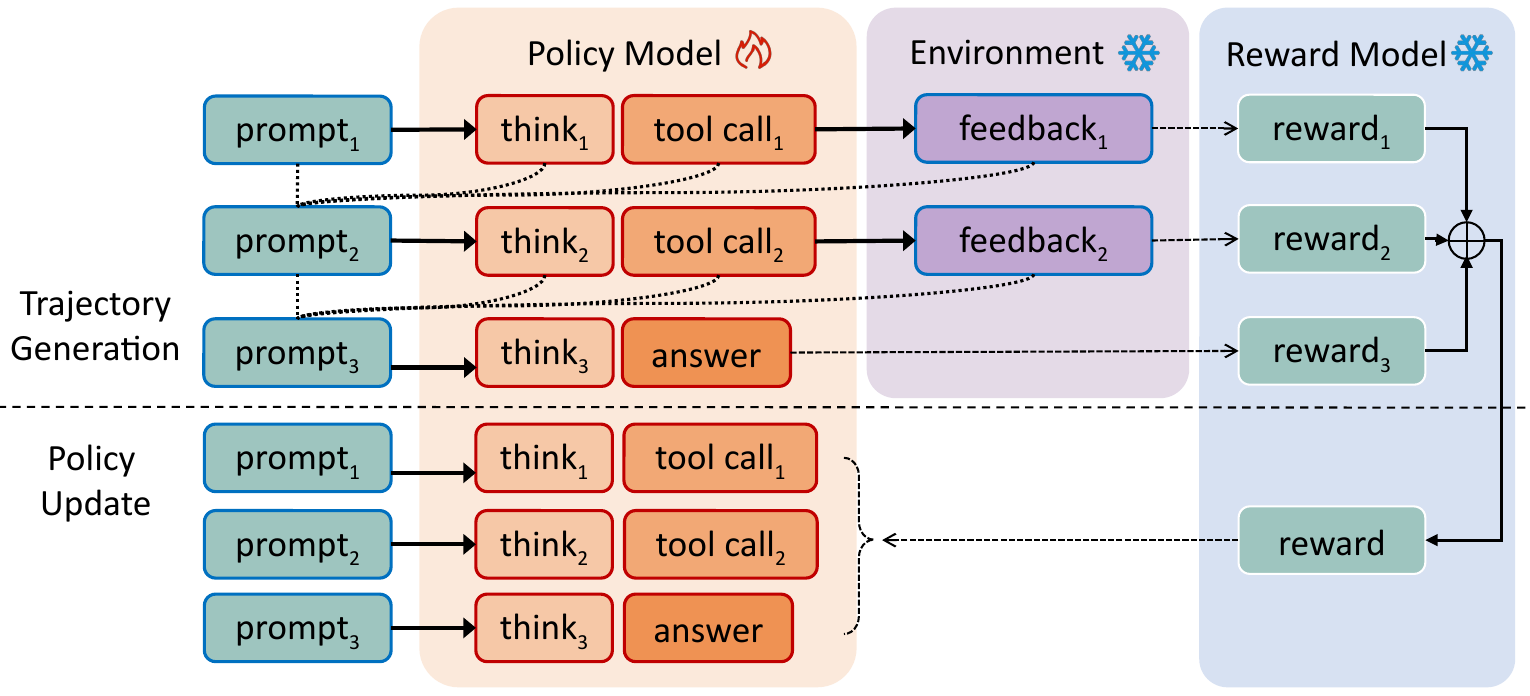}
    \caption{Illustration of multi-turn agentic RL in AlphaApollo. During generation, per-turn rewards are assigned based on model outputs and environment feedback, and summed to form the trajectory reward. During policy update, the policy is updated at each turn with non-model outputs masked.}
\label{fig:turn-level-optimization}
\vspace{-.1in}
\end{figure}

Based on the turn-level agentic reasoning system (Sec.~\ref{ssec: agentic reasoning}) that decouples model-generated content from tool responses, AlphaApollo incorporates VeRL~\citep{sheng2024hybridflow} (an open-source framework for RL) into a \textit{turn-level} optimization for stable and flexible agentic learning, 
along with versatile SFT/RL/parameter-efficient algorithms for versatile learning configurations.

\vspace{-.05in}
\paragraph{Turn-level optimization.}
The full trajectory $\tau_{T}$ comprises multiple turns of model-generated content and the environment feedback.
Notably, optimizing on the environment feedback can introduce instability~\citep{jin2025search, feng2025group}.
We discuss the trajectory-level optimization in Appendix~\ref{app: discussion_agentic_learning}.
For stable learning on the multi-turn interaction, AlphaApollo adopts turn-level optimization.
This paradigm decouples model generations $o_t$ from environment feedback $f_t$, thereby facilitating targeted optimization on $o_t$. 
For example, the turn-level GRPO is formulated as:
\begin{equation}
\begin{aligned}
\mathcal{J}_{\text{Turn-GRPO}}(\theta)
\;=\;
\mathbb{E}_{(q,a) \sim \mathcal{D},\{o^{(i)}\}_{i=1}^{N} \sim \pi_{\text{old}}(\cdot|q)}
\Big[
\frac{1}{N}
\sum_{i=1}^{N}
\frac{1}{T^{(i)}}\sum_{t=1}^{T^{(i)}}
\frac{1}{|o^{(i)}_t|}\sum_{k=1}^{|o^{(i)}_t|}
\min  \Big(\frac{\pi_\theta\big(o_{t,k}^{(i)} \mid p_{t}, o_{t,<k}^{(i)}\big)}
{\pi_{\text{old}}\big(o_{t,k}^{(i)} \mid p_{t}, o_{t,<k}^{(i)}\big)}\, A_{t,k}^{(i)},\; 
\\
\operatorname{clip}\big(\frac{\pi_\theta\big(o_{t,k}^{(i)} \mid p_{t}, o_{t,<k}^{(i)}\big)}
{\pi_{\text{old}}\big(o_{t,k}^{(i)} \mid p_{t}, o_{t,<k}^{(i)}\big)}, \,1-\epsilon,\,1+\epsilon\big)\,A_{t,k}^{(i)}
\Big)-\beta\,\mathbb{D}_{\mathrm{KL}}\bigl[\pi_\theta \,\big\|\,\pi_{\mathrm{ref}}\bigr] \,
\Big],
\label{eq:alphaapollo_grpo_objective}
\end{aligned}
\end{equation}
where $o_{t,k}^{(i)}$ denotes $k$-th token of the model's output in $t$-th turn of the $i$-th trajectory,
$N$ is the number of group samples for GRPO, $A_{t,k}^{(i)} = \frac{r(\tau_{T^{(i)}}) - \bar{r}}{\sigma_r}$ where $r$ is the reward by evaluating the answer proposed in turn $T$, $\bar{r}$ and $\sigma_r$ are the baseline and the normalization across all $N$ trajectories of one question.
$A$ indicates the advantage and $\epsilon$ is the clipping bound, and $\beta\,\mathbb{D}_{\mathrm{KL}}[\pi_\theta \,\big\|\,\pi_{\mathrm{ref}}]$ is the KL divergence term using an unbiased estimator~\citep{kl_approx} weighted by $\beta$.
Here, token importance ratios 
$
\nicefrac{\pi_\theta(o_{t,k}^{(i)} \mid p_t, o_{t,<k}^{(i)})}{\pi_{\text{old}}(o_{t,k}^{(i)} \mid p_t, o_{t,<k}^{(i)})}
$ 
are computed conditioned on $p_{t}$ and generated tokens $o_{t,<k}$, which incorporate the query and prior turns' generations and tool responses. 

Similarly, the turn-level SFT is formulated as:

\begin{equation}
\mathcal{J}_{\text{Turn-SFT}}(\theta)
= \mathbb{E}_{\tau_{T}\sim \mathcal{D}_{\text{SFT}}} 
\left[ 
\sum_{t=1}^{T}
\sum_{k=1}^{|o_t|} 
\log \pi_\theta(o_{t, k} \mid p_t, o_{t, <k}) 
\right],
\end{equation}
where $\tau_T$ is a full trajectory curated from the multi-turn agentic reasoning (Sec.~\ref{ssec: agentic reasoning}).

\vspace{-.05in}
\paragraph{Post-training features.} To create a versatile and user-friendly agentic learning system, AlphaApollo incorporates the following post-training features:
\begin{itemize}[leftmargin=*]
    \item \textbf{Algorithms:} AlphaApollo supports common SFT/RL algorithms for post-training, such as PPO~\citep{schulman2017proximal}, GRPO~\citep{shao2024deepseekmath}, and DAPO~\citep{yu2025dapo}.
    \item \textbf{Models:} AlphaApollo enables agentic learning across various foundation model families, including Qwen2.5~\citep{yang2024qwen2}, Qwen3~\citep{yang2025qwen3}, and Llama3.2~\citep{grattafiori2024llama}. Additionally, it provides a multi-modal preprocessing and training pipeline for post-training with multi-modal foundation models, such as Qwen2.5-VL families~\citep{bai2025qwen2}.
    \item \textbf{Flexible advantage estimation:} 
    AlphaApollo supports both trajectory- and turn-level advantage estimation in RL, making it compatible with emerging process reward modeling algorithms.
    \item \textbf{Parameter-efficient fine-turning:} Leveraging FSDP and vLLM as backends, AlphaApollo enables LoRA~\citep{hu2022lora} for efficient post-training of large-scale models.
\end{itemize}

\subsection{Multi-round Agentic Evolution}
\label{ssec: agentic evolution}

\begin{figure*}[!t]
\centering
\includegraphics[width=0.8\textwidth]{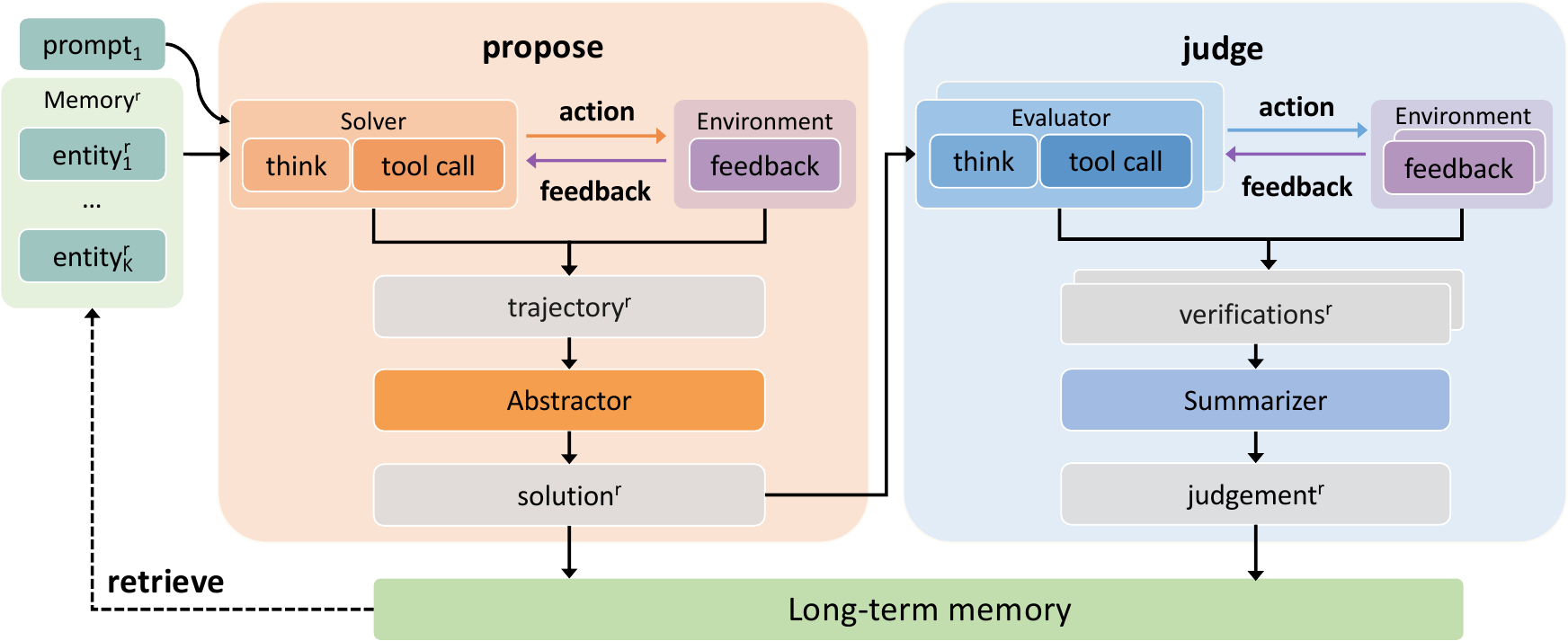}
\caption{Illustration of multi-round agentic evolution in AlphaApollo. The model iteratively refines its strategies through a propose-judge-update evolutionary loop. A long-term memory is introduced to prevent future errors and promote efficient strategies in subsequent rounds.}
\label{fig:test-time-evolution}
\vspace{-.1in}
\end{figure*}

AlphaApollo employs a test-time evolution mechanism to iteratively refine solutions (Figure~\ref{fig:test-time-evolution}). Operating via a propose-judge-update loop, the system coordinates multiple agents to generate solutions, evaluate them, and store the judgment in long-term memory to guide subsequent rounds.

\vspace{-.05in}
\paragraph{Overview.}
We denote the system's initial input as $p_1$ and the long-term memory as $\mathcal{M}$, initialized as $\mathcal{M}^{(1)} = \emptyset$. The process iterates over multiple rounds $r = \{1, \dots, R\}$. In each round, the system coordinates four specialized agents $\pi$ to instantiate the pipeline, yielding a \textit{candidate solution} $\hat{s}^{(r)}$ and \textit{diagnostic judgment} $j^{(r)}$. 
These outputs are formed as a memory entity $(\hat{s}^{(r)}, j^{(r)})$, which is committed to $\mathcal{M}$. The $\mathcal{M}^{(r+1)}$ is retrieved from $\mathcal{M}$ to condition the policy $\pi$ for the subsequent evolution.
Detailed prompts and implementations can be found in Appendix~\ref{app: evolution prompts}.
The pipeline is formulated as follows:
\begin{equation}
\begin{aligned}
\textbf{1. Propose:}& \quad \tau^{(r)} \sim \pi_{\text{solver}}(\cdot \mid p_1, \mathcal{M}^{(r)}),
\quad
\hat{s}^{(r)} \sim \pi_{\text{abs}}(\tau^{(r)}); 
\\
\textbf{2. Reflection:}& \quad \mathcal{V}^{(r)} = \{v_i\}_{i=1}^N \sim \pi_{\text{eval}}(\cdot \mid \hat{s}^{(r)}), \quad j^{(r)} \sim \pi_{\text{sum}}(\mathcal{V}^{(r)}); 
\\
\textbf{3. Update:}& 
\quad 
\mathcal{M} = \mathcal{M} \cup \{ (\hat{s}^{(r)}, j^{(r)}) \},
\quad
\mathcal{M}^{(r+1)} = \mathrm{Retrieve}(\mathcal{M}).
\end{aligned}
\label{eqn:evolving_pipeline}
\end{equation}
We detail the role of each agent in this pipeline below.
\begin{itemize}[leftmargin=*, itemsep=4pt, topsep=4pt]
\item \texttt{Solver} ($\pi_{\text{solver}}$): \textit{Given a problem and memory state, generates a multi-turn reasoning trajectory.} The solver interacts with the environment to generate a complete trajectory $\tau^{(r)}$.
\item \texttt{Abstractor} ($\pi_{\text{abs}}$): \textit{Given $\tau^{(r)}$, generates a condensed solution.}
The abstractor compresses $\tau^{(r)}$ into a condensed solution $\hat{s}^{(r)}$ to fit the context window while preserving essential information.
\item \texttt{Evaluator} ($\pi_{\text{eval}}$): \textit{Given $\hat{s}^{(r)}$, generates verifications $\mathcal{V}^{(r)}$.} For deterministic tasks, it integrates execution feedback. For open-ended tasks, it aggregates $N$ sampled critiques via \emph{majority voting}.
\item \texttt{Summarizer} ($\pi_{\text{sum}}$): \textit{Given $\mathcal{V}^{(r)}$, generates comprehensive judgement $j^{(r)}$.} Synthesize $\mathcal{V}^{(r)}$ into a high-level judgment $j^{(r)}$, remove redundancy to provide clean advice for $\pi_\text{solver}$ in future rounds.
\end{itemize}

\vspace{-.05in}
\paragraph{Long-term memory for evolution.}
To enable long-horizon evolution, the memory module supports efficient retrieval to prevent excessive context, including the overlong content of correctness and solution.
Specifically, the long-term memory operates in two phases:
\begin{itemize}[leftmargin=*, itemsep=4pt, topsep=4pt]
\item \textbf{Store:} 
At the end of round $r$, the system transitions the memory state by appending the new entity: $\mathcal{M} \cup \{ (\hat{s}^{(r)}, j^{(r)}) \}$. This accumulation ensures the history of exploration is preserved.
\item \textbf{Retrieval:} 
To yield $\mathcal{M}^{(r+1)}$, the system selects the Top-$K$ entries from the memory under a weighted scoring function prioritizing (1) the correct solution and (2) shorter solutions in correct groups.
\end{itemize}
Retrieving high-level judgment $j$ promotes promising strategies while preventing recurrent errors. Furthermore, the module employs thread-safe locks to decouple storage from execution, enabling asynchronous agents to reliably synchronize insights regardless of latency.


\paragraph{Parallel Evolution.} To accelerate reasoning, AlphaApollo distributes the evolution pipeline across multiple worker threads. This architecture supports heterogeneous solvers (e.g., mixing open-source models with proprietary APIs or varying sampling temperatures). Agents synchronize via the shared long-term memory: when one solver commits a high-quality solution, it becomes available to others, fostering collective intelligence that guides the evolution toward a better solution.

\section{Experiments}
\label{sec: exp}

\paragraph{Experiment setup.} 
We evaluate AlphaApollo across various models, from Qwen2.5-1.5B-Insturct to Qwen2.5-14B-Instruct. We evaluate on AIME24, AIME25, CMIMC, HMMT (Feb and Nov), BRUMO, and SMT. 
We set a maximum of 4 interaction rounds. Details are in Appendix~\ref{app: discussion_agentic_reasoning}, \ref{app: discussion_agentic_learning}, and \ref{app: discussion_agentic_evolution}.

\begin{table*}[t!]
\small
\centering
\caption{Agentic reasoning results (Avg@32/Pass@32 in \%). Base is evaluated without training and tools; AlphaApollo is evaluated with tools enabled without training. \textbf{Bold} marks the better results.
}
\vspace{-8pt}
\label{tab:mainresults}
\setlength{\tabcolsep}{5.2pt}
\renewcommand{\arraystretch}{1.1}
\fontsize{8}{8}\selectfont
\begin{tabular}{lcc cc cc}
\toprule
\multirow{3}{*}{\textbf{Dataset}} 
& \multicolumn{2}{c}{\textbf{Qwen2.5-3B-Instruct}}
& \multicolumn{2}{c}{\textbf{Qwen2.5-7B-Instruct}}
& \multicolumn{2}{c}{\textbf{Qwen2.5-14B-Instruct}} \\
\cmidrule(lr){2-3}\cmidrule(lr){4-5}\cmidrule(lr){6-7}
& \textbf{Base} & \textbf{AlphaApollo}
& \textbf{Base} & \textbf{AlphaApollo}
& \textbf{Base} & \textbf{AlphaApollo} \\
\midrule

\textbf{AIME24}
& 5.21 / 26.67
& \textbf{5.52} / \textbf{30.00} 
& \textbf{12.19} / 36.67
& 8.85 / \textbf{56.67} 
& 13.44 / 46.67
& \textbf{16.98}/ \textbf{60.00} 
\\

\textbf{AIME25}
& \textbf{3.23} / \textbf{36.67}
& 2.19 / 23.33
& \textbf{8.23} / 36.67
& 6.15 / \textbf{36.67}
& \textbf{12.29} / 43.33
& 11.77 / \textbf{46.67}
\\

\textbf{CMIMC25}
& 1.17 / 17.50
& \textbf{3.36} / \textbf{30.00} 
& 4.02 / 30.00
& \textbf{7.42} / \textbf{40.00}
& 4.61 / 27.50
& \textbf{11.48} / \textbf{40.00}
\\

\textbf{HMMT25 Feb}
& 0.52 / 10.00
& \textbf{2.60} / \textbf{16.67} 
& 2.08 / 23.33
& \textbf{6.67} / \textbf{33.33} 
& 3.23 / 23.33
& \textbf{9.58} / \textbf{40.00} 
\\

\textbf{HMMT25 Nov}
& \textbf{3.02} / \textbf{20.00}
& 2.50 / 23.33 
& 5.00 / 23.33
& \textbf{6.04} / \textbf{26.67} 
& 5.73 / 20.00
& \textbf{7.29} / \textbf{23.33} 
\\

\textbf{BRUMO25}
& \textbf{11.25} / 40.00
& 8.44 / \textbf{46.67} 
& \textbf{18.23} / 50.00
& 17.18 / \textbf{50.00}
& 22.29 / 43.33
& \textbf{22.60} / \textbf{60.00} 
\\

\textbf{SMT 2025}
& 8.25 / 32.08
& \textbf{8.43} / \textbf{41.51} 
& \textbf{11.62} / 39.62
& 9.08 / \textbf{41.51} 
& 14.15 / 41.51
& \textbf{14.74} / \textbf{49.06} 
\\

\midrule

\multirow{2}{*}{\textbf{Average}}
& \multirow{2}{*}{4.66 / 26.13}
& \textbf{4.72} / \textbf{30.22} 
& \multirow{2}{*}{8.77 / 34.23}
& \textbf{8.77} / \textbf{40.69} 
& \multirow{2}{*}{10.82 / 35.10}
& \textbf{13.49} / \textbf{45.58} 
\\
& & {\scriptsize (0.06$\uparrow$) / (4.09$\uparrow$)}
& & {\scriptsize (0.00$\uparrow$) / (6.46$\uparrow$)}
& & {\scriptsize (2.67$\uparrow$) / (10.48$\uparrow$)}
\\
\bottomrule
\end{tabular}
\vspace{-8pt}
\end{table*}

\begin{table*}[t!]
\centering
\caption{
Agentic learning results (Avg@32 in \%) for AlphaApollo. 
No-training evaluates AlphaApollo using tools without training. 
LE (MATH-LightEval~\citep{hendrycksmath2021}), LIMR~\citep{li2025limr}, and DS (DeepScaleR~\citep{deepscaler2025}) denote training the model on the corresponding datasets. 
}
\vspace{-8pt}
\label{tab:training_results}
\setlength{\tabcolsep}{1.5pt}
\renewcommand{\arraystretch}{1.1}
\fontsize{8}{8}\selectfont
\begin{tabular}{lccccccccccccc}
\toprule
\multirow{4}{*}{\textbf{Dataset}} &
\multicolumn{4}{c}{\textbf{Qwen2.5-1.5B-Instruct}} &
\multicolumn{4}{c}{\textbf{Qwen2.5-3B-Instruct}} &
\multicolumn{4}{c}{\textbf{Qwen2.5-7B-Instruct}} \\
\cmidrule(lr){2-5}\cmidrule(lr){6-9}\cmidrule(lr){10-13}
& \multirow{3}{*}{\textbf{\makecell{No-\\training}}}&  \multicolumn{3}{c}{\textbf{AlphaApollo (training)}} & \multirow{3}{*}{\textbf{\makecell{No-\\training}}} & \multicolumn{3}{c}{\textbf{AlphaApollo (training)}}& \multirow{3}{*}{\textbf{\makecell{No-\\training}}} & \multicolumn{3}{c}{\textbf{AlphaApollo (training)}}\\
\cmidrule(lr){3-5}\cmidrule(lr){7-9}\cmidrule(lr){11-13}
& & +LE & +LIMR & +DS &
 & +LE & +LIMR & +DS &
  & +LE & +LIMR & +DS \\
\midrule

\textbf{AIME24}
& 0.63 &
3.77 & 3.74 & \textbf{8.96}
& 5.52 &
9.14 & 14.35 & \textbf{20.92}
& 8.85 &
22.91 & 19.40 & \textbf{25.50}

\\

\textbf{AIME25}
& 0.73 &
2.68 & \textbf{15.36} & 14.67
& 2.19
& 13.06 & \textbf{14.14} & 9.33
& 6.15
& 16.58 & 15.28 & \textbf{17.58}
\\

\textbf{CMIMC25}
& 0.63
& 5.25 & \textbf{7.78} & 7.11
& 3.36
& 6.32 & 7.80 & \textbf{10.58}
& 7.42
& 13.16 & 14.14 & \textbf{16.97}
\\

\textbf{HMMT25 Feb}
& 1.35
& 4.47 & \textbf{7.27} & 6.51
& 2.60 
& 12.27 & 12.61 & \textbf{14.07}
& 6.67
& 13.33 & 9.21 & \textbf{18.30}
\\

\textbf{HMMT25 Nov}
& 0.83
& 1.73 & \textbf{5.75} & 4.93
& 2.50 
& \textbf{6.48} & 5.29 & 4.21
& 6.04 
& 11.21 & 12.28& \textbf{13.11} 
\\

\textbf{BRUMO25}
& 1.98
& 8.77 & \textbf{15.96} & 14.98
& 8.44 
& \textbf{23.22} & 21.30 & 21.62
& 17.18
& 30.26 & 27.70& \textbf{33.10} 
\\

\textbf{SMT 2025}
& 1.36 
& 5.83 & \textbf{12.07} & 10.31
& 8.43 &
13.11 & \textbf{14.60} & 10.72
& 9.08 &
16.65 & \textbf{18.00} & 17.88
\\

\midrule

\multirow{2}{*}{\textbf{Average}}
& \multirow{2}{*}{1.07} 
& 4.64 & \textbf{9.70} & 9.64
& \multirow{2}{*}{4.72}&
11.94 & 12.87 & \textbf{13.35}
& \multirow{2}{*}{8.77} 
& 17.73 & 16.57& \textbf{20.35} \\

& & {\scriptsize (3.57$\uparrow$)} & {\scriptsize (8.63$\uparrow$)} & {\scriptsize (8.57$\uparrow$)} 
& 
& {\scriptsize (7.22$\uparrow$)}  & {\scriptsize (8.15$\uparrow$ )} & {\scriptsize (8.33$\uparrow$)} 
&
& {\scriptsize (8.96$\uparrow$)}  & {\scriptsize (7.80$\uparrow$)}  & {\scriptsize (11.58$\uparrow$)} 
\\
\bottomrule
\end{tabular}
\vspace{-8pt}
\end{table*}

\begin{table*}[t!]
\centering
\caption{
Agentic evolution results (accuracy in \%). 
w/o Evolution uses tools without evolution, while w/ Evolution enables agentic evolution with tools. 
}
\vspace{-8pt}
\label{tab:evolving_results}
\setlength{\tabcolsep}{8pt}
\renewcommand{\arraystretch}{1.1}
\fontsize{8}{8}\selectfont
\begin{tabular}{@{}lcccccc@{}}
\toprule
\multirow{3}{*}{\textbf{Datasets}} & 
\multicolumn{2}{c}{\textbf{Qwen2.5-3B-Instruct}} &
\multicolumn{2}{c}{\textbf{Qwen2.5-7B-Instruct}} &
\multicolumn{2}{c}{\textbf{Qwen2.5-14B-Instruct}}
\\
\cmidrule(lr){2-3} \cmidrule(lr){4-5} \cmidrule(lr){6-7} 
& \textbf{w/o Evolution} & \textbf{w/ Evolution} & \textbf{w/o Evolution} & \textbf{w/ Evolution} & \textbf{w/o Evolution} & \textbf{w/ Evolution}
\\
\midrule

\textbf{AIME24} 
& 6.67 & \textbf{10.00}
& 11.67 & \textbf{12.50}
& 19.17 & \textbf{23.33} 
\\

\textbf{AIME25} 
& \textbf{7.50} & \textbf{7.50}
& 8.33 & \textbf{11.67}
& 18.33 & \textbf{24.17} 
\\

\textbf{CMIMC 25} 
& 3.12 & \textbf{6.25}
& 10.00 & \textbf{12.50}
& 15.00 & \textbf{19.38} 
\\

\textbf{HMMT25 Feb} 
& 3.33 & \textbf{5.00} 
& 10.00 & \textbf{10.83} 
& 15.83 & \textbf{20.00}
\\

\textbf{HMMT25 Nov} 
& 1.67 & \textbf{4.17}
& \textbf{7.50} & \textbf{7.50} 
& 7.50 & \textbf{12.50}
\\

\textbf{BRUMO25} 
& 7.50 & \textbf{12.50}
& 11.67 & \textbf{17.50} 
& 26.67 & \textbf{31.67} 
\\

\textbf{SMT 2025} 
& 7.08 & \textbf{8.49}
& 8.02 & \textbf{8.96}
& 13.21 & \textbf{16.51}
\\

\midrule
\multirow{2}{*}{\textbf{Average}} 
& \multirow{2}{*}{5.27} & \textbf{7.70} 
& \multirow{2}{*}{9.60} & \textbf{11.64} 
& \multirow{2}{*}{16.53} & \textbf{21.08} 
\\

& & {\scriptsize (2.43$\uparrow$)} &
& {\scriptsize (2.04$\uparrow$)} &
&  {\scriptsize (4.55$\uparrow$)} 
\\
\bottomrule
\end{tabular}
\vspace{-8pt}
\end{table*}

\paragraph{Main results.}
We evaluate AlphaApollo from three aspects: agentic reasoning, agentic learning, and agentic evolution.
The corresponding results are summarized in Tables~\ref{tab:mainresults}, \ref{tab:training_results}, and \ref{tab:evolving_results}, respectively. 

\begin{figure}[t!]
    \centering
    \begin{subfigure}[t]{0.32\textwidth}
        \centering
        \includegraphics[width=\linewidth]{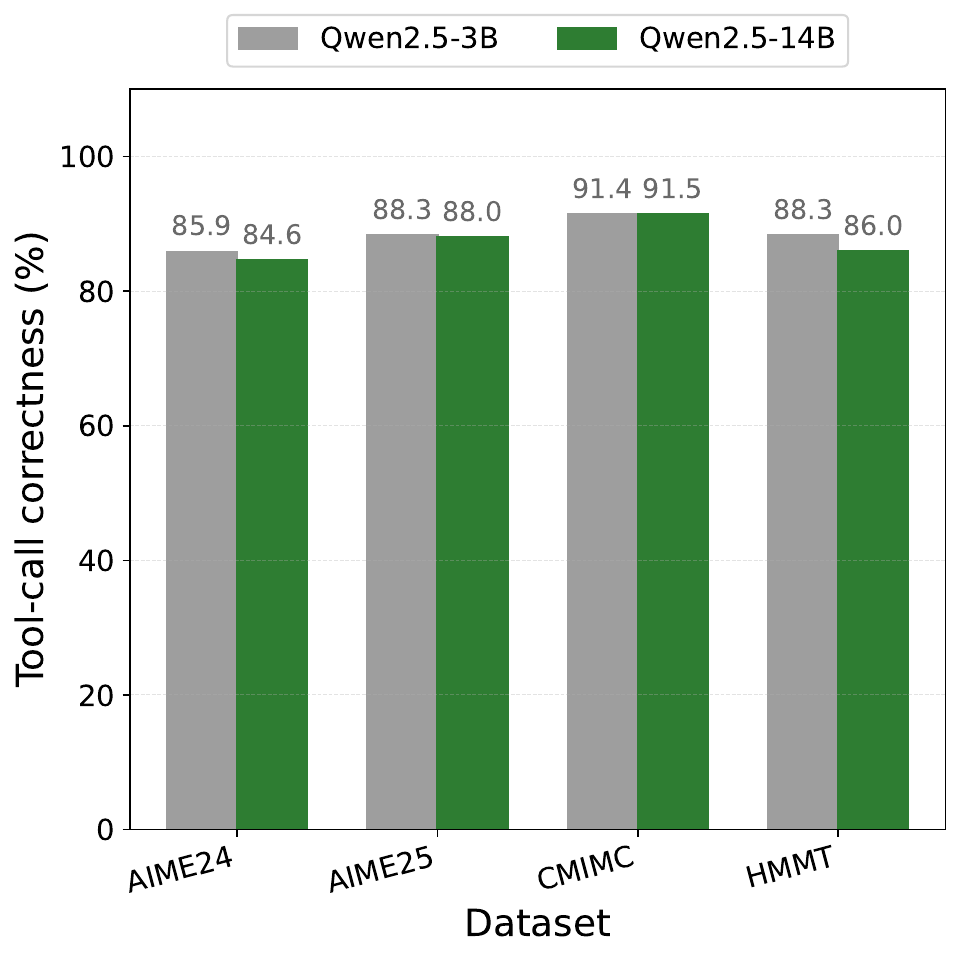}
        \vspace{-12pt}
        \caption{Tool-call success rate (\%) in reasoning with AlphaApollo.}
        \label{fig: tool-call-correctness}
    \end{subfigure}
    \hfill
    \begin{subfigure}[t]{0.32\textwidth}
        \centering
        \includegraphics[width=\linewidth]{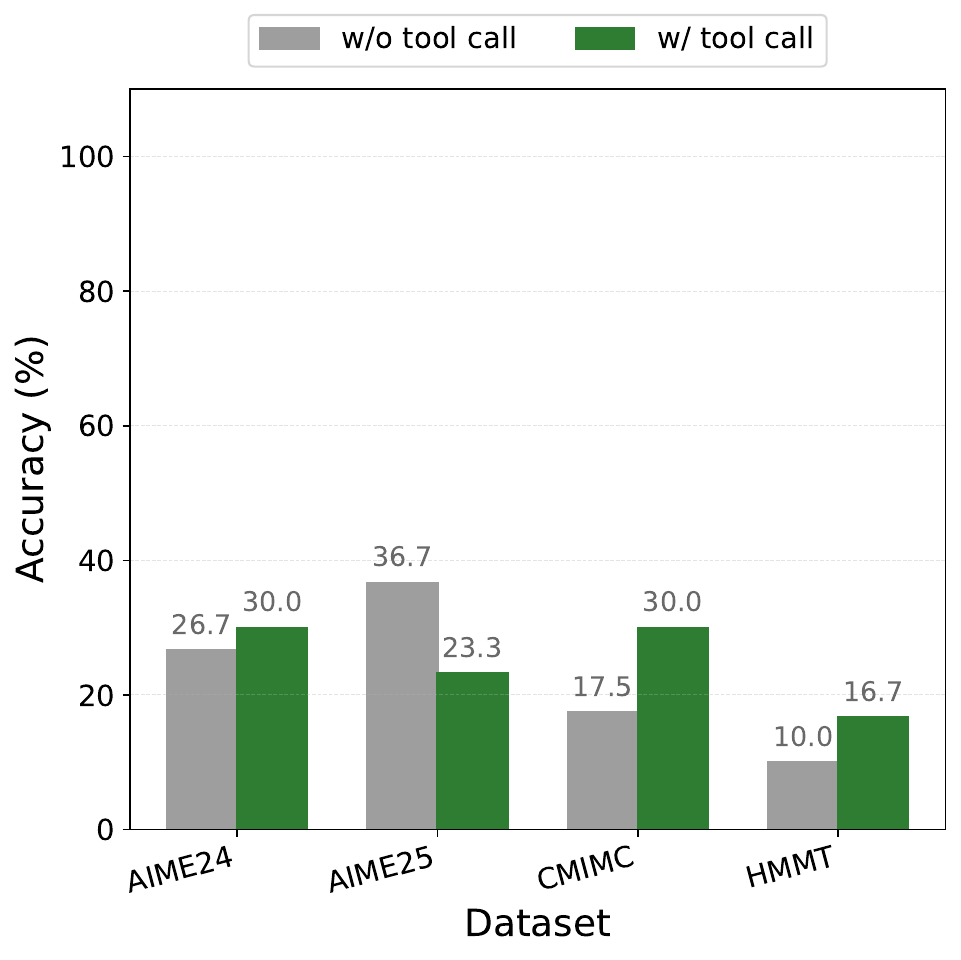}
        \vspace{-12pt}
        \caption{Accuracy (\%) w/ and w/o tool calls on Qwen2.5-3B-Instruct.}
        \label{fig: qwen25_3b_tool_vs_notool}
        
    \end{subfigure}
    \hfill
    \begin{subfigure}[t]{0.32\textwidth}
        \centering
        \includegraphics[width=\linewidth]{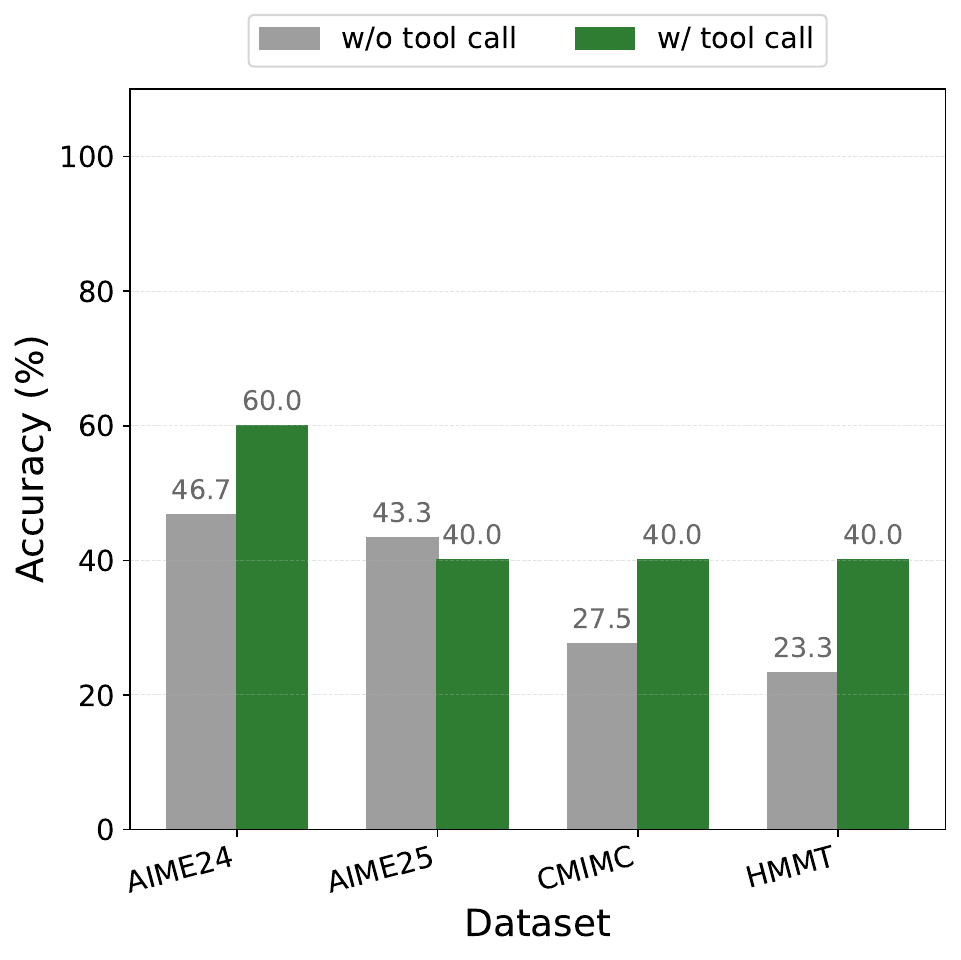}
        \vspace{-12pt}
        \caption{Accuracy (\%) w/ and w/o tool calls on Qwen2.5-14B-Instruct.}
        \label{fig: qwen25_14b_tool_vs_notool}
    \end{subfigure}
    \vspace{-8pt}
    \caption{Tool-call performance and accuracy enhancements in AlphaApollo using Qwen models. }
    \label{fig: analysis1}
\vspace{-8pt}
\end{figure}

\paragraph{Agentic reasoning with tools enables consistent performance gain across model scales.}
In Table~\ref{tab:mainresults}, Figures~\ref{fig: qwen25_3b_tool_vs_notool}, and \ref{fig: qwen25_14b_tool_vs_notool}, AlphaApollo consistently outperforms no-tool versions across all models, demonstrating improvements from 4.66\% to 4.72\% on Qwen2.5-3B-Instruct and from 10.82\% to 13.49\% on Qwen2.5-14B-Instruct.
The observed gains in reasoning capability primarily stem from improved tool-call reliability. In Figure~\ref{fig: tool-call-correctness}, AlphaApollo’s model-friendly tool-calling module achieves a success rate exceeding 85\% across all evaluated datasets.
These demonstrate that AlphaApollo consistently excels in boosting FMs' capabilities, both in reasoning and tool utilization.

\begin{figure}[t!]
    \centering
    \begin{subfigure}[t]{0.32\textwidth}
        \centering
        \includegraphics[width=\linewidth]{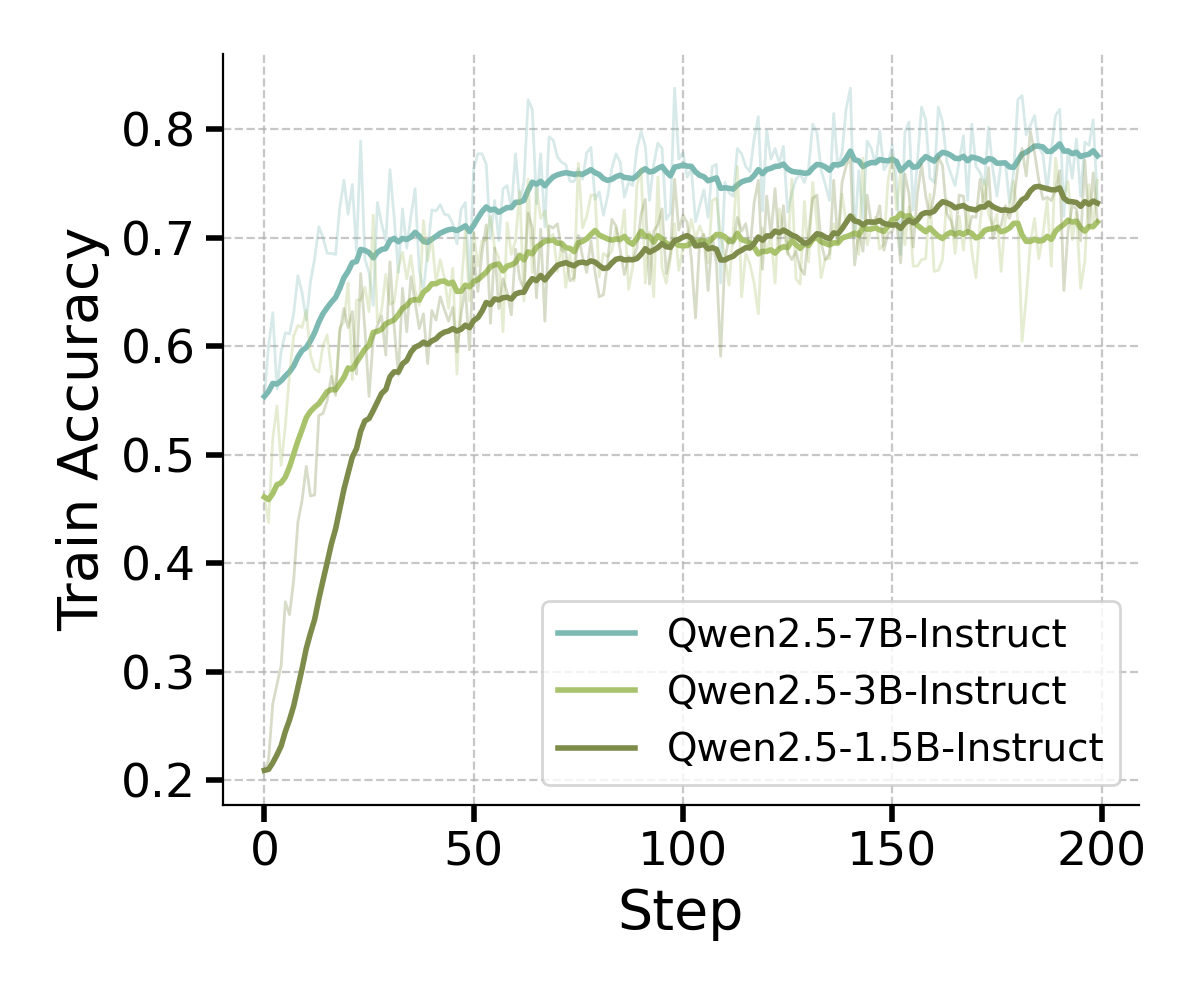}
        \vspace{-20pt}
        \caption{MATH-LightEval}
        \label{fig: training-lightevalmath-succ-rate}
    \end{subfigure}
    \hfill
    \begin{subfigure}[t]{0.32\textwidth}
        \centering
        \includegraphics[width=\linewidth]{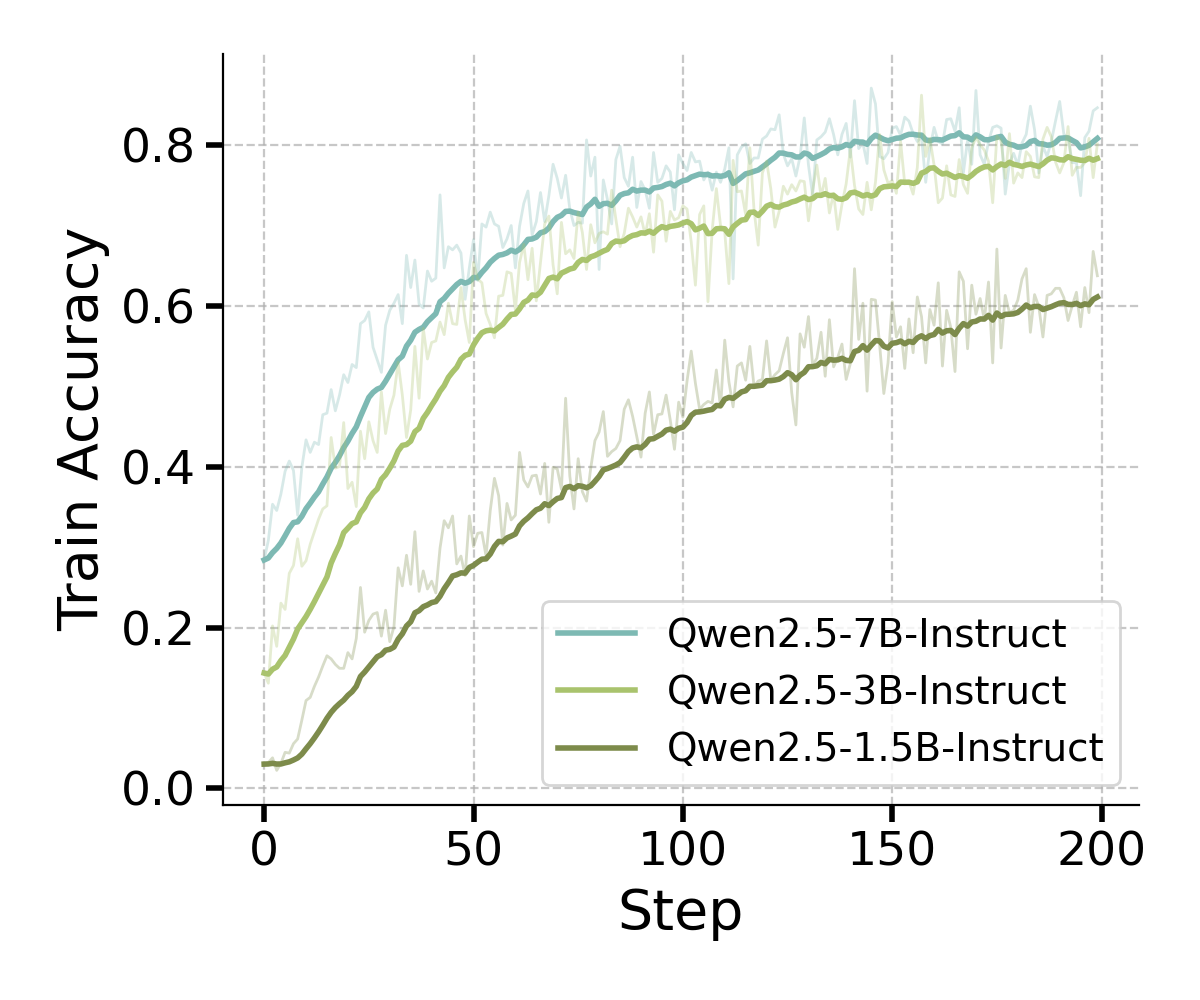}
        \vspace{-20pt}
        \caption{LIMR}
        \label{fig: training-limr-succ-rate}
    \end{subfigure}
    \hfill
    \begin{subfigure}[t]{0.32\textwidth}
        \centering
        \includegraphics[width=\linewidth]{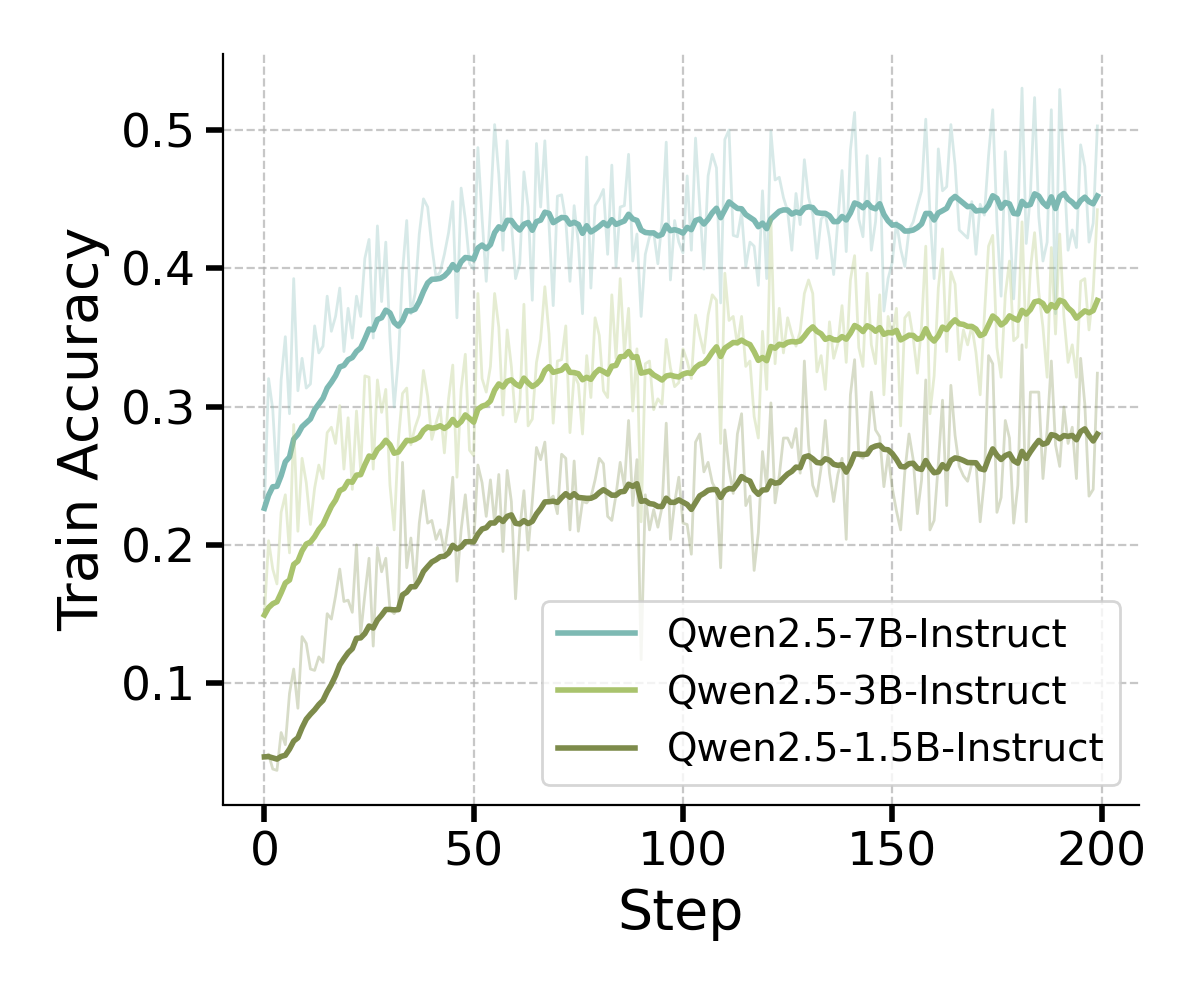}
        \vspace{-20pt}
        \caption{DeepScaleR}
        \label{fig: training-deepscaler-succ-rate}
    \end{subfigure} 
    \vspace{-6pt}
    \caption{{Training accuracy curves of experiments in Table~\ref{tab:training_results}.}}
    \label{fig: training accuracy curves}
\vspace{-8pt}
\end{figure}

\begin{figure}[t!]
    \centering
    \begin{subfigure}[t]{0.32\textwidth}
        \centering
        \includegraphics[width=\linewidth]{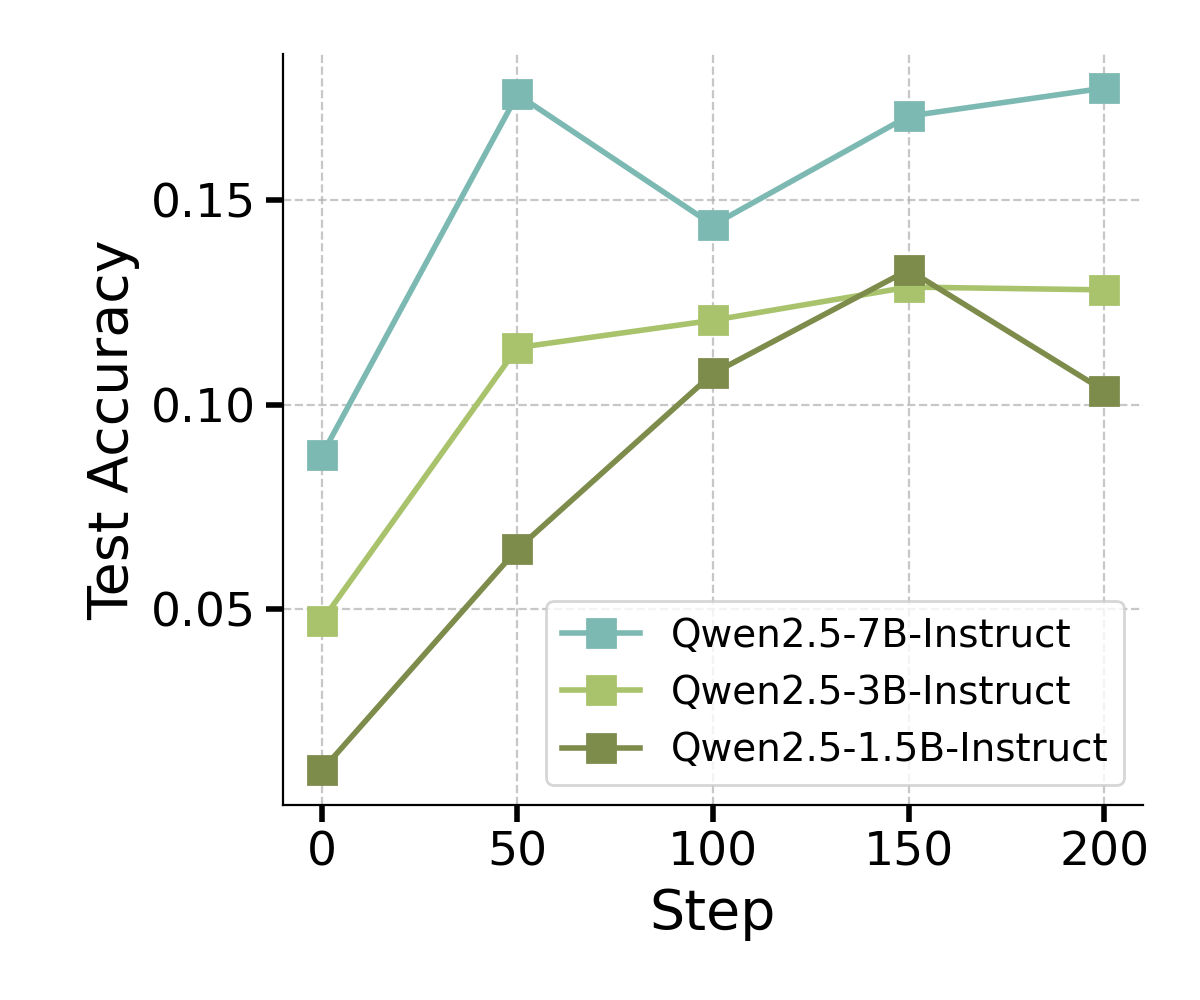}
        \vspace{-20pt}
        \caption{MATH-LightEval}
        \label{fig: val-lightevalmath-succ-rate}
    \end{subfigure}
    \hfill
    \begin{subfigure}[t]{0.32\textwidth}
        \centering
        \includegraphics[width=\linewidth]{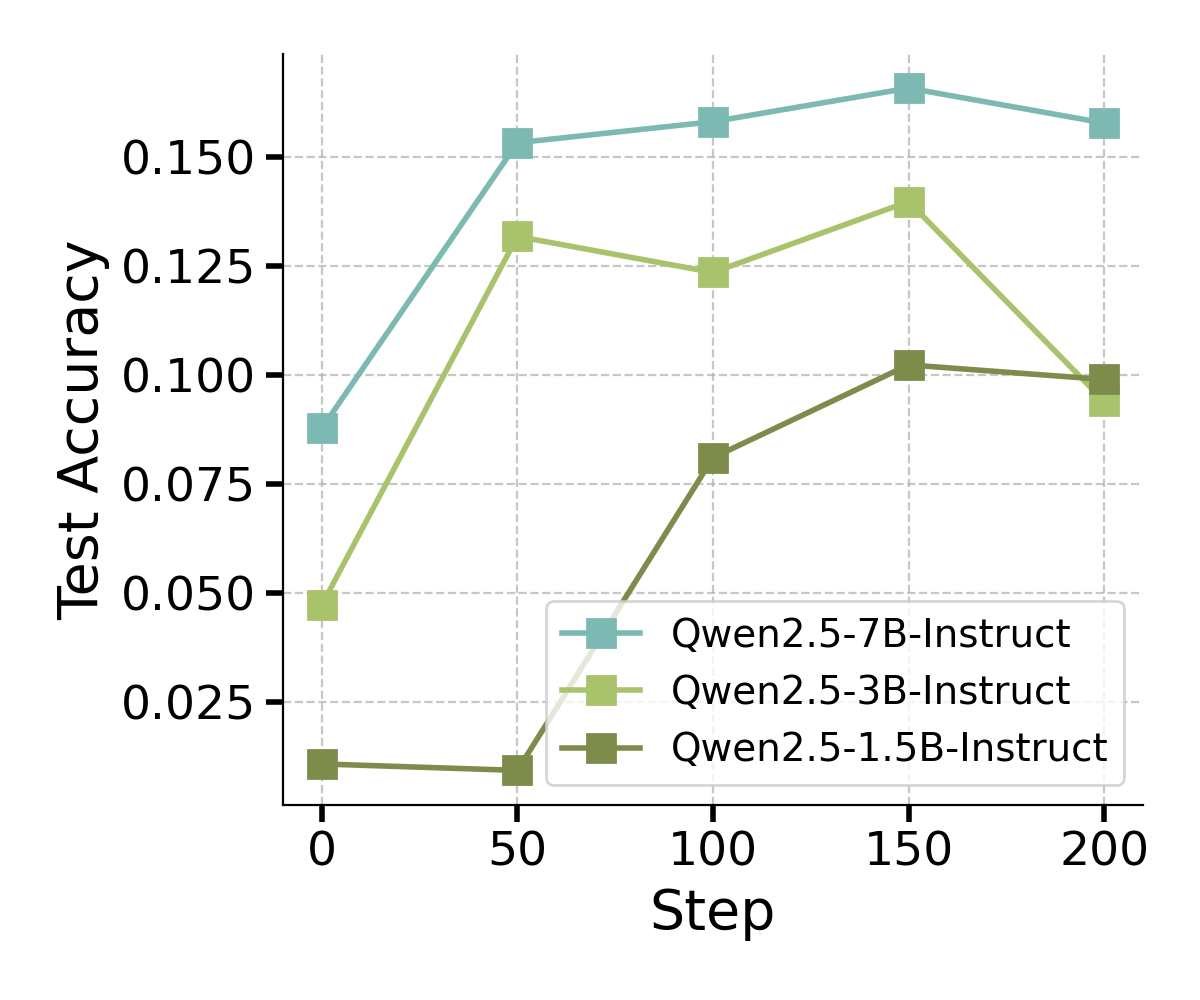}
        \vspace{-20pt}
        \caption{LIMR}
        \label{fig: val-limr-succ-rate}
    \end{subfigure}
    \hfill
    \begin{subfigure}[t]{0.32\textwidth}
        \centering
        \includegraphics[width=\linewidth]{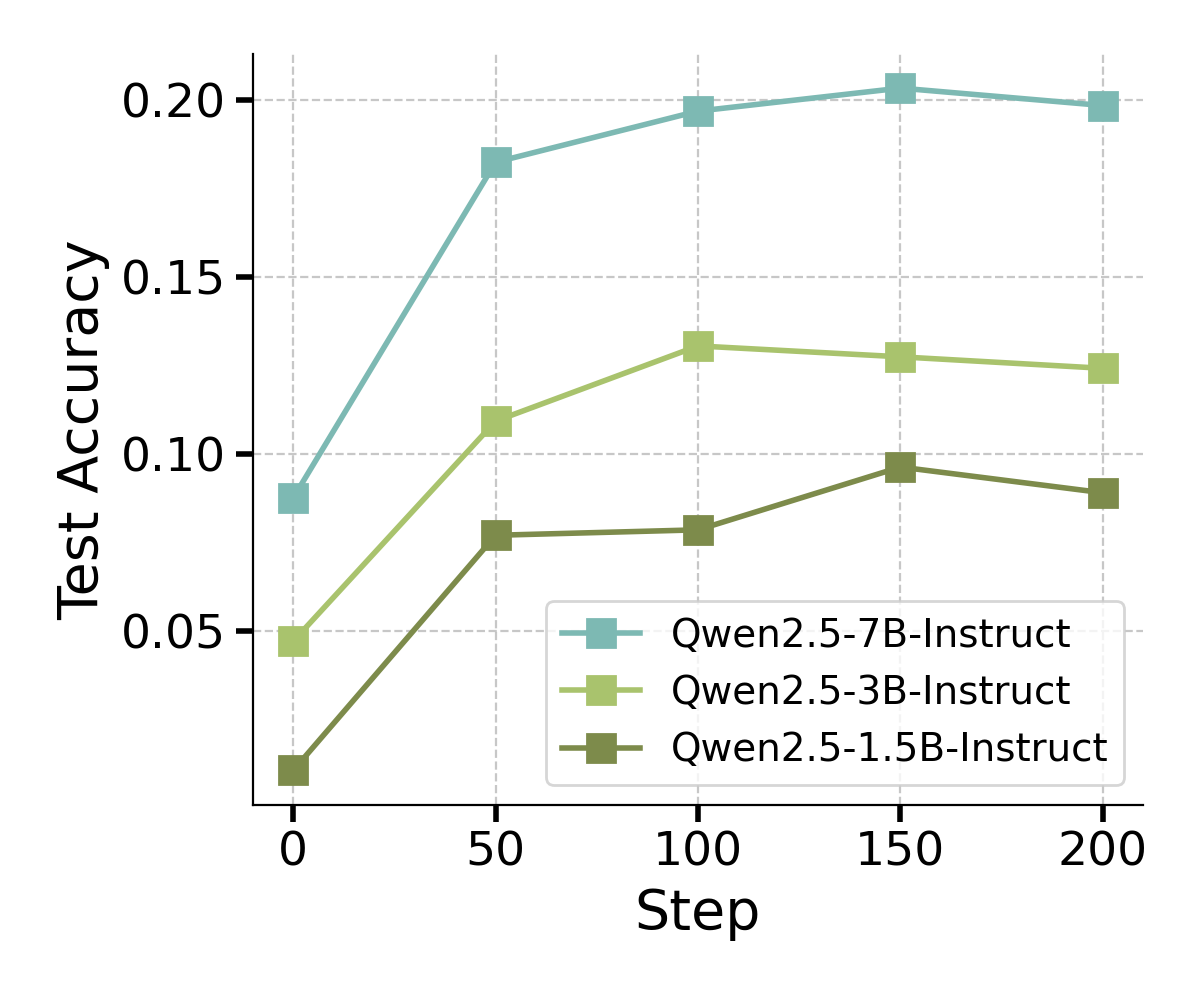}
        \vspace{-20pt}
        \caption{DeepScaleR}
        \label{fig: val-deepscaler-succ-rate}
    \end{subfigure} 
    \vspace{-6pt}
    \caption{Validation accuracy curves of experiments in Table~\ref{tab:training_results}.}
    \label{fig: validation accuracy curves}
\vspace{-16pt}
\end{figure}

\begin{figure}[t!]
\centering
\hfill
\begin{subfigure}[t]{0.32\textwidth}
\includegraphics[width=\linewidth]{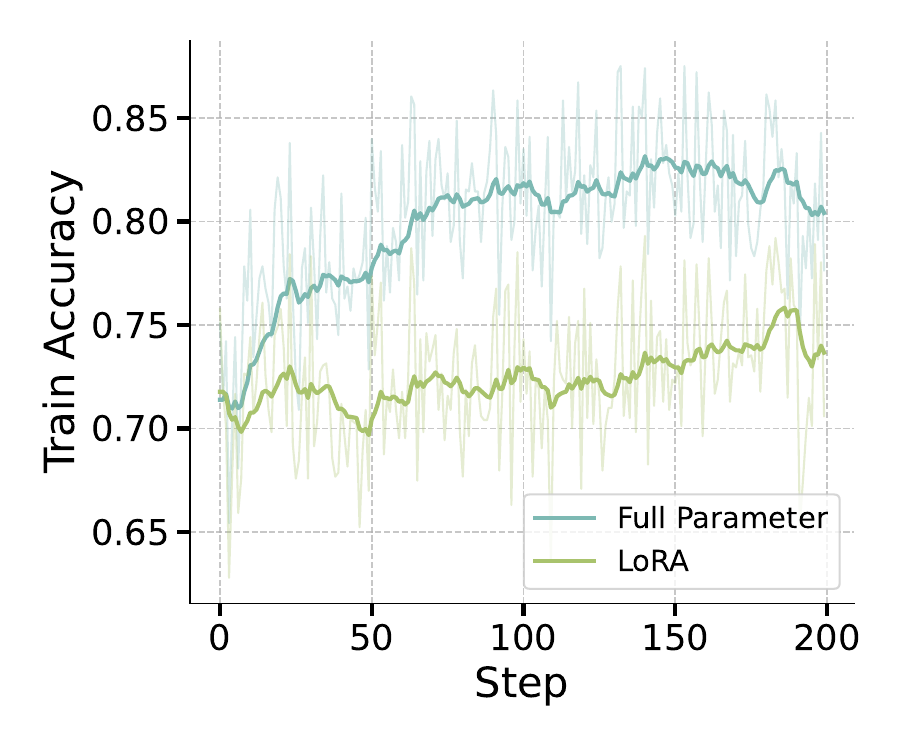}
\end{subfigure}%
\hfill
\begin{subfigure}[t]{0.32\textwidth}
\includegraphics[width=\linewidth]{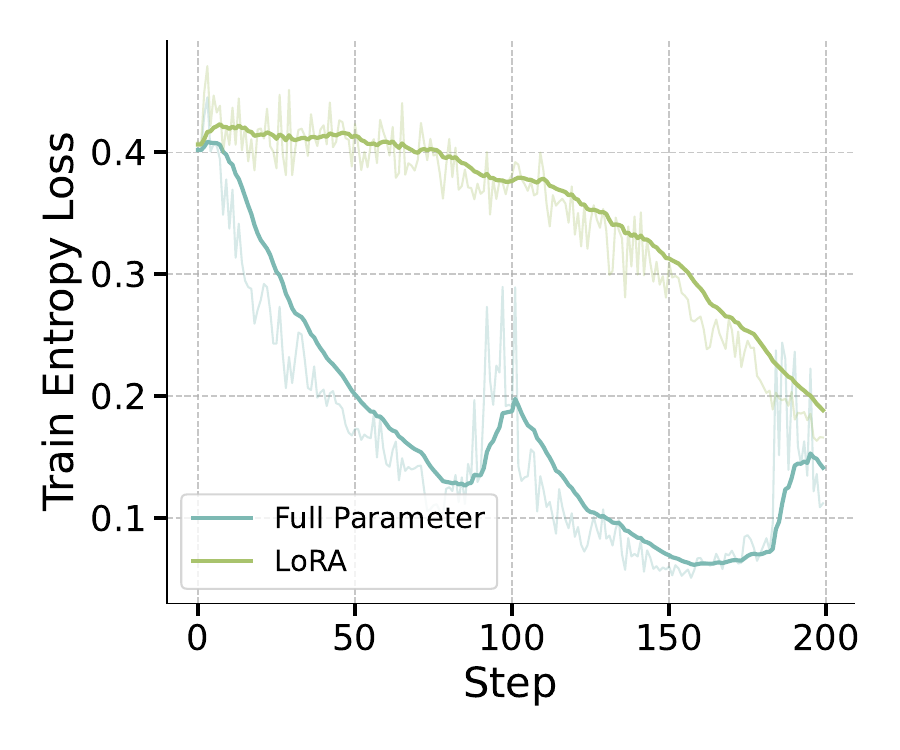}
\end{subfigure}
\hfill
\begin{subfigure}[t]{0.32\textwidth}
\includegraphics[width=\linewidth]{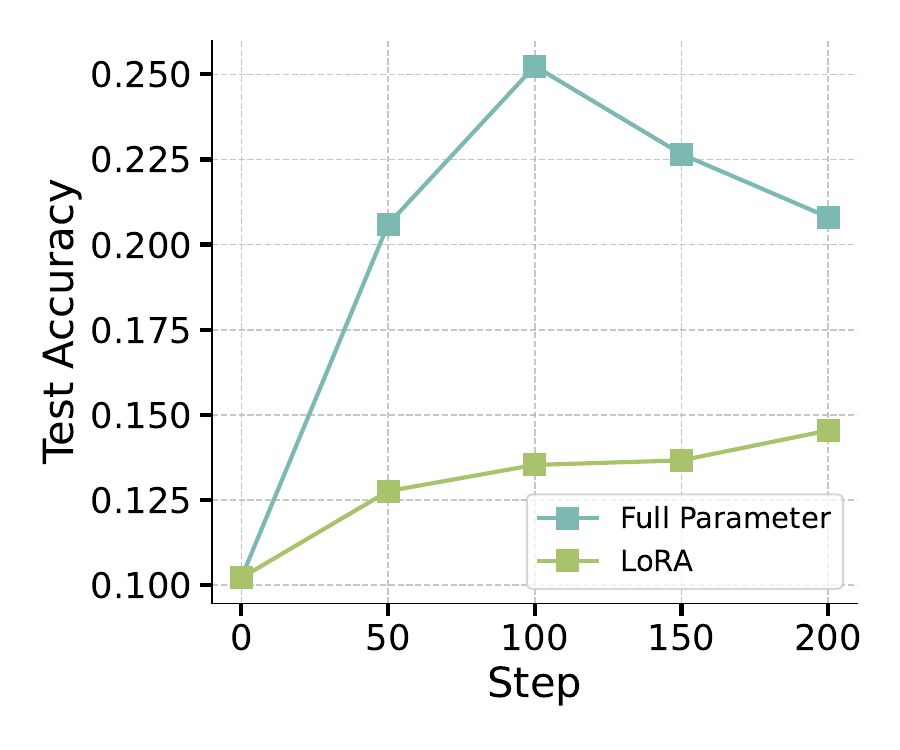}
\end{subfigure}
    \vspace{-16pt}
\caption{Comparison of training and validation in performance and entropy: Full parameters vs. LoRA on Qwen2.5-14B-Instruct with MATH-LightEval.}
\label{fig:full-para-vs-lora}
\vspace{-8pt}
\end{figure}

\paragraph{Agentic learning in AlphaApollo substantially improves reasoning across model scales.}
As shown in Table~\ref{tab:training_results}, training on DeepScaleR consistently yields large gains over tool-augmented baselines without training, improving average accuracy by +8.57 on Qwen2.5-1.5B-Instruct, +8.33 on Qwen2.5-3B-Instruct (MATH-LightEval, from 4.72\% to 13.35\%), and +11.58 on Qwen2.5-7B-Instruct, demonstrating strong scalability of agentic reasoning improvements. Figures~\ref{fig: training accuracy curves} and \ref{fig: validation accuracy curves} further show steady performance growth on both training and test sets. We additionally compare full-parameter and LoRA training (Figure~\ref{fig:full-para-vs-lora}). Full-parameter optimization exhibits faster learning dynamics, with quicker accuracy gains, reduced entropy loss, and consistently higher final performance, indicating the advantage of full-capacity adaptation for agentic reasoning.

\begin{figure}[t!]
    \centering
    \begin{subfigure}[t]{0.32\textwidth}
        \centering
        \includegraphics[width=\linewidth]{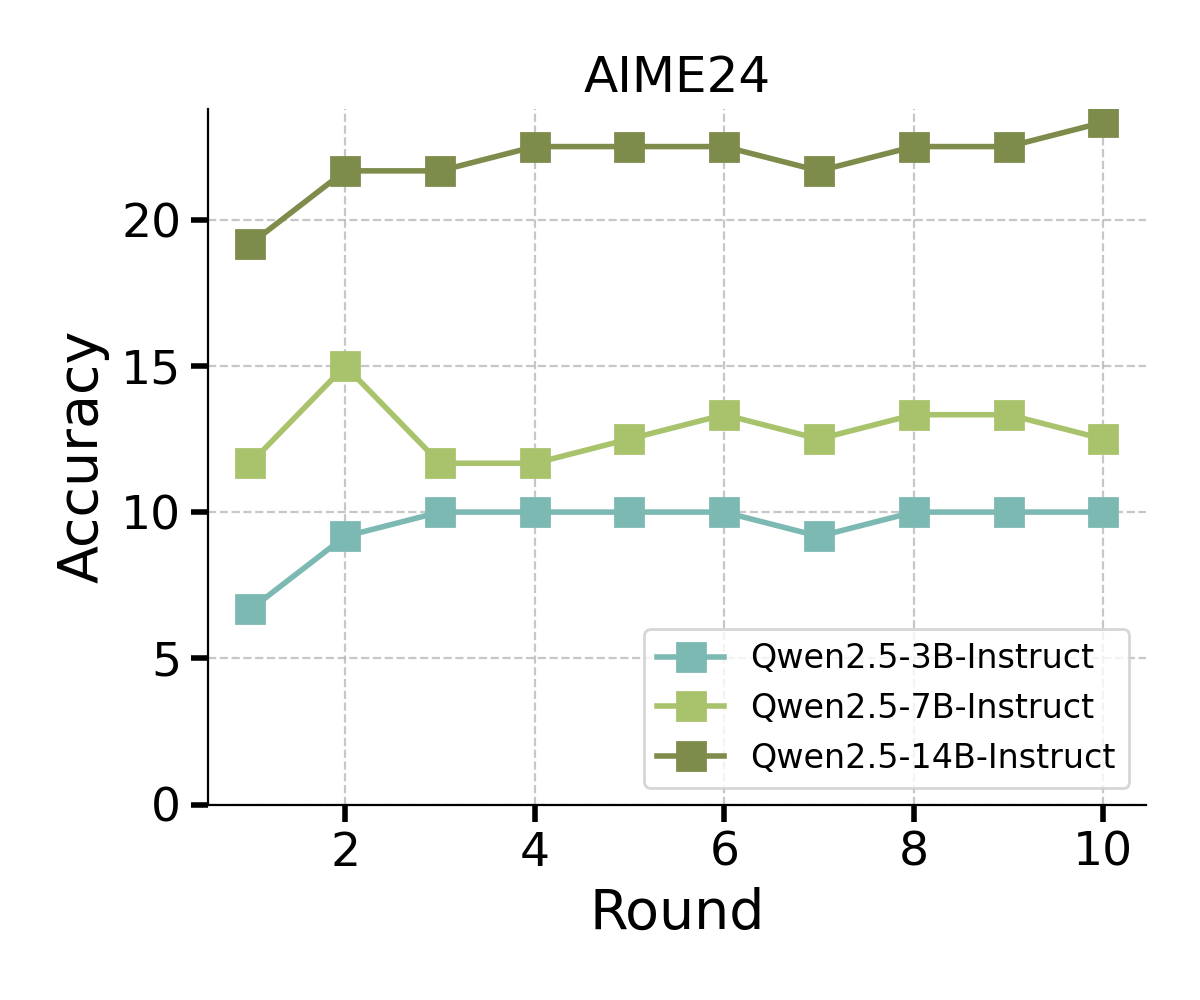}
        \label{fig: tool-call-correctness-4b}
    \end{subfigure}
    \hfill
    \begin{subfigure}[t]{0.32\textwidth}
        \centering
        \includegraphics[width=\linewidth]{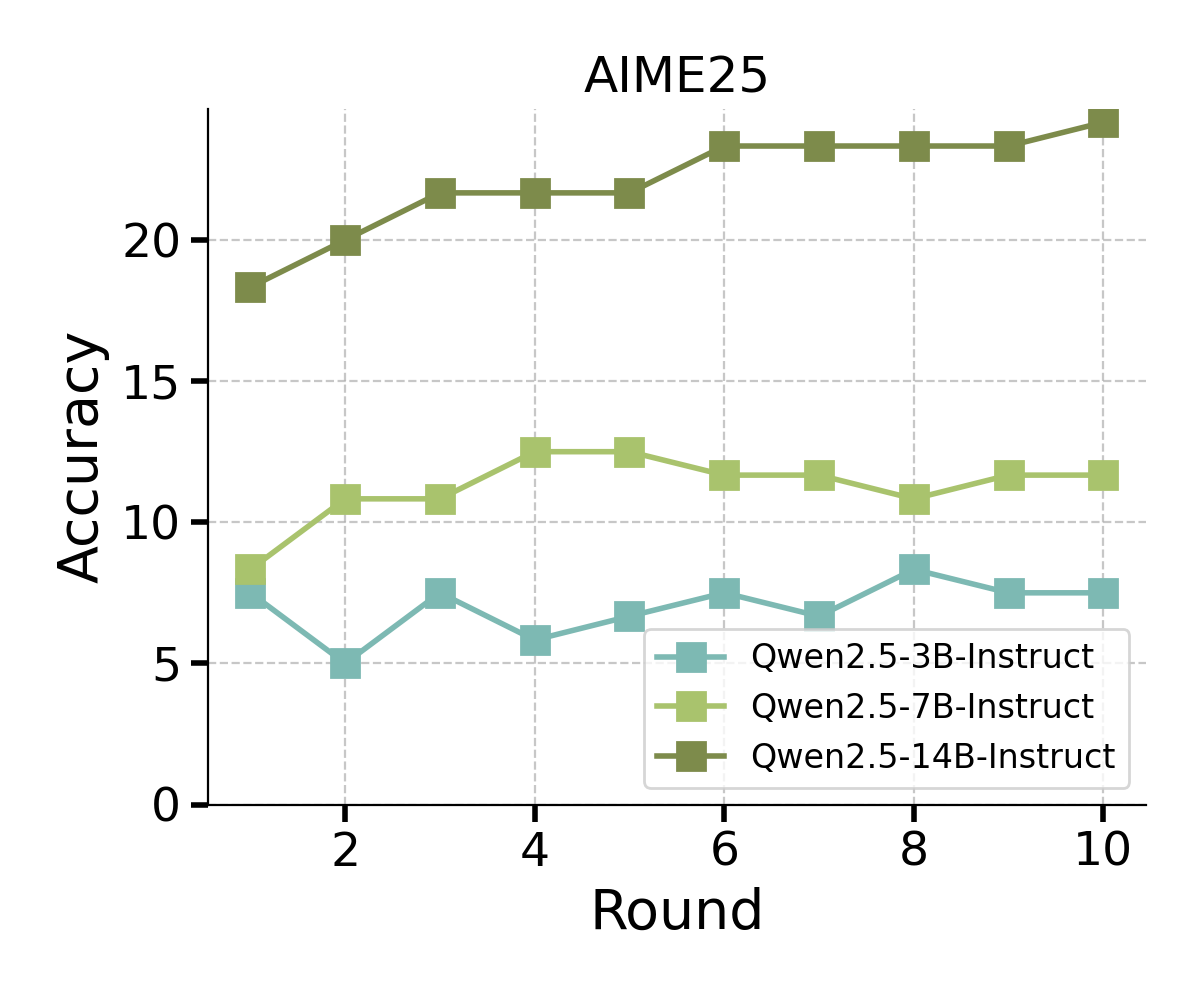}
        \label{fig: tool-call-correctness-30b}
    \end{subfigure}
    \hfill
    \begin{subfigure}[t]{0.32\textwidth}
        \centering
        \includegraphics[width=\linewidth]{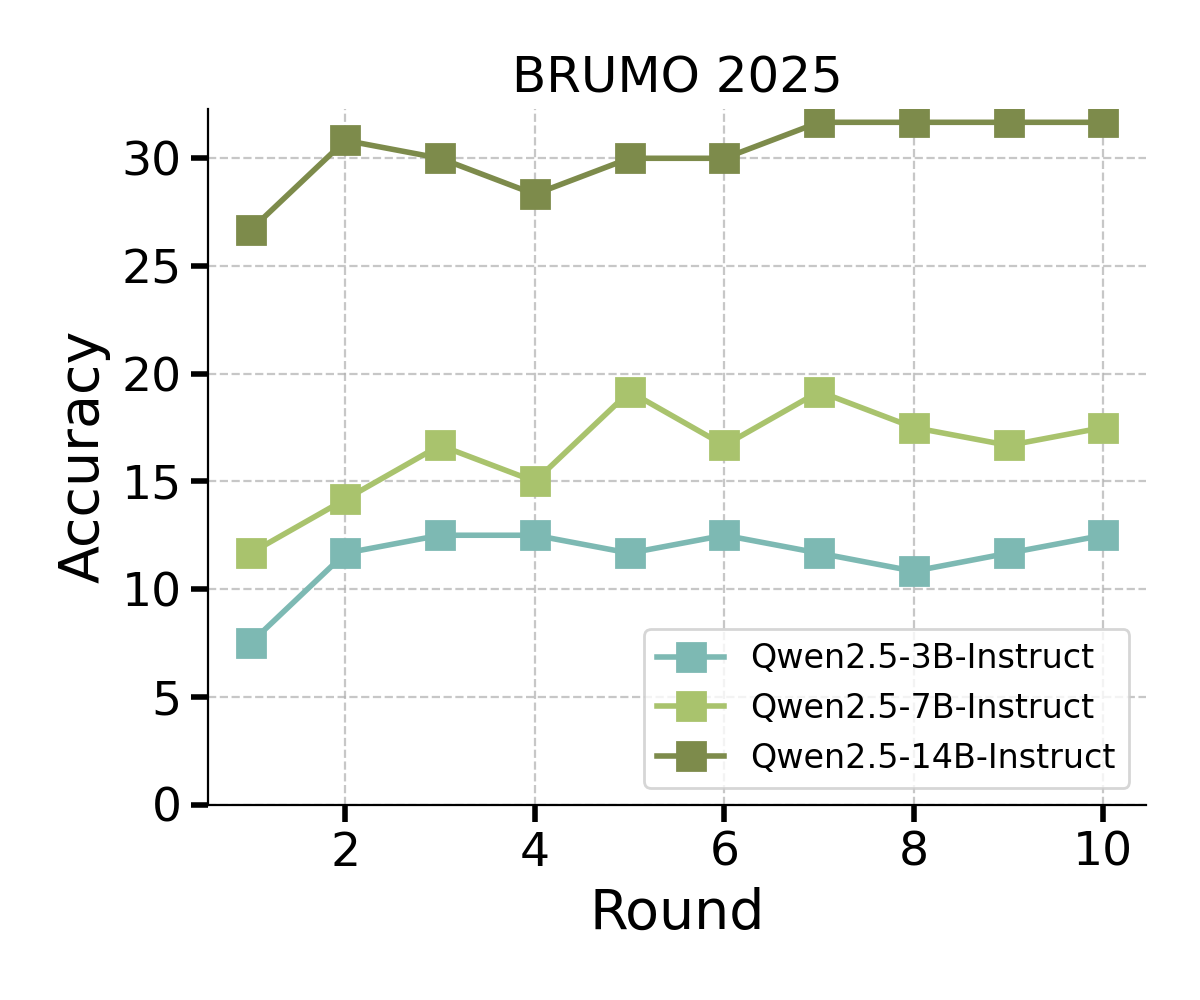}
        \label{fig: tool-call-correctness-branches}
    \end{subfigure}
    \vspace{-16pt}
    \caption{Accuracy across successive self-evolution rounds for different model scales. Performance consistently improves with more evolution rounds, demonstrating the effectiveness and stability of the self-evolving mechanism.}
    \label{fig: agentic-evolving}
    \vspace{-8pt}
\end{figure}

\paragraph{Agentic evolution further amplifies reasoning capability through iterative self-improvement.}
In Table~\ref{tab:evolving_results}, enabling evolution (+Evo) improves over the non-evolving baseline (w/o Evo). The strongest gains occur on Qwen2.5-14B-Instruct, where average accuracy increases from 16.53\% to 21.08\% (+4.55), including large boosts on AIME24 and AIME25. Improvements also persist on other backbones, with average gains of +2.43 on Qwen2.5-3B-Instruct. 
Figure~\ref{fig: agentic-evolving} reveals a stable upward trend across evolving rounds with clear scale-dependent gaps, indicating that agentic evolution reliably enhances reasoning performance and scales the benefit with model capability.

\noindent
Overall, these results show that AlphaApollo delivers consistent, scalable gains by unifying agentic reasoning, learning, and evolution, enabling reliable tool use and iterative refinement of reasoning.

\begin{figure}[t!]
\begin{minipage}[t]{0.50\textwidth}
\begin{tcolorbox}[colback=cyan!5,colframe=cyan!40!black,title=Case study: Decomposition]
\scriptsize 
\ldots, we need to \textbf{break it down into several steps}: \ldots

Let's start by calculating \ldots
\begin{tcolorbox}[colback=cyan!5,colframe=cyan!60!black!60,boxrule=0.3pt,left=2pt,right=2pt,top=1pt,bottom=1pt,title=Python Interpreter Calling]
\scriptsize 
from math import comb \\
(\ldots)
\end{tcolorbox}
The result of the code execution is: 210 \\

Next, let's calculate \ldots 
\begin{tcolorbox}[colback=cyan!5,colframe=cyan!60!black!60,boxrule=0.3pt,left=2pt,right=2pt,top=1pt,bottom=1pt,title=Python Interpreter Calling]
\scriptsize 
from math import comb \\
(\ldots)
\end{tcolorbox}

The result of the code execution is: 115, 90, 24, 1 \\
\ldots

Thus, the final answer is: $\boxed{116}$.
\end{tcolorbox}
\end{minipage}
\begin{minipage}[t]{0.50\textwidth}
\begin{tcolorbox}[colback=OliveGreen!5,colframe=OliveGreen!70!Black!70,title=Case study: Correction]
\scriptsize 
\ldots Let's write a Python script \ldots
\begin{tcolorbox}[colback=OliveGreen!5,colframe=OliveGreen!50!Black!40,boxrule=0.3pt,left=2pt,right=2pt,top=1pt,bottom=1pt,title=Python Interpreter Calling]
def find\_largest\_N():\\
(\ldots)
\end{tcolorbox}

The result of the code execution is: None

It seems there was an issue \ldots \textbf{Let's re-evaluate \ldots}

\ldots

We will re-run the script with a more detailed check \ldots
\begin{tcolorbox}[colback=OliveGreen!5,colframe=OliveGreen!50!Black!40,boxrule=0.3pt,left=2pt,right=2pt,top=1pt,bottom=1pt,title=Python Interpreter Calling]
\scriptsize 
def find\_largest\_N():\\
(\ldots)
\end{tcolorbox}

The result of the code execution is: 5694

\ldots

Thus, the final answer is: $\boxed{699}$.
\end{tcolorbox}
\end{minipage}

\begin{minipage}[t]{0.50\textwidth}
\begin{tcolorbox}[colback=RoyalPurple!5,colframe=RoyalPurple!70,title=Case study: Verification]
\scriptsize 
\ldots Now, we need to compute \(xy\):
\[ xy = \left( \frac{10}{a} \right) \left( \frac{5a}{2} \right) = \frac{50a}{2a} = 25 \]
\textbf{Let’s verify the solution \ldots}
\begin{tcolorbox}[colback=RoyalPurple!5,colframe=RoyalPurple!40,boxrule=0.3pt,left=2pt,right=2pt,top=1pt,bottom=1pt,title=Python Interpreter Calling]
\scriptsize 
from sympy import symbols, Eq, solve, log \\
x, y, a = symbols('x y a') \\
eq1 = Eq(x*a, 10) \\
eq2 = Eq(4*y/a, 10) \\
sol = solve((eq1, eq2), (x, y)) \\
xy = sol[x] * sol[y] \\
xy.simplify()
\end{tcolorbox}

The result of the code execution is: 25

The value of $xy$ is $\boxed{25}$.
\end{tcolorbox}
\end{minipage}
\begin{minipage}[t]{0.50\textwidth}
\begin{tcolorbox}[colback=Salmon!5,colframe=Salmon!90!Black,title=Case study: Backtracking]
\scriptsize 
\ldots 
Let's implement the above steps \ldots
\begin{tcolorbox}[colback=Salmon!5,colframe=Salmon!70!,boxrule=0.3pt,left=2pt,right=2pt,top=1pt,bottom=1pt,title=Python Interpreter Calling]
\scriptsize 
from sympy import * \\
(\ldots)
\end{tcolorbox}
TypeError: cannot determine truth value of Relational.

It appears that the direct symbolic solution approach using SymPy did not work as expected due to the complexity of the equations. 

\textbf{Let’s try a different approach \ldots} 
\begin{tcolorbox}[colback=Salmon!5,colframe=Salmon!70!,boxrule=0.3pt,left=2pt,right=2pt,top=1pt,bottom=1pt,title=Python Interpreter Calling]
\scriptsize 
\# Iterate over possible values of a and b\\
(\ldots)
\end{tcolorbox}
The result of the code execution is: 601

Thus, the final answer is: $\boxed{601}$.
\end{tcolorbox}
\end{minipage}
\vspace{-16pt}
\caption{Case studies of models' cognitive behaviors exhibited by AlphaApollo. 
}
\label{fig:case_studies}
\end{figure}

\paragraph{Case studies.}
Under AlphaApollo, models exhibit diverse cognitive reasoning behaviors, rooted in their pre-training data and enabling deliberate reasoning.
We showcase in Figure~\ref{fig:case_studies} and
present complete case studies in Appendix~\ref{app: case cognitive behaviors} while summarizing following agentic reasoning behaviors:

\vspace{-4pt}
\begin{itemize}[leftmargin=*, noitemsep]
\setlength\itemsep{2pt}
\vspace{-6pt}
\item \textbf{Decomposition}. The model breaks down a complex problem into smaller, more manageable sub-problems. This strategy not only reduces cognitive load but also increases the likelihood of solving each component correctly, which in turn contributes to the accuracy of the final solution.

\item \textbf{Correction}. During the reasoning process, the model frequently identifies potential mistakes in intermediate steps and revises them. Such self-corrective behavior shows that the model can refine its outputs dynamically rather than strictly following an error-prone initial trajectory.

\item \textbf{Verification}. The model actively checks intermediate results against either external tools or internal consistency rules. This verification step functions as a safeguard, filtering out unreasonable solutions and ensuring that the final answer is logically sound.

\item \textbf{Backtracking}. When encountering contradictions, the model is capable of retracing earlier steps and exploring alternative reasoning paths.
This behavior resembles human-like problem solving, where a failed attempt triggers a systematic search for better strategies.
\vspace{-4pt}
\end{itemize}

\section{Conclusion}


We presented AlphaApollo, a self-evolving agentic reasoning system that addresses limited long-horizon reasoning capacity and unreliable test-time refinement by orchestrating models and tools. 
AlphaApollo combines multi-turn agentic reasoning, turn-level agentic learning, and multi-round agentic evolution to enable structured tool use, stable RL optimization, and iterative self-improvement with tool-assisted verifications and long-horizon memory. 
Across seven math reasoning benchmarks and multiple model scales, AlphaApollo delivers consistent gains through reliable tool use, multi-turn reinforcement learning, and further improvements from test-time evolution.
\section*{Acknowledgement}

We want to thank Guanyu Jiang for his help in testing the early version of AlphaApollo and Yicheng Zhou for his help in testing the evolution part of AlphaApollo.

\bibliography{ref}

@article{qi2025learning,
  title={Learning to Reason Across Parallel Samples for LLM Reasoning},
  author={Qi, Jianing and Ye, Xi and Tang, Hao and Zhu, Zhigang and Choi, Eunsol},
  journal={arXiv preprint arXiv:2506.09014},
  year={2025}
}

@inproceedings{
yoran2023answering,
title={Answering Questions by Meta-Reasoning over Multiple Chains of Thought},
author={Ori Yoran and Tomer Wolfson and Ben Bogin and Uri Katz and Daniel Deutch and Jonathan Berant},
booktitle={EMNLP},
year={2023},
}

@article{naik2023diversity,
  title={Diversity of thought improves reasoning abilities of LLMs},
  author={Naik, Ranjita and Chandrasekaran, Varun and Yuksekgonul, Mert and Palangi, Hamid and Nushi, Besmira},
  journal={arXiv preprint arXiv:2310.07088},
  year={2023}
}

@article{taubenfeld2025confidence,
  title={Confidence improves self-consistency in llms},
  author={Taubenfeld, Amir and Sheffer, Tom and Ofek, Eran and Feder, Amir and Goldstein, Ariel and Gekhman, Zorik and Yona, Gal},
  journal={arXiv preprint arXiv:2502.06233},
  year={2025}
}

@article{knappe2024semantic,
  title={Semantic Self-Consistency: Enhancing Language Model Reasoning via Semantic Weighting},
  author={Knappe, Tim and Li, Ryan and Chauhan, Ayush and Chhua, Kaylee and Zhu, Kevin and O'Brien, Sean},
  journal={arXiv preprint arXiv:2410.07839},
  year={2024}
}

@article{pan2025octotools,
  title={OctoTools: An Agentic Framework with Extensible Tools for Complex Reasoning}, 
  author={Pan Lu and Bowen Chen and Sheng Liu and Rahul Thapa and Joseph Boen and James Zou},
  journal={arXiv preprint arXiv:2502.11271},
  year={2025} 
}

@article{schulman2017proximal,
  title={Proximal policy optimization algorithms},
  author={Schulman, John and Wolski, Filip and Dhariwal, Prafulla and Radford, Alec and Klimov, Oleg},
  journal={arXiv preprint arXiv:1707.06347},
  year={2017}
}

@article{shao2024deepseekmath,
  title={Deepseekmath: Pushing the limits of mathematical reasoning in open language models},
  author={Shao, Zhihong and Wang, Peiyi and Zhu, Qihao and Xu, Runxin and Song, Junxiao and Bi, Xiao and Zhang, Haowei and Zhang, Mingchuan and Li, YK and Wu, Y and others},
  journal={arXiv preprint arXiv:2402.03300},
  year={2024}
}

@article{yang2024qwen2,
  title={Qwen2. 5 technical report},
  author={Yang, An and Yang, Baosong and Zhang, Beichen and Hui, Binyuan and Zheng, Bo and Yu, Bowen and Li, Chengyuan and Liu, Dayiheng and Huang, Fei and Wei, Haoran and others},
  journal={arXiv preprint arXiv:2412.15115},
  year={2024}
}

@article{yang2025qwen3,
  title={Qwen3 Technical Report}, 
  author={An Yang and Anfeng Li and Baosong Yang and Beichen Zhang and Binyuan Hui and Bo Zheng and Bowen Yu and Chang Gao and Chengen Huang and Chenxu Lv and Chujie Zheng and Dayiheng Liu and Fan Zhou and Fei Huang  and others},
  journal={arXiv preprint arXiv:2505.09388},
  year={2025} 
}

@inproceedings{hendrycksmath2021,
  title={Measuring Mathematical Problem Solving With the MATH Dataset},
  author={Dan Hendrycks and Collin Burns and Saurav Kadavath and Akul Arora and Steven Basart and Eric Tang and Dawn Song and Jacob Steinhardt},
  booktitle={NeurIPS},
  year={2021}
}

@article{yu2025dapo,
  title={Dapo: An open-source llm reinforcement learning system at scale},
  author={Yu, Qiying and Zhang, Zheng and Zhu, Ruofei and Yuan, Yufeng and Zuo, Xiaochen and Yue, Yu and Fan, Tiantian and Liu, Gaohong and Liu, Lingjun and Liu, Xin and others},
  journal={arXiv preprint arXiv:2503.14476},
  year={2025}
}

@article{snell2024scaling,
  title={Scaling llm test-time compute optimally can be more effective than scaling model parameters},
  author={Snell, Charlie and Lee, Jaehoon and Xu, Kelvin and Kumar, Aviral},
  journal={arXiv preprint arXiv:2408.03314},
  year={2024}
}

@article{muennighoff2025s1,
  title={s1: Simple test-time scaling},
  author={Muennighoff, Niklas and Yang, Zitong and Shi, Weijia and Li, Xiang Lisa and Fei-Fei, Li and Hajishirzi, Hannaneh and Zettlemoyer, Luke and Liang, Percy and Cand{\`e}s, Emmanuel and Hashimoto, Tatsunori},
  journal={arXiv preprint arXiv:2501.19393},
  year={2025}
}

@article{phan2025humanity,
  title={Humanity's last exam},
  author={Phan, Long and Gatti, Alice and Han, Ziwen and Li, Nathaniel and Hu, Josephina and Zhang, Hugh and Zhang, Chen Bo Calvin and Shaaban, Mohamed and Ling, John and Shi, Sean and others},
  journal={arXiv preprint arXiv:2501.14249},
  year={2025}
}

@article{chollet2025arc,
  title={Arc-agi-2: A new challenge for frontier ai reasoning systems},
  author={Chollet, Francois and Knoop, Mike and Kamradt, Gregory and Landers, Bryan and Pinkard, Henry},
  journal={arXiv preprint arXiv:2505.11831},
  year={2025}
}

@article{meurer2017sympy,
  title={SymPy: symbolic computing in Python},
  author={Meurer, Aaron and Smith, Christopher P and Paprocki, Mateusz and {\v{C}}ert{\'\i}k, Ond{\v{r}}ej and Kirpichev, Sergey B and Rocklin, Matthew and Kumar, AMiT and Ivanov, Sergiu and Moore, Jason K and Singh, Sartaj and others},
  journal={PeerJ Computer Science},
  year={2017}
}

@inproceedings{lewis2020retrieval,
  title={Retrieval-augmented generation for knowledge-intensive nlp tasks},
  author={Lewis, Patrick and Perez, Ethan and Piktus, Aleksandra and Petroni, Fabio and Karpukhin, Vladimir and Goyal, Naman and K{\"u}ttler, Heinrich and Lewis, Mike and Yih, Wen-tau and Rockt{\"a}schel, Tim and others},
  booktitle={NeurIPS},
  year={2020}
}

@article{wang2025hierarchical,
  title={Hierarchical Reasoning Model},
  author={Wang, Guan and Li, Jin and Sun, Yuhao and Chen, Xing and Liu, Changling and Wu, Yue and Lu, Meng and Song, Sen and Yadkori, Yasin Abbasi},
  journal={arXiv preprint arXiv:2506.21734},
  year={2025}
}

@article{yang2024number,
  title={Number cookbook: Number understanding of language models and how to improve it},
  author={Yang, Haotong and Hu, Yi and Kang, Shijia and Lin, Zhouchen and Zhang, Muhan},
  journal={arXiv preprint arXiv:2411.03766},
  year={2024}
}

@inproceedings{wang2022self,
  title={Self-consistency improves chain of thought reasoning in language models},
  author={Wang, Xuezhi and Wei, Jason and Schuurmans, Dale and Le, Quoc and Chi, Ed and Narang, Sharan and Chowdhery, Aakanksha and Zhou, Denny},
  booktitle={ICLR},
  year={2022}
}

@article{gao2025survey,
  title={A survey of self-evolving agents: On path to artificial super intelligence},
  author={Gao, Huan-ang and Geng, Jiayi and Hua, Wenyue and Hu, Mengkang and Juan, Xinzhe and Liu, Hongzhang and Liu, Shilong and Qiu, Jiahao and Qi, Xuan and Wu, Yiran and others},
  journal={arXiv preprint arXiv:2507.21046},
  year={2025}
}

@inproceedings{madaan2023self,
  title={Self-refine: Iterative refinement with self-feedback},
  author={Madaan, Aman and Tandon, Niket and Gupta, Prakhar and Hallinan, Skyler and Gao, Luyu and Wiegreffe, Sarah and Alon, Uri and Dziri, Nouha and Prabhumoye, Shrimai and Yang, Yiming and others},
  booktitle={NeurIPS},
  year={2023}
}

@inproceedings{
shinn2023reflexion,
title={Reflexion: language agents with verbal reinforcement learning},
author={Noah Shinn and Federico Cassano and Ashwin Gopinath and Karthik R Narasimhan and Shunyu Yao},
booktitle={NeurIPS},
year={2023},
}

@article{yuksekgonul2025optimizing,
  title={Optimizing generative AI by backpropagating language model feedback},
  author={Yuksekgonul, Mert and Bianchi, Federico and Boen, Joseph and Liu, Sheng and Lu, Pan and Huang, Zhi and Guestrin, Carlos and Zou, James},
  journal={Nature},
  year={2025}
}

@inproceedings{kim2023language,
  title={Language models can solve computer tasks},
  author={Kim, Geunwoo and Baldi, Pierre and McAleer, Stephen},
  booktitle={NeurIPS},
  year={2023}
}

@inproceedings{
huang2024large,
title={Large Language Models Cannot Self-Correct Reasoning Yet},
author={Jie Huang and Xinyun Chen and Swaroop Mishra and Huaixiu Steven Zheng and Adams Wei Yu and Xinying Song and Denny Zhou},
booktitle={ICLR},
year={2024}
}

@article{paul2023refiner,
  title={Refiner: Reasoning feedback on intermediate representations},
  author={Paul, Debjit and Ismayilzada, Mete and Peyrard, Maxime and Borges, Beatriz and Bosselut, Antoine and West, Robert and Faltings, Boi},
  journal={arXiv preprint arXiv:2304.01904},
  year={2023}
}

@article{novikov2025alphaevolve,
  title={AlphaEvolve: A coding agent for scientific and algorithmic discovery},
  author={Novikov, Alexander and V{\~u}, Ng{\^a}n and Eisenberger, Marvin and Dupont, Emilien and Huang, Po-Sen and Wagner, Adam Zsolt and Shirobokov, Sergey and Kozlovskii, Borislav and Ruiz, Francisco JR and Mehrabian, Abbas and others},
  journal={arXiv preprint arXiv:2506.13131},
  year={2025}
}

@inproceedings{
yang2024large,
title={Large Language Models as Optimizers},
author={Chengrun Yang and Xuezhi Wang and Yifeng Lu and Hanxiao Liu and Quoc V Le and Denny Zhou and Xinyun Chen},
booktitle={ICLR},
year={2024}
}

@inproceedings{qu2024recursive,
  title={Recursive introspection: Teaching language model agents how to self-improve},
  author={Qu, Yuxiao and Zhang, Tianjun and Garg, Naman and Kumar, Aviral},
  booktitle={NeurIPS},
  year={2024}
}

@article{chen2023teaching,
  title={Teaching large language models to self-debug},
  author={Chen, Xinyun and Lin, Maxwell and Sch{\"a}rli, Nathanael and Zhou, Denny},
  journal={arXiv preprint arXiv:2304.05128},
  year={2023}
}

@inproceedings{zheng2024sglang,
  title={Sglang: Efficient execution of structured language model programs},
  author={Zheng, Lianmin and Yin, Liangsheng and Xie, Zhiqiang and Sun, Chuyue Livia and Huang, Jeff and Yu, Cody Hao and Cao, Shiyi and Kozyrakis, Christos and Stoica, Ion and Gonzalez, Joseph E and others},
  booktitle={NeurIPS},
  year={2024}
}

@inproceedings{kwon2023efficient,
  title={Efficient memory management for large language model serving with pagedattention},
  author={Kwon, Woosuk and Li, Zhuohan and Zhuang, Siyuan and Sheng, Ying and Zheng, Lianmin and Yu, Cody Hao and Gonzalez, Joseph and Zhang, Hao and Stoica, Ion},
  booktitle={SOSP},
  year={2023}
}

@misc{openaiapi,
  title  = {OpenAI API Platform},
  author = {{OpenAI}},
  year   = {2025},
  url    = {https://platform.openai.com}
}

@misc{Huggingface,
  title  = {Huggingface Transformers},
  author = {{Huggingface}},
  year   = {2025},
  url    = {https://huggingface.co/docs/transformers}
}

@inproceedings{gao2024confucius,
  title={Confucius: Iterative tool learning from introspection feedback by easy-to-difficult curriculum},
  author={Gao, Shen and Shi, Zhengliang and Zhu, Minghang and Fang, Bowen and Xin, Xin and Ren, Pengjie and Chen, Zhumin and Ma, Jun and Ren, Zhaochun},
  booktitle={AAAI},
  year={2024}
}

@article{numpy,
 title         = {Array programming with {NumPy}},
 author        = {Charles R. Harris and K. Jarrod Millman and St{\'{e}}fan J.
                 van der Walt and Ralf Gommers and Pauli Virtanen and David
                 Cournapeau and Eric Wieser and Julian Taylor and Sebastian
                 Berg and Nathaniel J. Smith and Robert Kern and Matti Picus
                 and Stephan Hoyer and Marten H. van Kerkwijk and Matthew
                 Brett and Allan Haldane and Jaime Fern{\'{a}}ndez del
                 R{\'{i}}o and Mark Wiebe and Pearu Peterson and Pierre
                 G{\'{e}}rard-Marchant and Kevin Sheppard and Tyler Reddy and
                 Warren Weckesser and Hameer Abbasi and Christoph Gohlke and
                 Travis E. Oliphant},
 year          = {2020},
 journal       = {Nature},
}

@article{scipy,
  title={SciPy 1.0: Fundamental Algorithms for Scientific Computing in Python},
  author={Virtanen, Pauli and Gommers, Ralf and Oliphant, Travis E. and Haberland, Matt and Reddy, Tyler and Cournapeau, David and others},
  journal={Nature Methods},
  year={2020},
}

@article{chen2023program,
  title={Program of Thoughts Prompting: Disentangling Computation from Reasoning for Numerical Reasoning Tasks},
  author={Wenhu Chen and Xueguang Ma and Xinyi Wang and William W. Cohen},
  journal={Transactions on Machine Learning Research},
  year={2023}
}

@article{liu2024apigen,
  title={APIGen: Automated Pipeline for Generating Verifiable and Diverse Function-Calling Datasets},
  author={Zuxin Liu and Thai Hoang and Jianguo Zhang and Ming Zhu and Tian Lan and Shirley Kokane and Juntao Tan and Weiran Yao and Zhiwei Liu and Yihao Feng and Rithesh Murthy and Liangwei Yang and Silvio Savarese and Juan Carlos Niebles and Huan Wang and Shelby Heinecke and Caiming Xiong},
  journal={arXiv preprint arXiv:2406.18518},
  year={2024}
}

@inproceedings{shen2024small,
  title={Small LLMs Are Weak Tool Learners: A Multi-LLM Agent},
  author={Weizhou Shen and Chenliang Li and Hongzhan Chen and Ming Yan and Xiaojun Quan and Hehong Chen and Ji Zhang and Fei Huang},
  booktitle={EMNLP},
  year={2024}
}

@inproceedings{qin2024toolllm,
  title={ToolLLM: Facilitating Large Language Models to Master 16000+ Real-world APIs},
  author={Yujia Qin and Shihao Liang and Yining Ye and Kunlun Zhu and Lan Yan and Yaxi Lu and Yankai Lin and Xin Cong and Xiangru Tang and Bill Qian and Sihan Zhao and Lauren Hong and Runchu Tian and Ruobing Xie and Jie Zhou and Mark Gerstein and Dahai Li and Zhiyuan Liu and Maosong Sun},
  booktitle={ICLR},
  year={2024}
}

@article{chai2025scimaster,
  title={SciMaster: Towards General-Purpose Scientific AI Agents, Part I. X-Master as Foundation: Can We Lead on Humanity's Last Exam?},
  author={Jingyi Chai and Shuo Tang and Rui Ye and Yuwen Du and Xinyu Zhu and Mengcheng Zhou and Yanfeng Wang and Weinan E and Yuzhi Zhang and Linfeng Zhang and Siheng Chen},
  journal={arXiv preprint arXiv:2507.05241},
  year={2025}
}

@inproceedings{wu2024autogen,
  title={Autogen: Enabling next-gen LLM applications via multi-agent conversations},
  author={Wu, Qingyun and Bansal, Gagan and Zhang, Jieyu and Wu, Yiran and Li, Beibin and Zhu, Erkang and Jiang, Li and Zhang, Xiaoyun and Zhang, Shaokun and Liu, Jiale and others},
  booktitle={COLM},
  year={2024}
}

@inproceedings{hong2024metagpt,
  title={MetaGPT: Meta programming for a multi-agent collaborative framework},
  author={Hong, Sirui and Zhuge, Mingchen and Chen, Jonathan and Zheng, Xiawu and Cheng, Yuheng and Zhang, Ceyao and Wang, Jinlin and Wang, Zili and Yau, Steven Ka Shing and Lin, Zijuan and others},
  year={2024},
  booktitle={ICLR}
}

@inproceedings{shen2023hugginggpt,
  title={Hugginggpt: Solving ai tasks with chatgpt and its friends in hugging face},
  author={Shen, Yongliang and Song, Kaitao and Tan, Xu and Li, Dongsheng and Lu, Weiming and Zhuang, Yueting},
  booktitle={NeurIPS},
  year={2023}
}

@article{li2024improving,
  title={Improving multi-agent debate with sparse communication topology},
  author={Li, Yunxuan and Du, Yibing and Zhang, Jiageng and Hou, Le and Grabowski, Peter and Li, Yeqing and Ie, Eugene},
  journal={arXiv preprint arXiv:2406.11776},
  year={2024}
}

@article{agashe2023llm,
  title={Llm-coordination: evaluating and analyzing multi-agent coordination abilities in large language models},
  author={Agashe, Saaket and Fan, Yue and Reyna, Anthony and Wang, Xin Eric},
  journal={arXiv preprint arXiv:2310.03903},
  year={2023}
}

@article{zhou2025agentfly,
  title={AgentFly: Fine-tuning LLM Agents without Fine-tuning LLMs},
  author={Zhou, Huichi and Chen, Yihang and Guo, Siyuan and Yan, Xue and Lee, Kin Hei and Wang, Zihan and Lee, Ka Yiu and Zhang, Guchun and Shao, Kun and Yang, Linyi and others},
  journal={arXiv preprint arXiv:2508.16153},
  year={2025}
}

@article{qiu2025alita,
  title={Alita: Generalist agent enabling scalable agentic reasoning with minimal predefinition and maximal self-evolution},
  author={Qiu, Jiahao and Qi, Xuan and Zhang, Tongcheng and Juan, Xinzhe and Guo, Jiacheng and Lu, Yifu and Wang, Yimin and Yao, Zixin and Ren, Qihan and Jiang, Xun and others},
  journal={arXiv preprint arXiv:2505.20286},
  year={2025}
}

@article{jiang2025verltool,
  title={VerlTool: Towards Holistic Agentic Reinforcement Learning with Tool Use},
  author={Jiang, Dongfu and Lu, Yi and Li, Zhuofeng and Lyu, Zhiheng and Nie, Ping and Wang, Haozhe and Su, Alex and Chen, Hui and Zou, Kai and Du, Chao and others},
  journal={arXiv preprint arXiv:2509.01055},
  year={2025}
}

@article{shang2025rstar2,
  title={rStar2-Agent: Agentic Reasoning Technical Report},
  author={Shang, Ning and Liu, Yifei and Zhu, Yi and Zhang, Li Lyna and Xu, Weijiang and Guan, Xinyu and Zhang, Buze and Dong, Bingcheng and Zhou, Xudong and Zhang, Bowen and others},
  journal={arXiv preprint arXiv:2508.20722},
  year={2025}
}

@article{xue2025simpletir,
  title={Simpletir: End-to-end reinforcement learning for multi-turn tool-integrated reasoning},
  author={Xue, Zhenghai and Zheng, Longtao and Liu, Qian and Li, Yingru and Zheng, Xiaosen and Ma, Zejun and An, Bo},
  journal={arXiv preprint arXiv:2509.02479},
  year={2025}
}

@inproceedings{feng2025group,
  title={Group-in-group policy optimization for llm agent training},
  author={Feng, Lang and Xue, Zhenghai and Liu, Tingcong and An, Bo},
  booktitle={NeurIPS},
  year={2025}
}

@article{bai2025qwen2,
  title={Qwen2. 5-vl technical report},
  author={Bai, Shuai and Chen, Keqin and Liu, Xuejing and Wang, Jialin and Ge, Wenbin and Song, Sibo and Dang, Kai and Wang, Peng and Wang, Shijie and Tang, Jun and others},
  journal={arXiv preprint arXiv:2502.13923},
  year={2025}
}

@article{chai2025rlfactory,
  title={RLFactory: A Plug-and-Play Reinforcement Learning Post-Training Framework for LLM Multi-Turn Tool-Use},
  author={Chai, Jiajun and Yin, Guojun and Xu, Zekun and Yue, Chuhuai and Jia, Yi and Xia, Siyu and Wang, Xiaohan and Jiang, Jiwen and Li, Xiaoguang and Dong, Chengqi and others},
  journal={arXiv preprint arXiv:2509.06980},
  year={2025}
}

@inproceedings{du2024improving,
  title={Improving factuality and reasoning in language models through multiagent debate},
  author={Du, Yilun and Li, Shuang and Torralba, Antonio and Tenenbaum, Joshua B and Mordatch, Igor},
  booktitle={ICML},
  year={2024}
}

@misc{kl_approx,
    title = {Approximating KL Divergence} ,
    author = {Schulman, John},
    url = {http://joschu.net/blog/kl-approx.html},
    year = {2020}
}

@inproceedings{
hu2022lora,
title={Lo{RA}: Low-Rank Adaptation of Large Language Models},
author={Edward J Hu and yelong shen and Phillip Wallis and Zeyuan Allen-Zhu and Yuanzhi Li and Shean Wang and Lu Wang and Weizhu Chen},
booktitle={ICLR},
year={2022}
}

@article{jin2025search,
  title={Search-r1: Training llms to reason and leverage search engines with reinforcement learning},
  author={Jin, Bowen and Zeng, Hansi and Yue, Zhenrui and Yoon, Jinsung and Arik, Sercan and Wang, Dong and Zamani, Hamed and Han, Jiawei},
  journal={arXiv preprint arXiv:2503.09516},
  year={2025}
}

@article{wang2025ragen,
  title={Ragen: Understanding self-evolution in llm agents via multi-turn reinforcement learning},
  author={Wang, Zihan and Wang, Kangrui and Wang, Qineng and Zhang, Pingyue and Li, Linjie and Yang, Zhengyuan and Jin, Xing and Yu, Kefan and Nguyen, Minh Nhat and Liu, Licheng and others},
  journal={arXiv preprint arXiv:2504.20073},
  year={2025}
}

@article{chen2023fireact,
  title={FireAct: Toward Language Agent Fine-tuning},
  author={Chen, Baian and Shu, Chang and Shareghi, Ehsan and Collier, Nigel and Narasimhan, Karthik and Yao, Shunyu},
  journal={arXiv preprint arXiv:2310.05915},
  year={2023},
}

@article{zeng2023agenttuning,
  title={AgentTuning: Enabling Generalized Agent Abilities for LLMs},
  author={Zeng, Aohan and Liu, Mingdao and Lu, Rui and Wang, Bowen and Liu, Xiao and Dong, Yuxiao and Tang, Jie},
  journal={arXiv preprint arXiv:2310.12823},
  year={2023},
}

@article{subramaniam2025multiagentft,
  title={Multiagent Finetuning: Self Improvement with Diverse Reasoning Chains},
  author={Subramaniam, Vighnesh and Du, Yilun and Tenenbaum, Joshua B. and Torralba, Antonio and Li, Shuang and Mordatch, Igor},
  journal={arXiv preprint arXiv:2501.05707},
  year={2025},
}

@article{silver2017mastering,
  title={Mastering chess and shogi by self-play with a general reinforcement learning algorithm},
  author={Silver, David and Hubert, Thomas and Schrittwieser, Julian and Antonoglou, Ioannis and Lai, Matthew and Guez, Arthur and Lanctot, Marc and Sifre, Laurent and Kumaran, Dharshan and Graepel, Thore and others},
  journal={arXiv preprint arXiv:1712.01815},
  year={2017}
}

@article{silver2016mastering,
  title={Mastering the game of Go with deep neural networks and tree search},
  author={Silver, David and Huang, Aja and Maddison, Chris J and Guez, Arthur and Sifre, Laurent and Van Den Driessche, George and Schrittwieser, Julian and Antonoglou, Ioannis and Panneershelvam, Veda and Lanctot, Marc and others},
  journal={Nature},
  year={2016}
}

@article{zhang2024accessing,
  title={Accessing gpt-4 level mathematical olympiad solutions via monte carlo tree self-refine with llama-3 8b},
  author={Zhang, Di and Huang, Xiaoshui and Zhou, Dongzhan and Li, Yuqiang and Ouyang, Wanli},
  journal={arXiv preprint arXiv:2406.07394},
  year={2024}
}

@article{rabby2024mc,
  title={MC-NEST: Enhancing Mathematical Reasoning in Large Language Models leveraging a Monte Carlo Self-Refine Tree},
  author={Rabby, Gollam and Keya, Farhana and Auer, S{\"o}ren},
  journal={arXiv preprint arXiv:2411.15645},
  year={2024}
}

@inproceedings{zhang2025llama,
  title={LLaMA-Berry: Pairwise Optimization for Olympiad-level Mathematical Reasoning via O1-like Monte Carlo Tree Search},
 author={Zhang, Di and Wu, Jianbo and Lei, Jingdi and Che, Tong and Li, Jiatong and Xie, Tong and Huang, Xiaoshui and Zhang, Shufei and Pavone, Marco and Li, Yuqiang and others},
  booktitle={NAACL},
  year={2025}
}

@article{chang2025step,
  title={Step-level Verifier-guided Hybrid Test-Time Scaling for Large Language Models},
  author={Chang, Kaiyan and Shi, Yonghao and Wang, Chenglong and Zhou, Hang and Hu, Chi and Liu, Xiaoqian and Luo, Yingfeng and Ge, Yuan and Xiao, Tong and Zhu, Jingbo},
  journal={arXiv preprint arXiv:2507.15512},
  year={2025}
}

@article{nakano2021webgpt,
  title={Webgpt: Browser-assisted question-answering with human feedback},
  author={Nakano, Reiichiro and Hilton, Jacob and Balaji, Suchir and Wu, Jeff and Ouyang, Long and Kim, Christina and Hesse, Christopher and Jain, Shantanu and Kosaraju, Vineet and Saunders, William and others},
  journal={arXiv preprint arXiv:2112.09332},
  year={2021}
}

@inproceedings{suris2023vipergpt,
  title={Vipergpt: Visual inference via python execution for reasoning},
  author={Sur{\'\i}s, D{\'\i}dac and Menon, Sachit and Vondrick, Carl},
  booktitle={ICCV},
  year={2023}
}

@article{press2022selfask,
  title={Measuring and narrowing the compositionality gap in language models},
  author={Press, Ofir and Zhang, Muru and Min, Sewon and Schmidt, Ludwig and Smith, Noah A and Lewis, Mike},
  journal={arXiv preprint arXiv:2210.03350},
  year={2022}
}

@article{gao2023retrieval,
  title={Retrieval-augmented generation for large language models: A survey},
  author={Gao, Yunfan and Xiong, Yun and Gao, Xinyu and Jia, Kangxiang and Pan, Jinliu and Bi, Yuxi and Dai, Yixin and Sun, Jiawei and Wang, Haofen and Wang, Haofen},
  journal={arXiv preprint arXiv:2312.10997},
  year={2023}
}

@article{sheng2024hybridflow,
  title   = {HybridFlow: A Flexible and Efficient RLHF Framework},
  author  = {Guangming Sheng and Chi Zhang and Zilingfeng Ye and Xibin Wu and Wang Zhang and Ru Zhang and Yanghua Peng and Haibin Lin and Chuan Wu},
  year    = {2024},
  journal = {arXiv preprint arXiv: 2409.19256}
}

@article{edge2024local,
  title={From local to global: A graph rag approach to query-focused summarization},
  author={Edge, Darren and Trinh, Ha and Cheng, Newman and Bradley, Joshua and Chao, Alex and Mody, Apurva and Truitt, Steven and Metropolitansky, Dasha and Ness, Robert Osazuwa and Larson, Jonathan},
  journal={arXiv preprint arXiv:2404.16130},
  year={2024}
}

@inproceedings{
asai2024selfrag,
title={Self-{RAG}: Learning to Retrieve, Generate, and Critique through Self-Reflection},
author={Akari Asai and Zeqiu Wu and Yizhong Wang and Avirup Sil and Hannaneh Hajishirzi},
booktitle={ICLR},
year={2024}
}

@inproceedings{gao2023pal,
  title={Pal: Program-aided language models},
  author={Gao, Luyu and Madaan, Aman and Zhou, Shuyan and Alon, Uri and Liu, Pengfei and Yang, Yiming and Callan, Jamie and Neubig, Graham},
  booktitle={ICML},
  year={2023},
}

@article{grattafiori2024llama,
  title={The llama 3 herd of models},
  author={Grattafiori, Aaron and Dubey, Abhimanyu and Jauhri, Abhinav and Pandey, Abhinav and Kadian, Abhishek and Al-Dahle, Ahmad and Letman, Aiesha and Mathur, Akhil and Schelten, Alan and Vaughan, Alex and others},
  journal={arXiv preprint arXiv:2407.21783},
  year={2024}
}

@inproceedings{moritz2018ray,
  title={Ray: A distributed framework for emerging $\{$AI$\}$ applications},
  author={Moritz, Philipp and Nishihara, Robert and Wang, Stephanie and Tumanov, Alexey and Liaw, Richard and Liang, Eric and Elibol, Melih and Yang, Zongheng and Paul, William and Jordan, Michael I and others},
  booktitle={OSDI},
  year={2018}
}

@misc{deepscaler2025,
  title={DeepScaleR: Surpassing O1-Preview with a 1.5B Model by Scaling RL},
  author={Michael Luo and Sijun Tan and Justin Wong and Xiaoxiang Shi and William Tang and Manan Roongta and Colin Cai and Jeffrey Luo and Tianjun Zhang and Erran Li and Raluca Ada Popa and Ion Stoica},
  year={2025},
  note={Notion Blog}
}

@article{li2025limr,
  title={Limr: Less is more for rl scaling},
  author={Li, Xuefeng and Zou, Haoyang and Liu, Pengfei},
  journal={arXiv preprint arXiv:2502.11886},
  year={2025}
}
\bibliographystyle{ref-style}

\clearpage
\onecolumn
\appendix 
\etocdepthtag.toc{mtappendix}
\etocsettagdepth{mtchapter}{none}
\etocsettagdepth{mtappendix}{subsection}
\renewcommand{\contentsname}{Appendix}
\tableofcontents
\clearpage

\section{Related Work}
\label{sec: related work}

In this section, we systematically review prior work related to our three key features: tool-integrated reasoning in Section~\ref{ssec:tir}, multi-model reasoning in Section~\ref{ssec:multi-model reasoning}, and test-time iteration in Section~\ref{ssec:test-time iteration}. 
Analogous to the Apollo Program, where \textit{diverse experts} built \textit{specialized tools} to \textit{iteratively} launch the Apollo missions, these three features are essential to our AlphaApollo system. 



\subsection{Tool-integrated Reasoning}
\label{ssec:tir}

As aforementioned, the capabilities of FMs in tackling complex problems are limited by their insufficient computational ability and domain knowledge.
Tool-integrated reasoning shows effectiveness in mitigating these shortcomings, enabling FMs to leverage external tools to bridge gaps in knowledge and arithmetic.
The following introduces three directions for this paradigm.

\paragraph{Tool-integrated methods.}

Early tool-integrated methods leverage external tools to enhance FMs reasoning, addressing limitations in knowledge and computation. 
(1) For knowledge, Retrieval-Augmented Generation (RAG)~\citep{lewis2020retrieval, nakano2021webgpt} integrates knowledge bases to furnish FMs with essential facts for accurate inference, though its efficacy hinges on precise retrieval of contextually relevant information without noise~\citep{gao2023retrieval}. To ensure retrieval reliability, Self-Ask~\citep{press2022selfask} decomposes queries into sub-questions for targeted retrieval, Self-RAG~\citep{asai2024selfrag} verifies and filters irrelevant chunks, and GraphRAG~\citep{edge2024local} structures knowledge as context graphs to capture relational relevance. 
(2) For computation, Python has emerged as an effective tool for precise computation. Program-aided Language Models  (PAL)~\citep{gao2023pal} and Program of Thoughts (PoT)~\citep{chen2023program} incorporate Python code generation and execution within the reasoning process for arithmetic and logical tasks, while ViperGPT~\citep{suris2023vipergpt} extends this to vision, using code to process images and enrich multimodal reasoning.

\paragraph{Agentic frameworks.}
Beyond tool-integrated methods, agentic frameworks build flexible, tool-integrated environments that allow FMs to invoke tools dynamically, rather than adhering to predefined calling stages. For instance, SciMaster~\citep{chai2025scimaster} introduces agentic workflows that support dynamic reasoning via a Python-based tool for computation and retrieval, augmented by test-time scaling through reflection, solution refinement, and answer selection mechanisms to enhance FMs' ability on complex problems. 
Similarly, OctoTools~\citep{pan2025octotools} offers a reliable framework that integrates diverse tools through detailed tool cards describing their functions and utilities; it deploys a query analyzer agent to select a task-specific tool subset based on tool cards. 
This framework enables efficient tool usage when extensive tools are available, yielding more adaptive tool utilization for complex problems.
Additionally, Alita~\citep{qiu2025alita} proposes a framework for dynamically generating task-specific tools from code. It also leverages web search to iteratively refine both the reasoning process and the design of tools to optimize the solution for real-world tasks.

\paragraph{Learning frameworks.}

While the integration of multiple tools unlocks the reasoning potential of FMs, how to let FMs harness the usage of these tools in solving complex problems remains a significant challenge.
To address this, tool-learning frameworks~\citep{qin2024toolllm, liu2024apigen, gao2024confucius} enhance FMs' tool utilization through targeted post-training. Notably, several frameworks are tailored for tool-integrated long-horizon reasoning, enabling trainable multi-round interactions with tools. For instance, VerlTool~\citep{jiang2025verltool}, RL-Factory~\citep{chai2025rlfactory}, and rStar2-Agent~\citep{shang2025rstar2} employ unified tool managers to create model-friendly tool-use environments, supported by established RL training pipelines via VeRL~\citep{sheng2024hybridflow}. Although these frameworks effectively integrate RL methods into tool-integrated reasoning, the inherent dynamics of such reasoning—particularly long-horizon planning and the incorporation of external tools—pose significant challenges to stable and efficient optimization. To mitigate this, Verl-Agent~\citep{feng2025group} introduces a step-independent rollout mechanism and customizable memory modules, alleviating inherent long-horizon reasoning difficulties. Whereas SimpleTIR~\citep{xue2025simpletir} detects and filters trajectories featuring ``void turns''—instances where reasoning collapses and destabilizes multi-turn agentic training—thereby promoting more robust optimization.

\subsection{Multi-model Reasoning}
\label{ssec:multi-model reasoning}

Multi-model reasoning leverages the strengths of multiple models and allocates sub-tasks across models, which may adopt diverse roles rather than strictly complementary capabilities, to increase accuracy, robustness, and scalability in complex problem-solving.
Representative paradigms include collaborative strategies and adversarial debate, with multi-agent fine-tuning further strengthening the system. We elaborate on each paradigm as follows.

\paragraph{Collaboration.}
Collaboration coordinates multiple models that work \emph{synergistically}, with each contributing specialized capabilities toward a shared objective. 
AutoGen~\citep{wu2024autogen} provides a framework for multi-agent conversations that power next-generation FM applications, enabling agents to jointly tackle tasks such as coding and question answering. 
MetaGPT~\citep{hong2024metagpt} employs role-based multi-agent collaboration and emphasizes structured, human-like workflows with standard operating procedures (SOPs), assigning roles such as product manager and engineer to inject domain expertise and improve efficiency in software-development tasks.
In addition, HuggingGPT~\citep{shen2023hugginggpt} leverages FMs to orchestrate Hugging Face models across modalities, integrating vision and speech to handle multimodal tasks. 

\paragraph{Debate.}
Debate mechanisms engage multiple models in \emph{mutual critique and refinement}, often paired with tool use to verify facts and resolve ambiguities. 
The MAD framework~\citep{du2024improving} formalizes multi-round proposal, cross-examination, and revision among independent FMs before a final judgment, improving mathematical and strategic reasoning, reducing hallucinations, and working even with black-box models under task-agnostic prompts. 
CoA~\citep{li2024improving} advances this paradigm with a sparse communication topology that lowers computational cost relative to fully connected settings while preserving reasoning performance in experiments with GPT and Mistral. 
Complementing these efforts, LLM-Coordination~\citep{agashe2023llm} examines multi-agent behavior in pure coordination games, finding that FMs excel at environment comprehension yet underperform on theory-of-mind reasoning.

\paragraph{Multi-agent fine-tuning.}
Multi-agent fine-tuning jointly optimizes role-specialized models so that, as a group, they plan, call tools, and summarize more effectively than a single FM. 
Early systems fine-tune on agent trajectories (FireAct~\citep{chen2023fireact}) or curated agent-instruction data (AgentTuning~\citep{zeng2023agenttuning}) to endow general agent abilities, providing a foundation for role-based optimization. 
An emerging direction explicitly fine-tunes a society of models from a common base using inter-agent data to diversify skills and improve coordination~\citep{subramaniam2025multiagentft}. 
\citet{shen2024small} decomposes tool-learning into planner, caller, and summarizer roles, each fine-tuned on sub-tasks, achieving superior performance on ToolBench and surpassing single-FM approaches. AgentFly~\citep{zhou2025agentfly} advances this paradigm by introducing memory-based online reinforcement learning for agent fine-tuning without updating FM weights. 
These methods demonstrate how multi-agent fine-tuning enhances coordination and tool usage in multi-turn tasks.

\subsection{Test-time Iteration}
\label{ssec:test-time iteration}

Test-time iteration strategies improve FM reasoning by using extra computational resources to refine solutions or sample and verify diverse answers during inference, leveraging pretraining knowledge.
Broadly, test-time iteration methods are categorized into parallel, sequential, and mixed strategies, distinguished by the interaction between iterations.

\paragraph{Parallel iteration.} 
Most early test-time scaling methods utilize parallel iteration, where the model generates multiple independent reasoning trajectories. The final answer is then selected through various aggregation mechanisms.
Self-Consistency~\citep{wang2022self} employs simple voting, Semantic Self-Consistency~\citep{knappe2024semantic} leverages semantic similarity, CISC~\citep{taubenfeld2025confidence} utilizes confidence metrics, and MCR~\citep{yoran2023answering} implements model-decided final answers. Additionally, DIVSE~\citep{naik2023diversity} enhances the diversity of the reasoning trajectory by reformulating the original question to encourage exploration of a broader solution space. While this strategy enhances FM reasoning by allocating more computational budget to the reasoning process~\citep{snell2024scaling}, it is fundamentally limited by the lack of interaction among different reasoning trajectories. This isolation limits the model in leveraging from previous attempts or refining its approach based on earlier outputs. The independence between trajectories constrains the potential for iterative improvement and fails to capitalize on insights that could emerge from comparing or combining intermediate reasoning steps~\citep{qi2025learning}.

\paragraph{Sequential iteration.}
Sequential iteration strategies leverage prior reasoning processes and outcomes to identify limitations and iteratively refine subsequent generations. Notably, Self-Refine methods~\citep{madaan2023self, qu2024recursive, chen2023teaching} prompt FMs to revise initial outputs, generating improved solutions. Similarly, \citet{muennighoff2025s1} employs a deliberation mechanism, replacing the end-of-sequence token with the keyword `wait' to enable continuous reasoning post-answer generation. OPRO~\citep{yang2024large} iteratively optimizes prompts based on prior outcomes to yield superior solutions. While these approaches enhance reasoning, their reliance on self-correction without external supervision can lead to unreliable outcomes~\citep{huang2024large}. In contrast, Reflexion~\citep{shinn2023reflexion}, RCI~\citep{kim2023language}, and Refiner~\citep{paul2023refiner} incorporate external feedback from environments, code executors, and critic models, respectively, providing robust supervision and improving performance. Furthermore, methods like Alpha-Evolve~\citep{novikov2025alphaevolve} and TextGrad~\citep{yuksekgonul2025optimizing} extend sequential iteration to scientific domains, demonstrating FMs' potential in tackling complex challenges, such as code optimization and molecular synthesis.

\paragraph{Mixed iteration.} 
Mixed iteration strategies integrate parallel and sequential iteration to combine the exploratory breadth of independent trajectories with the refinement depth of conditioned generations. This hybrid approach optimizes test-time compute allocation, often surpassing pure parallel or sequential methods on complex reasoning tasks by balancing exploration (through parallel iteration) and exploitation (via sequential refinement). A prominent example is Monte Carlo Tree Search (MCTS)~\citep{silver2016mastering, silver2017mastering}, which employs multi-round node expansion through parallel iteration and outcome simulation via sequential iteration. Recent advancements apply MCTS to enhance FM reasoning by treating intermediate thoughts as tree nodes and reasoning outcomes as leaf nodes, iterating through node expansion and outcome simulation to derive optimized solutions~\citep{zhang2024accessing, rabby2024mc, zhang2025llama}. Notably, \citet{chang2025step} incorporates self-refinement into node exploration, enabling MCTS to both identify the best node for subsequent iterations and optimize it for maximum exploitation.

\section{Agentic Reasoning: Implementation Details and Discussion}
\label{app: discussion_agentic_reasoning}

\subsection{The Computational Module}
\label{ssec: computation}

\begin{figure}[t!]
\centering
\includegraphics[width=1.0\linewidth]{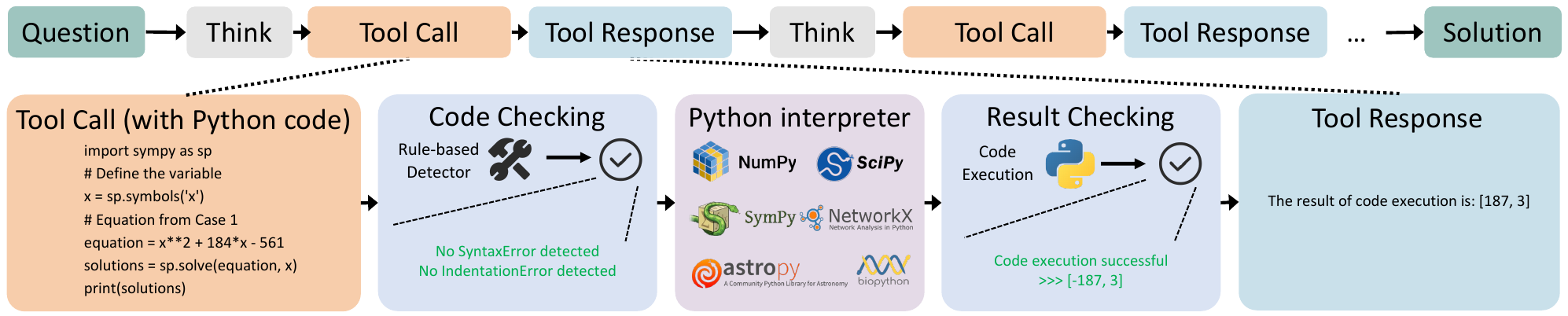}
\caption{
The pipeline of the computational module in processing code.
}
\label{fig: computation}
\end{figure}

This module integrates a Python interpreter with several external libraries of Python, as illustrated in Figure~\ref{fig: computation}. 
Compared to other programming languages, Python’s extensive ecosystem provides a more powerful computational environment for addressing complex reasoning tasks. 
Representative libraries are listed below:

\begin{itemize}[leftmargin=*]
\item \textbf{SymPy}~\citep{meurer2017sympy} is a computer algebra system for symbolic mathematics, enabling the manipulation and solution of mathematical expressions in closed form. 
For example, it can determine the exact roots of the cubic equation $x^3 - 2x + 1$ within Python.  

\item \textbf{NumPy}~\citep{numpy} is a fundamental library for fast, vectorized numerical computing, providing powerful support for $N$-dimensional arrays and matrices. 
Typical tasks include linear algebra operations such as computing the dot product of two matrices.  

\item \textbf{SciPy}~\citep{scipy} is a comprehensive library of advanced numerical algorithms for science and engineering, supporting integration, ordinary differential equations, optimization, signal processing, sparse linear algebra, statistics, and more. 
For instance, it can readily solve problems like finding the minimum of $f(x) = \sin(x) + x^2$ over the interval $[-3,3]$.  
\end{itemize}

\paragraph{Python environment and code execution.}
Notably, this computational module shares the same Python environment as the AlphaApollo project, enabling users to easily extend the toolset by installing new Python libraries. 
When this module receives a tool-call request containing Python code from the model, it generates a temporary Python file and executes it in an individual subprocess, which is isolated from AlphaApollo's main process for safety reasons.

\paragraph{Error correction.}
The model-generated code can contain errors.
To improve the robustness against these errors, AlphaApollo's computational module incorporates a rule-based error-correction.
When the model generates Python code, the rule-based approach detects and automatically corrects errors identifiable through predefined rules, and then passes the corrected code to the Python interpreter for execution. 

In rule-based error-correction, all \texttt{IndentationError} can be detected, with most being automatically correctable. 
The rule-based approach verifies the legality of indentation line by line, identifying and removing unnecessary space tokens to ensure proper indentation throughout the code. 
Additionally, certain \texttt{SyntaxError}, such as extraneous markdown blocks wrapping the code, can be detected and resolved. 
This method extracts the core Python code from wrapped content, ensuring the generated code is executable.


In addition, for errors such as \texttt{ValueError}, \texttt{TypeError}, \texttt{AttributeError}, and \texttt{ImportError}, which commonly arise when using external libraries (e.g., calling \texttt{sympy.factorial} with a negative argument, although it only accepts non-negative values), the computational module can invoke the retrieval module (Sec.~\ref{ssec: information}) for further guidance. 
By combining this capability with the two-fold error-correction mechanism, the computational module creates a model-friendly Python environment that empowers foundation models to perform code-augmented reasoning.

\subsection{The Retrieval Module}
\label{ssec: information}

\begin{figure}[t!]
    \centering
    \includegraphics[width=1.0\linewidth]{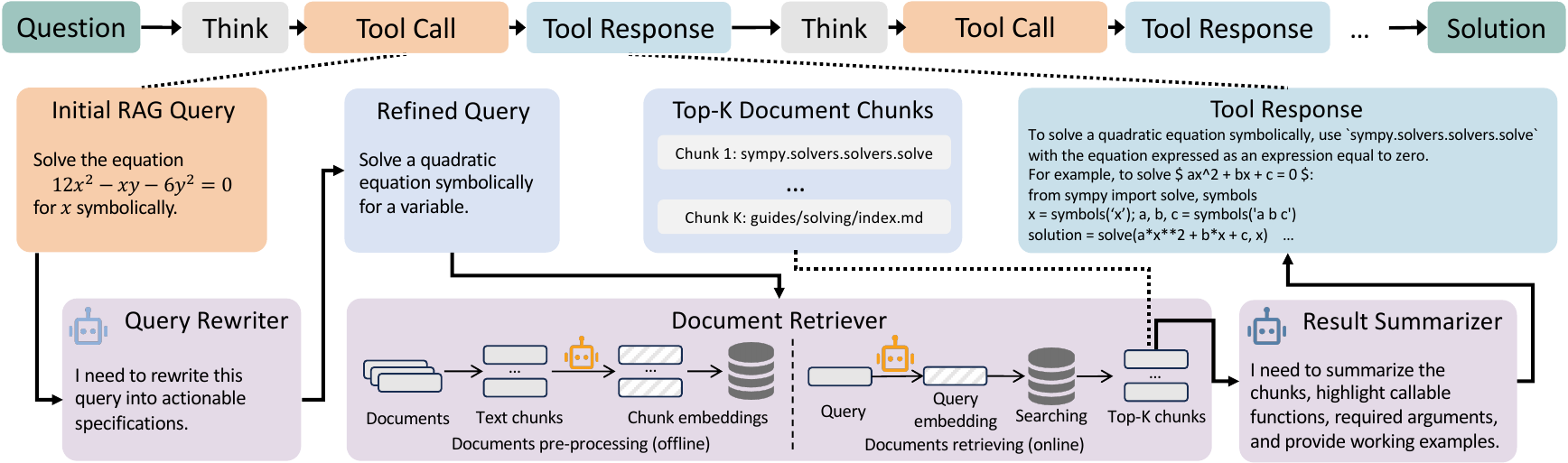}
    \caption{Schematic illustrating the information retrieval process of the retrieve module.
    The purple components correspond to the Retrieval Server. For clarity, Client 2 is omitted from the illustration.
    }
    \label{fig:module-retrieval}
\end{figure}

Foundation models demonstrate notable proficiency in generating code for widely used Python libraries. However, they often exhibit limitations when interfacing with less common libraries in solving complex problems, such as \textit{NetworkX} for graphical problems or \textit{SymPy} for symbolic mathematical problems.
This can result in hallucinated or erroneous function calls, which compromise the reliability of tool integration.
To mitigate this limitation, we augment AlphaApollo with a retrieval module that improves function invocation accuracy by \textit{retrieving relevant functions and usage examples from library documentation}. 
Here, we offload the underlying complexities of information retrieval and processing to the retrieval module, allowing the (main) model to focus more on formulating the retrieval queries rather than selecting retrieval systems or information sources.
As shown in Figure~\ref{fig:module-retrieval}, the retrieval module comprises three core components: a query rewriter, a document retriever, and a result summarizer.
The module follows a single-pass workflow that rewrites queries for clarity, retrieves semantically relevant documents, and summarizes documents to remove potentially redundant content. 
Specifically, each component functions as follows:
\begin{itemize}[leftmargin=*]
\item
\textbf{Query rewriter.} The workflow begins with the \textit{query rewriter}, which transforms the initial query into a retrieval-friendly specification. Instead of passing detailed task descriptions directly, the rewriter abstracts them into generalizable forms that emphasize functional intent. For example, an over-detailed query ``Solve the equation $12x^2 - xy - 6y^2 = 0$ for $x$ symbolically'' is rewritten as a more suitable one, ``Solve a quadratic equation symbolically for a variable.'' 
This reformulation abstracts away numerical details while retaining the core intent, resulting in more relevant retrieval terms for the retriever. We implement the query rewriter using an instruction-following model with a tailored prompt template. 

\item
\textbf{Document retriever.} After rewriting the query, the \textit{document retriever} searches for relevant information in the module's indexed corpus, which includes the source code of a Python library and its associated documentation.
Technically, we partition the corpus into \textit{overlapped chunks}, which could improve retrieval effectiveness by aligning units with sentence-level semantics and reducing the inclusion of irrelevant content.
We implement overlapped chunking with a fixed-length sliding window, allowing adjacent chunks to share overlapping tokens and thus preserve cross-boundary context. These chunks are then encoded by the embedding model and stored in a \textit{vector database}.
When the retriever receives a refined query from the rewriter, it first encodes the query using the same embedding model employed during document indexing. The retriever then conducts a cosine-similarity search over the database to locate the top-$K$ chunk embeddings relevant to the query. The corresponding document segments are retrieved and assembled into a composite context, which is then passed to the summarization module.

\item
\textbf{Result summarizer.} Finally, the \textit{result summarizer} filters and summarizes the retrieved context into a concise response. Its role is to highlight callable functions, required arguments, and minimal working examples rather than returning raw documentation.
For instance, when the retriever provides (i) demonstrations of \texttt{sympy.solve}, (ii) module descriptions of \texttt{solveset}, and (iii) general API notes on \texttt{sympy}, the summarizer distills them into an actionable tool description. It identifies \texttt{solve} as the appropriate function, specifies its key arguments, and generates a compact example such as: \texttt{from sympy import solve, symbols; a, b, c, x = symbols('a b c x'); solve(ax**2 + bx + c, x)}.
\end{itemize}

In summary, the retrieval module aims to improve function invocation in Python.
It enables models to utilize external Python libraries to guide the generation and refinement of the generated code, grounding the model-generated code in reliable documentation.

\subsection{Experimental Settings}
\label{app: experimental settings reasoning}

In this setting, we enable tool-augmented reasoning as a baseline without any training or agentic evolution. The model performs multi-turn interactions with access to computational (Python code execution) and retrieval (local RAG) tools, limited to a maximum of 4 rounds ($ T_\text{max}=4 $) to facilitate step-by-step problem-solving. We set the temperature to 0.6 for balanced exploration, with history length capped at 8 to maintain context across turns. No additional learning or verification mechanisms are applied, and each question is evaluated in a single-pass manner.

\subsection{Prompts}
\label{app: prompts: system prompt}
In this part, we provide the employed prompts in the AlphaApollo in Figures~\ref{fig: system prompt no his} and~\ref{fig: system prompt with his}.

\begin{figure}[t!]
\centering
\begin{tcolorbox}[
    title={System prompt of the first turn},
    colframe=lightblue,
    boxrule=0.3mm, width=0.9\textwidth, arc=3mm, auto outer arc=true, fontupper=\footnotesize
]
You are a math problem solver agent tasked with solving the given math problem step-by-step.\\

Your question: \{question\}\\

Now it's your turn to respond to the current step.

You should first conduct the reasoning process. This process MUST be enclosed within \textless think \textgreater \textless /think \textgreater tags. 

After completing your reasoning, choose only one of the following actions (do not perform multiple actions at the same time):

1) \textless python\_code\textgreater...\textless /python\_code\textgreater: If computation/checking is helpful, emit exactly ONE \textless python\_code\textgreater...\textless /python\_code\textgreater block with pure Python 3. Inspect the \textless tool\_response\textgreater (stdout from your code). If it disagrees with your reasoning, correct yourself.

2) \textless local\_rag\textgreater...\textless/local\_rag\textgreater: You have access to a RAG System tool to search for documentation or examples (Supported repos: sympy, scipy, numpy, math, cmath, fractions, itertools). Emit exactly ONE \textless local\_rag\textgreater...\textless/local\_rag\textgreater block with a JSON object. Inspect the returned \textless tool
\_response\textgreater (RAG result). If it disagrees with your reasoning, correct yourself. For example: \textless local\_rag\textgreater \{\{"repo\_name": "sympy", "query": "your query here"\}\}\textless /local\_rag\textgreater.

3) \textless answer\textgreater...\textless /answer\textgreater: If you are ready to provide the self-contained solution, provide the answer only inside \textless answer\textgreater...\textless /answer\textgreater, formatted in LaTeX, e.g., \textbackslash \textbackslash boxed\{\{...\}\}.

\end{tcolorbox}
\caption{In the system prompt of the first turn, we instruct the model to solve math problems step by step, decide which tools to use, and present the final answer in a boxed format.}
\label{fig: system prompt no his}
\end{figure}

\begin{figure}[t!]
\centering
\begin{tcolorbox}[
    title={System prompt for subsequent turns},
    colframe=lightblue,
    boxrule=0.3mm, width=0.9\textwidth, arc=3mm, auto outer arc=true, fontupper=\footnotesize
]
You are a math problem solver agent tasked with solving the given math problem step-by-step.\\

Your question: \{question\}\\

Prior to this step, you have already taken \{step\_count\} step(s).
Below is the interaction history:
\{memory\_context\}

Now it's your turn to respond to the current step.
You should first conduct the reasoning process. This process MUST be enclosed within
\textless think\textgreater\ \textless/think\textgreater\ tags.

After completing your reasoning, choose only one of the following actions (do not perform multiple actions at the same time):

1) \textless python\_code\textgreater...\textless/python\_code\textgreater:
If computation/checking is helpful, emit exactly ONE
\textless python\_code\textgreater...\textless/python\_code\textgreater block with pure Python 3.
Inspect the \textless tool\_response\textgreater\ (stdout from your code).
If it disagrees with your reasoning, correct yourself.

2) \textless local\_rag\textgreater...\textless/local\_rag\textgreater:
You have access to a RAG System tool to search for documentation or examples
(Supported repos: sympy, scipy, numpy, math, cmath, fractions, itertools).
Emit exactly ONE \textless local\_rag\textgreater...\textless/local\_rag\textgreater
block with a JSON object.
Inspect the returned \textless tool\_response\textgreater\ (RAG result).
If it disagrees with your reasoning, correct yourself.
For example:
\textless local\_rag\textgreater\{\{"repo\_name": "sympy", "query": "your query here"\}\}\textless/local\_rag\textgreater.

3) \textless answer\textgreater...\textless/answer\textgreater:
If you are ready to provide the self-contained solution, provide the answer only inside
\textless answer\textgreater...\textless/answer\textgreater,
formatted in LaTeX, e.g., \textbackslash\textbackslash boxed\{\{...\}\}.
\end{tcolorbox}
\caption{In the system prompt of subsequent turns, we integrate the memory from previous turns and instruct the model to continue solving the math problem step by step, decide which tools to use, and present the final answer in a boxed format.}
\label{fig: system prompt with his}
\end{figure}

\section{Agentic Learning: Implementation Details and Discussion}
\label{app: discussion_agentic_learning}

Prior agentic learning systems, such as VerlTool~\citep{jiang2025verltool} and RL-Factory~\citep{chai2025rlfactory}, stabilize agentic learning by employing a masking strategy that assigns zero weight to tokens from tool responses, thereby excluding them from optimization.
Formally, considering the trajectory sampling process of GRPO as an example, the objective for trajectory-level RL optimization is:
\begin{equation}
\begin{aligned}
\mathcal{J}_{\text{Trajectory-GRPO}}(\theta)
&=
\mathbb{E}_{\substack{p_1\sim \mathcal{D}\\\{y^{(i)}_k\}_{i=1}^N \sim \pi_{\text{old}}(\cdot|p_1)}}
\Bigl[
\frac{1}{N}
\sum_{i=1}^{N}\frac{1}{|y^{(i)}|}\sum_{k=1}^{|y^{(i)}|}
\min  \Big(m(y_k^{(i)})
\frac{\pi_\theta\big(y_k^{(i)} \mid p_1, y_{<k}^{(i)}\big)}
{\pi_{\text{old}}\big(y_k^{(i)} \mid p_1, y_{<k}^{(i)}\big)} A_k^{(i)},\\
&
\operatorname{clip}\big(m(y_t^{(i)})
\frac{\pi_\theta\big(y_k^{(i)} \mid p_1, y_{<k}^{(i)}\big)}
{\pi_{\text{old}}\big(y_k^{(i)} \mid p_1, y_{<k}^{(i)}\big)}, 1-\epsilon,1+\epsilon\big)A_k^{(i)} 
\Big) -\beta\mathbb{D}_{\mathrm{KL}}\big[\pi_\theta \big\|\pi_{\mathrm{ref}}\big]
\Bigr]
\label{eq:previous_grpo_objective}
\end{aligned}
\end{equation}
where $y_k^{(i)}$ denotes $k$-th token of the $(i)$-th sampled trajectory, $m(y_k^{(i)}) = \mathds{1}\big(y_k^{(i)} \in \mathcal{O}\big)$ indicates the masking strategy that optimize the policy with the model-generation content $\mathcal{O}$. 

Similarly, the SFT in trajectory-level optimization learns given the full agentic reasoning trajectory while masking tokens from external tools.

\begin{equation}
\mathcal{J}_{\text{Trajectory-SFT}}(\theta) \;=\; \mathbb{E}_{(x,y)\sim \mathcal{D}} \left[ \sum_{k=1}^{|y|} m(y_k) \log \pi_\theta(y_k \mid y_{<k}, x) \right],
\end{equation}
Although the masking strategy prevents the policy from learning tool responses, thereby enhancing stability, the long-horizon nature of agentic reasoning can still destabilize training. Low-probability tokens arising in extended trajectories may compromise optimization, leading to reward collapse, as evidenced in studies on agentic reinforcement learning~\citep{xue2025simpletir,wang2025ragen}.

\subsection{Experimental Settings}
\label{app: experimental settings learning}

In this phase, we apply a learning rate of 1e-6, a training batch size of 128,
and a group size of 8.
The temperature
is set to 0.4 to ensure stable performance monitoring throughout the reinforcement learning process.
The training datasets consist of Lighteval-Math, LIMR, and DeepScaleR for comprehensive mathematical reasoning coverage. To demonstrate the generality of our framework, we trained on each of these datasets separately in our experiments.

\subsection{Prompts}
\label{app: prompts: learning prompt}

We use the same prompts as in Appendix~\ref{app: prompts: system prompt}, shown in Figures~\ref{fig: system prompt no his} and \ref{fig: system prompt with his}.
\section{Agentic Evolution: Implementation Details and Discussion}
\label{app: discussion_agentic_evolution}

\subsection{Experimental Settings}
\label{app: experimental settings evolution}

\begin{table*}[h!]
\small
\centering
\setlength{\tabcolsep}{5pt}
\renewcommand{\arraystretch}{1.1}
\fontsize{9}{9}\selectfont
\caption{
The detailed settings of models in agentic evolution. Here, output length refers to the maximum response length specified during evaluation of evolution.
}
\label{tab:model-info}
\begin{tabular}{clcc}
\toprule
\textbf{\phantom{xx}Model Family\phantom{xx}}&\textbf{\phantom{xx}\phantom{xx}Model Name\phantom{xx}}& \makecell{Maximum\\context length}& \makecell{Output\\length} \\
\midrule
\multirow{4}{*}{Qwen2.5~\citep{yang2024qwen2}}
&Qwen2.5-1.5B-Instruct & 128K & 8k \\
&Qwen2.5-3B-Instruct & 128K & 8k  \\
&Qwen2.5-7B-Instruct & 128K & 8k \\
&Qwen2.5-14B-Instruct & 128K & 8k \\
\bottomrule
\end{tabular}
\end{table*}

We show the configuration of agentic evolution in Table~\ref{tab:model-info}.
This phase involves no training and utilizes a dual-model configuration. The policy model operates with a temperature of 0.7 to encourage exploration and diverse solution generation, while the verifier uses a temperature of 0.4. The self-evolving process runs for 10 rounds, the verifier conducts repeated verification for 5 times to ensure majority judgment. Both models share a maximum token limit of 8k to accommodate complex reasoning trajectories.

\begin{figure}[t!]
\centering
\definecolor{Solver}{HTML}{F5B988} 
\definecolor{Abstractor}{HTML}{BF6000}
\definecolor{Evaluator}{HTML}{9DC3E6}
\definecolor{Summarizer}{HTML}{A2BCE2}
\begin{tcolorbox}[
    title={Prompt for \texttt{Solver} with $\mathcal{M}^{(r)}$ for the first turn},
    colframe=Solver,
    boxrule=0.3mm, width=0.9\textwidth, arc=3mm, auto outer arc=true, fontupper=\footnotesize
]
You are a math problem solver agent tasked with solving the given math problem step-by-step.\\

Your question: \{question\}\\

Below are the previous solutions and their verification feedback:
\{previous\_solutions\}\\

The \textbackslash boxed\{1\} within feedback indicates that the previous solution was correct, and \textbackslash boxed\{0\} indicates that the previous solution was incorrect. Use the previous solutions and their verification feedback to guide your current step. Remember that if the previous solution was incorrect, you should correct your reasoning and try again.\\

Now it's your turn to respond to the current step.
You should first conduct the reasoning process. This process MUST be enclosed within
\textless think\textgreater\ \textless/think\textgreater\ tags. After completing your reasoning, choose only one of the following actions (do not perform multiple actions at the same time):\\

1) \textless python\_code\textgreater...\textless/python\_code\textgreater:
If computation/checking is helpful, emit exactly ONE
\textless python\_code\textgreater...\textless/python\_code\textgreater block with pure Python 3.
Inspect the \textless tool\_response\textgreater\ (stdout from your code).
If it disagrees with your reasoning, correct yourself.

2) \textless local\_rag\textgreater...\textless/local\_rag\textgreater:
You have access to a RAG System tool to search for documentation or examples
(Supported repos: sympy, scipy, numpy, math, cmath, fractions, itertools).
Emit exactly ONE \textless local\_rag\textgreater...\textless/local\_rag\textgreater
block with a JSON object.
Inspect the returned \textless tool\_response\textgreater\ (RAG result).
If it disagrees with your reasoning, correct yourself.
For example:
\textless local\_rag\textgreater\{\{"repo\_name": "sympy", "query": "your query here"\}\}\textless/local\_rag\textgreater.

3) \textless answer\textgreater...\textless/answer\textgreater:
If you are ready to provide the self-contained solution, provide the answer only inside
\textless answer\textgreater...\textless/answer\textgreater,
formatted in LaTeX, e.g., \textbackslash\textbackslash boxed\{...\}.
\end{tcolorbox}
\caption{Prompt for \texttt{Solver} with $\mathcal{M}^{(r)}$ for the first turn. The system instruction for the first turn, incorporating the set of previous solutions and their validity ($\mathcal{M}^{(r)}$).}
\label{fig:prompt-solver_init_turn_with_previous_solutions}
\end{figure}

\begin{figure}[t!]
\centering
\definecolor{Solver}{HTML}{F5B988} 
\definecolor{Abstractor}{HTML}{BF6000}
\definecolor{Evaluator}{HTML}{9DC3E6}
\definecolor{Summarizer}{HTML}{A2BCE2}
\begin{tcolorbox}[
    title={Prompt for \texttt{Solver} with $\mathcal{M}^{(r)}$ for subsequence turn},
    colframe=Solver,
    boxrule=0.3mm, width=0.9\textwidth, arc=3mm, auto outer arc=true, fontupper=\footnotesize
]
You are a math problem solver agent tasked with solving the given math problem step-by-step.\\

Your question: \{question\}\\

Below are the previous solutions and their verification feedback:
\{previous\_solutions\}\\

The \textbackslash boxed\{1\} within feedback indicates that the previous solution was correct, and \textbackslash boxed\{0\} indicates that the previous solution was incorrect. Use the previous solutions and their verification feedback to guide your current step. Remember that if the previous solution was incorrect, you should correct your reasoning and try again.\\

Prior to this step, you have already taken \{step\_count\} step(s).
Below is the interaction history:
\{memory\_context\}\\

Now it's your turn to respond to the current step.
You should first conduct the reasoning process. This process MUST be enclosed within
\textless think\textgreater\ \textless/think\textgreater\ tags. After completing your reasoning, choose only one of the following actions (do not perform multiple actions at the same time):\\

1) \textless python\_code\textgreater...\textless/python\_code\textgreater:
If computation/checking is helpful, emit exactly ONE
\textless python\_code\textgreater...\textless/python\_code\textgreater block with pure Python 3.
Inspect the \textless tool\_response\textgreater\ (stdout from your code).
If it disagrees with your reasoning, correct yourself.

2) \textless local\_rag\textgreater...\textless/local\_rag\textgreater:
You have access to a RAG System tool to search for documentation or examples
(Supported repos: sympy, scipy, numpy, math, cmath, fractions, itertools).
Emit exactly ONE \textless local\_rag\textgreater...\textless/local\_rag\textgreater
block with a JSON object.
Inspect the returned \textless tool\_response\textgreater\ (RAG result).
If it disagrees with your reasoning, correct yourself.
For example:
\textless local\_rag\textgreater\{\{"repo\_name": "sympy", "query": "your query here"\}\}\textless/local\_rag\textgreater.

3) \textless answer\textgreater...\textless/answer\textgreater:
If you are ready to provide the self-contained solution, provide the answer only inside
\textless answer\textgreater...\textless/answer\textgreater,
formatted in LaTeX, e.g., \textbackslash\textbackslash boxed\{...\}.
\end{tcolorbox}
\caption{Prompt for \texttt{Solver} with $\mathcal{M}^{(r)}$ for the subsequent turns. Instructions for subsequent turns where the agent must integrate the interaction history (\texttt{memory\_context}) with the evolutionary feedback from $\mathcal{M}^{(r)}$ to advance the solution step-by-step.}
\label{fig:prompt-solver_subseq_turn_with_previous_solutions}
\end{figure}

\begin{figure}[h!]
\centering
\definecolor{Solver}{HTML}{F8C9A8}
\definecolor{Abstractor}{HTML}{F5B988}
\definecolor{Evaluator}{HTML}{9DC3E6}
\definecolor{Summarizer}{HTML}{A2BCE2}
\begin{tcolorbox}[
    title={Prompt for \texttt{Abstractor}},
    colframe=Abstractor,
    boxrule=0.3mm, width=0.9\textwidth, arc=3mm, auto outer arc=true, fontupper=\footnotesize
]
\label{prompt:abstractor}
ROLE\\
You are a Mathematical Logic Auditor. Compress the interaction log into a Verification Brief.\\
\\
INPUT DATA\\
The log contains `\textless think\textgreater `, `\textless python\_code\textgreater `, `\textless tool\_response\textgreater `, and `\textless answer\textgreater ` tags.\\
\\
PROTOCOL\\
1. Filter Noise: Retain only the mathematical setup, derived constants, and successful logic. Discard syntax errors, backtracking, and internal monologue.\\
2. Track Origins: Explicitly differentiate between values computed via code and those asserted via intuition or external knowledge.\\
3. Format: Use LaTeX for math. Present as sequential Logical Checkpoints.\\
\\
OUTPUT TEMPLATE\\
Strategy: [Brief summary of the approach]\\
\\
Logical Checkpoints:\\
1. Setup: [Variable definitions and initial conditions]\\
2. Intermediate Result: [Key derived values]\\
3. Pivotal Step: [Crucial logic or calculation]\\
4. Resolution: [How the final result was reached]\\
\\
Final Claim: \textless answer\textgreater ...\textless /answer\textgreater \\
\\
TASK\\
Summarize:\\
\{content\}\\
\end{tcolorbox}
\caption{System instructions for \texttt{Abstractor}. The agent generates a structured summary by extracting the key mathematical setup, intermediate results, and the pivotal logic leading to the final answer, while discarding irrelevant content.}
\label{fig:prompt-abstractor}
\end{figure}

\begin{figure}[h!]
\centering
\definecolor{Solver}{HTML}{F8C9A8}
\definecolor{Abstractor}{HTML}{F5B988}
\definecolor{Evaluator}{HTML}{9DC3E6}
\definecolor{Summarizer}{HTML}{A2BCE2}
\begin{tcolorbox}[
    title={Prompt for \texttt{Evaluator} for the first turns.},
    colframe=Evaluator,
    boxrule=0.3mm, width=0.9\textwidth, arc=3mm, auto outer arc=true, fontupper=\footnotesize
]
You are a math verifier agent whose only job is to check whether the solver agent's proposed solution is correct.\\

Original question:\\
\{question\}\\

Solver's latest solution attempt:\\
\{solver\_solution\}\\

Now it's your turn to respond to the current step. You should first conduct the reasoning process. This process MUST be enclosed within \textless think\textgreater  \textless /think\textgreater tags. After completing your reasoning, choose only one of the following actions (do not perform both):\\

1) \textless python\_code\textgreater ...\textless /python\_code\textgreater : If computation/checking is helpful, emit exactly ONE \textless python\_code\textgreater ...\textless /python\_code\textgreater  block with pure Python 3. Inspect the \textless tool\_response\textgreater  (stdout from your code). If it disagrees with your reasoning, correct yourself.\\
2) \textless local\_rag\textgreater...\textless/local\_rag\textgreater:
You have access to a RAG System tool to search for documentation or examples
(Supported repos: sympy, scipy, numpy, math, cmath, fractions, itertools).
Emit exactly ONE \textless local\_rag\textgreater...\textless/local\_rag\textgreater
block with a JSON object.
Inspect the returned \textless tool\_response\textgreater\ (RAG result).
If it disagrees with your reasoning, correct yourself.
For example:
\textless local\_rag\textgreater\{\{"repo\_name": "sympy", "query": "your query here"\}\}\textless/local\_rag\textgreater.\\
3) \textless report\textgreater ...\textless /report\textgreater : If you are ready to conclude, wrap your verification report inside \textless report\textgreater ...\textless /report\textgreater  tags. The report should:\\
- Clearly state whether the policy solution appears correct or incorrect.\\
- Explain the key reasoning behind your judgment (keep it concise).\\
- End with your judgement in the format: \textbackslash boxed\{\{1\}\} if correct, or \textbackslash boxed\{\{0\}\} if incorrect.\\
- The judgement should be enclosed within \textless report\textgreater ...\textless /report\textgreater  tags.\\
\end{tcolorbox}
\caption{The system instructions for the evaluator agent for the first turn. It receives the problem and the candidate solution, utilizes available tools to audit the logic, and outputs a verification.}
\label{fig:prompt-evaluator_first_turn}
\end{figure}

\begin{figure}[h!]
\centering
\definecolor{Solver}{HTML}{F8C9A8}
\definecolor{Abstractor}{HTML}{F5B988}
\definecolor{Evaluator}{HTML}{9DC3E6}
\definecolor{Summarizer}{HTML}{A2BCE2}
\begin{tcolorbox}[
    title={Prompt for \texttt{Evaluator} for the subsequence turns.},
    colframe=Evaluator,
    boxrule=0.3mm, width=0.9\textwidth, arc=3mm, auto outer arc=true, fontupper=\footnotesize
]
You are a math verifier agent whose only job is to check whether the solver agent's proposed solution is correct.\\

Original question:\\
\{question\}\\

Solver's latest solution attempt:\\
\{solver\_solution\}\\

Prior to this step, you have already taken \{step\_count\} step(s).
Below is the interaction history:
\{memory\_context\}\\

Now it's your turn to respond to the current step. You should first conduct the reasoning process. This process MUST be enclosed within \textless think\textgreater  \textless /think\textgreater tags. After completing your reasoning, choose only one of the following actions (do not perform both):\\

1) \textless python\_code\textgreater ...\textless /python\_code\textgreater : If computation/checking is helpful, emit exactly ONE \textless python\_code\textgreater ...\textless /python\_code\textgreater  block with pure Python 3. Inspect the \textless tool\_response\textgreater  (stdout from your code). If it disagrees with your reasoning, correct yourself.\\
2) \textless local\_rag\textgreater...\textless/local\_rag\textgreater:
You have access to a RAG System tool to search for documentation or examples
(Supported repos: sympy, scipy, numpy, math, cmath, fractions, itertools).
Emit exactly ONE \textless local\_rag\textgreater...\textless/local\_rag\textgreater
block with a JSON object.
Inspect the returned \textless tool\_response\textgreater\ (RAG result).
If it disagrees with your reasoning, correct yourself.
For example:
\textless local\_rag\textgreater\{\{"repo\_name": "sympy", "query": "your query here"\}\}\textless/local\_rag\textgreater.\\
3) \textless report\textgreater ...\textless /report\textgreater : If you are ready to conclude, wrap your verification report inside \textless report\textgreater ...\textless /report\textgreater  tags. The report should:\\
- Clearly state whether the policy solution appears correct or incorrect.\\
- Explain the key reasoning behind your judgment (keep it concise).\\
- End with your judgement in the format: \textbackslash boxed\{\{1\}\} if correct, or \textbackslash boxed\{\{0\}\} if incorrect.\\
- The judgement should be enclosed within \textless report\textgreater ...\textless /report\textgreater  tags.\\
\end{tcolorbox}
\caption{The system instruction for \texttt{Evaluator} for subsequent turns.}
\label{fig:prompt-evaluator_subsequence_turn}
\end{figure}

\begin{figure}[h!]
\centering
\definecolor{Solver}{HTML}{F8C9A8}
\definecolor{Abstractor}{HTML}{F5B988}
\definecolor{Evaluator}{HTML}{9DC3E6}
\definecolor{Summarizer}{HTML}{A2BCE2}
\begin{tcolorbox}[
    title={Prompt for \texttt{Summarizer}},
    colframe=Summarizer,
    boxrule=0.3mm, width=0.9\textwidth, arc=3mm, auto outer arc=true, fontupper=\footnotesize
]
\label{prompt:summarizer}
You are a math verification report aggregator. Multiple evaluators have independently verified a solver agent's solution and reached the same judgment. Your task is to synthesize their reports into a single, comprehensive report.\\

Original question:\\
\{question\}\\

Solver agent's solution:\\
\{Solver\_solution\}\\

The evaluators have all concluded with judgment: \textbackslash boxed\{\{\{majority\_judgment\}\}\}\\

Below are the individual reports:\\
\{individual\_reports\}\\

Your task:\\
1. Analyze the key reasoning points from each verifier report.\\
2. Synthesize the most compelling arguments and evidence into a single, coherent report.\\
3. Ensure the aggregated report is comprehensive yet concise.\\
4. Maintain the same final judgment as the individual reports.\\

Wrap your aggregated report inside \textless report\textgreater ...\textless /report\textgreater  tags. The report should:\\
- Clearly state whether the policy solution is correct or incorrect.\\
- Combine the strongest reasoning from all verifier reports.\\
- End with the judgment: \textbackslash boxed\{majority\_judgment\}\\
\end{tcolorbox}
\caption{Instructions for summarizing verifications. The agent summarizes the key information from several concordant evaluators into a single representative judgment, maintaining the original consensus judgment.}
\label{fig:prompt-summarizer}
\end{figure}

\subsection{Prompts}
\label{app: evolution prompts}

In this part, we present the detailed system prompts governing the specialized agents within our agentic evolution framework (Sec.~\ref{ssec: agentic evolution}). These prompts dictate the specific behaviors, tool permissions, and interaction protocols for each role.

\texttt{Solver}: The Solver is responsible for generating solution trajectories. In the initial round, the prompts follow the standard protocols shown in Figures~\ref{fig: system prompt no his} and \ref{fig: system prompt with his} (Sec.~\ref{app: prompts: system prompt}). In subsequent rounds, the solver is augmented with long-term memory entities. Figure~\ref{fig:prompt-solver_init_turn_with_previous_solutions} illustrates the initialization prompt containing the history of past attempts ($\mathcal{M}^{(r)}$), while Figure~\ref{fig:prompt-solver_subseq_turn_with_previous_solutions} shows the instruction for intermediate steps, which integrates both the previous attempts and the local interaction context.

\texttt{Abstractor}: To facilitate efficient memory storage, the Abstractor (Figure~\ref{fig:prompt-abstractor}) is tasked with compressing raw interaction logs. It filters out computational noise, syntax errors, and backtracking, restructuring the successful reasoning path into concise \quot{logical checkpoints}. This process distinguishes between computed values and intuitive assertions, ensuring the stored memory is high-quality and retrieval-ready.

\texttt{Evaluator}: As shown in Figure~\ref{fig:prompt-evaluator_first_turn}, the agent is instructed to verify the solver's candidate solution against the original problem. The prompt enforces a "Think-Check-Report" lifecycle, explicitly authorizing the use of tools to validate intermediate steps and definitions before issuing a binary correctness judgment. For multi-turn verification sessions, the prompt in Figure~\ref{fig:prompt-evaluator_subsequence_turn} adapts to include the interaction history.

\texttt{Summarizer}: Designed for ensemble verification scenarios, the Summarizer (Figure~\ref{fig:prompt-summarizer}) is activated when multiple evaluators reach a consensus. It aggregates individual verification into a single judgement by synthesizing the arguments and removing redundancy, thereby providing a comprehensive justification.
\clearpage

\section{Additional Experimental Results}
\label{app: additional_experimental_results}

In this section, we provide additional results of the experiments.

\subsection{Agentic evolution}
\label{app: evolving_additional_results}

\begin{table*}[t!]
\centering
\caption{Agentic evolution results (accuracy in \%). w/o Evo uses tools without evolution; +Evo enables agentic evolution with tools. \textbf{Bold} marks the better result.}
\vspace{-8pt}
\label{tab:evolving_additional_results}
\setlength{\tabcolsep}{15pt}
\renewcommand{\arraystretch}{1.1}
\fontsize{11}{11}\selectfont
\begin{tabular}{@{}lcc@{}}
\toprule

\multirow{3}{*}{\textbf{Datasets}} & 
\multicolumn{2}{c}{\textbf{Qwen3-4B-Instruct-2507}} 
\\
\cmidrule(lr){2-3}
& \textbf{w/o Evo} & \textbf{+Evo}
\\
\midrule

\textbf{AIME24} 
& 44.98 & \textbf{81.68} 
\\

\textbf{AIME25} 
& 32.50 & \textbf{61.68} 
\\

\textbf{CMIMC 25} 
& \textbf{16.25} & \textbf{16.25} 
\\

\textbf{HMMT25 Feb} 
& 15.83 & \textbf{43.33}
\\

\textbf{HMMT25 Nov} 
& 23.30 & \textbf{42.53}
\\

\textbf{BRUMO25} 
& 24.15 & \textbf{51.68}
\\

\textbf{SMT 2025} 
& 30.65 & \textbf{32.58}
\\

\midrule
\multirow{2}{*}{\textbf{Average}} 
& \multirow{2}{*}{26.81} & \textbf{47.10}  
\\

&
& {\scriptsize (20.29$\uparrow$)}
\\
\bottomrule
\end{tabular}
\vspace{-8pt}
\end{table*}

\paragraph{Additional evidence on the Qwen3 series confirms the benefit of agentic evolution.}
Additionally, we conduct experiments on Qwen3-4B-Instruct (Table~\ref{tab:evolving_additional_results}) to further validate the effect of evolution. Enabling evolution (+Evo) substantially outperforms the non-evolving setting (w/o Evo), improving average accuracy from 26.81\% to 47.10\% (+20.29). The largest gains appear on AIME24 (\(+36.70\)) and AIME25 (\(+29.18\)), with similarly strong improvements on HMMT25 Feb (\(+27.50\)), BRUMO25 (\(+27.53\)), and HMMT25 Nov (\(+19.23\)). Overall, these results on the Qwen3 series further demonstrate that agentic evolution is robust and consistently improves mathematical reasoning with tools.
\section{Case Studies}
\label{app: case}

\subsection{Python Code Errors}
\label{app: case python code errors}

In this section, we show cases of model-generated Python code scripts with the AlphaApollo framework. Therein, partial \texttt{SyntaxError} and \texttt{IndentationError} can be solved with the rule-based error correction component as mentioned in Sec.~\ref{ssec: computation}, while other errors that cannot be solved with rules are refined with the model-based error correction component. We list them as follows:
\begin{itemize}
    \item Solved with the rule-based error correction component.
    \begin{itemize}
        \item \texttt{SyntaxError}(Figure~\ref{fig: syntaxerror can solved by rule});
        \item \texttt{IndentationError} (Figure~\ref{fig: indentationerror});
    \end{itemize}
    \item Solved with the model-based error correction component.
    \begin{itemize}
        \item \texttt{SyntaxError} (Figure~\ref{fig: syntaxerror cannot solved by rule});
        \item \texttt{NameError} (Figure~\ref{fig: nameerror});
        \item \texttt{IndexError} (Figure~\ref{fig: indexerror});
        \item \texttt{TypeError} (Figure~\ref{fig: typeerror});
        \item \texttt{ValueError} (Figure~\ref{fig: valueerror});
        \item \texttt{ImportError} (Figure~\ref{fig: importerror});
        \item \texttt{AttributeError} (Figure~\ref{fig: attribute error});
        \item \texttt{NotImplementedError} (Figure~\ref{fig: notimplementederror}).
    \end{itemize}
\end{itemize}

\begin{figure}[h!]
\centering
\begin{tcolorbox}[colback=gray!5!white, colframe=blue!75!black, 
title=Python Code, boxrule=0.3mm, width=0.9\textwidth, arc=3mm, auto outer arc=true, fontupper=\footnotesize]
``````python\\
number = 2024\\
binary\_str = bin(number)[2:]\\
print(f``Binary representation of {number}: {binary\_str}")\\
positions = [i for i, bit in enumerate(reversed(binary\_str)) if bit == `1']\\
print(f``Positions with 1s (from right, starting at 0): {positions}")\\
\textcolor{red}{"""}
\end{tcolorbox}
\centering
\begin{tcolorbox}[colback=gray!5!white, colframe=orange!75!black, 
title=Execution Result, boxrule=0.3mm, width=0.9\textwidth, arc=3mm, auto outer arc=true, fontupper=\footnotesize]
File ``main.py", line 7\\
    \textcolor{red}{\hspace*{20pt}"""}\\
    \hspace*{20pt}\^\\
SyntaxError: unterminated triple-quoted string literal (detected at line 7)
\end{tcolorbox}
\caption{The example of \texttt{SyntaxError} that can be solved with rules. We observe model-generated code often contains markdown code block format (\textit{i.e.}, ``````python ... """), which is invalid for the Python interpreter. To mitigate this kind of error, the rule-based error correction component automatically detects the format of code blocks and eliminates them before executing the code.}
\label{fig: syntaxerror can solved by rule}
\end{figure}
\vspace{-10pt}
\begin{figure}[h!]
\centering
\begin{tcolorbox}[colback=gray!5!white, colframe=blue!75!black, 
title=Python Code, boxrule=0.3mm, width=0.9\textwidth, arc=3mm, auto outer arc=true, fontupper=\footnotesize]
import sympy as sp\\

y = sp.symbols(`y')\\
\hspace*{8pt}\textcolor{red}{equation = y**3 - 27*y + 46}\\
\hspace*{8pt}\textcolor{red}{roots = sp.solve(equation, y)}\\
\hspace*{8pt}\textcolor{red}{print(roots)}
\end{tcolorbox}
\begin{tcolorbox}[colback=gray!5!white, colframe=orange!75!black, 
title=Execution Result, boxrule=0.3mm, width=0.9\textwidth, arc=3mm, auto outer arc=true, fontupper=\footnotesize]
File ``main.py", line 4\\
    \hspace*{20pt}\textcolor{red}{equation = y**3 - 27*y + 46}\\
    \hspace*{20pt}\^\\
    IndentationError: unexpected indent
\end{tcolorbox}
\caption{The example of \texttt{IndentationError} in model-generated code scripts. The rule-based error correction component automatically identifies the unexpected indentation in the code script and fixes it.}
\label{fig: indentationerror}
\end{figure}

\begin{figure}[h!]
\centering
\begin{tcolorbox}[colback=gray!5!white, colframe=blue!75!black, 
title=Python Code, boxrule=0.3mm, width=0.9\textwidth, arc=3mm, auto outer arc=true, fontupper=\footnotesize]
from itertools import combinations\_with\_replacement \par
\par
\# Calculate the number of non-negative integer solutions to a + b + c = 8 \par
\par
\# Using stars and bars formula: C(n + k -1, k-1) where n=8, k=3 \par
n = 8 \par
k = 3 \par
\textcolor{red}{solutions = (n + k -1) choose (k-1)} \par
\par
\# But let's compute it with code as well \par
count = 0 \par
for a in range(0, 9): \par
\hspace*{1em}for b in range(0, 9 - a): \par
\hspace*{2em}c = 8 - a - b \par
\hspace*{2em}if c \textgreater= 0 and c \textless=9: \par
\hspace*{3em}count +=1 \par
print(count) \par
\end{tcolorbox}
\begin{tcolorbox}[colback=gray!5!white, colframe=orange!75!black, 
title=Execution Result, boxrule=0.3mm, width=0.9\textwidth, arc=3mm, auto outer arc=true, fontupper=\footnotesize]
File ``main.py", line 6 \par
\hspace*{2em}\textcolor{red}{solutions = (n + k -1) choose (k-1)} \par
\hspace*{12em}\^{}\^{}\^{}\^{}\^{}\^{} \par
SyntaxError: invalid syntax \par
\end{tcolorbox}
\caption{The example of \texttt{SyntaxError} that cannot be solved with rules. The model-based error correction components provide specific feedback to the model via the tool response that instructs the model to identify and solve this error.}
\label{fig: syntaxerror cannot solved by rule}
\end{figure}

\begin{figure}[h!]
\centering

\begin{tcolorbox}[colback=gray!5!white, colframe=blue!75!black, 
title=Python Code, boxrule=0.3mm, width=0.9\textwidth, arc=3mm, auto outer arc=true, fontupper=\footnotesize]
from fractions import Fraction\\
\\
\# calculate combinations\\
\textcolor{red}{c0\_4 = math.comb(4, 4)}\\
\textcolor{red}{c1\_4 = math.comb(4, 3)}\\
\textcolor{red}{c2\_4 = math.comb(4, 2)}\\
\textcolor{red}{c3\_4 = math.comb(4, 1)}\\

\# Compute probabilities using exact fractions\\
P\_A = Fraction(c1\_4, c0\_4 + c1\_4 + c2\_4 + c3\_4)\\
numerator\_B = c0\_4 + c1\_4 * c6\_3\\
P\_B = Fraction((numerator\_B), c0\_4)\\
\\
conditional\_prob = P\_A / P\_B
\end{tcolorbox}

\begin{tcolorbox}[colback=gray!5!white, colframe=orange!75!black, 
title=Execution Result, boxrule=0.3mm, width=0.9\textwidth, arc=3mm, auto outer arc=true, fontupper=\footnotesize]
Traceback (most recent call last):\\
  \hspace*{12pt} File ``stdin'', line 4, in \\
    \hspace*{20pt} \textcolor{red}{c0\_4 = math.comb(4, 4)}\\
    \hspace*{54pt} \^ \ \^ \ \^ \ \^  \\
NameError: name `math' is not defined
\end{tcolorbox}
\caption{The example of \texttt{NameError} in model-generated code scripts. The model-based error correction component provides the likely root reason of this error (\textit{i.e.,} the model mistakenly uses undefined variables or Python packages) to help the model identify and fix this error.}
\label{fig: nameerror}
\end{figure}

\begin{figure}[h!]
\centering
\begin{tcolorbox}[colback=gray!5!white, colframe=blue!75!black, 
title=Python Code, boxrule=0.3mm, width=0.9\textwidth, arc=3mm, auto outer arc=true, fontupper=\footnotesize]
from sympy import symbols, solve, sqrt \par
\par
\# Define the variables \par
x, y, z = symbols(`x y z') \par
\par
\# Solve for vertices under the condition x \textless y \textless z \par
\# First vertex: x = y \par
x1 = 25 \par
y1 = 25 \par
z1 = 75 - 2*x1 \par
\par
\# Second vertex: y = z \par
y2 = 25 \par
x2 = 75 - 2*y2 \par
z2 = y2 \par
\par
\# Third vertex: x = z \par
z3 = 25 \par
x3 = z3 \par
y3 = 75 - 2*z3 \par
\par
\# Ensure x \textless y \textless z \par
vertices = [(x1, y1, z1), (x2, y2, z2), (x3, y3, z3)] \par
vertices = [(x, y, z) for x, y, z in vertices if x \textless y \textless z] \par
\par
\# Calculate the area of the triangle \par
\textcolor{red}{area = sqrt((vertices[0][0] - vertices[1][0])**2 + (vertices[0][1] - vertices[1][1])**2 \textbackslash} \par \hspace*{2em} \textcolor{red}{+ (vertices[0][2] - vertices[1][2])**2) * sqrt((vertices[0][0] - vertices[2][0])**2 \textbackslash} \par \hspace*{2em} \textcolor{red}{+ (vertices[0][1] - vertices[2][1])**2 + (vertices[0][2] - vertices[2][2])**2) \textbackslash/ 2} \par
\par
\# Simplify the area \par
area\_simplified = area.simplify() \par
area\_simplified \par
\end{tcolorbox}

\begin{tcolorbox}[colback=gray!5!white, colframe=orange!75!black, 
title=Execution Result, boxrule=0.3mm, width=0.9\textwidth, arc=3mm, auto outer arc=true, fontupper=\footnotesize]
Traceback (most recent call last): \par
\hspace*{2em}File ``main.py'', line 21, in \textless module\textgreater \par
\hspace*{4em}\textcolor{red}{area = sqrt((vertices[0][0] - vertices[1][0])**2 + (vertices[0][1] - vertices[1][1])**2 + (vertices[0][2] - vertices[1][2])**2) * sqrt((vertices[0][0] - vertices[2][0])**2 + (vertices[0][1] - vertices[2][1])**2 + (vertices[0][2] - vertices[2][2])**2) \textbackslash/ 2} \par
IndexError: list index out of range \par
\end{tcolorbox}

\caption{The example of \texttt{IndexError} in model-generated code scripts.}
\label{fig: indexerror}
\end{figure}

\begin{figure}[h!]
\centering
\begin{tcolorbox}[colback=gray!5!white, colframe=blue!75!black, 
title=Python Code, boxrule=0.3mm, width=0.9\textwidth, arc=3mm, auto outer arc=true, fontupper=\footnotesize]

import numpy as np \par
from scipy.optimize import root\_scalar \par
\par
\# Define the functions \par
def f(x): \par
\hspace*{1em}return abs(abs(x) - 1/2) \par
def g(x): \par
\hspace*{1em}return abs(abs(x) - 1/4) \par
def y\_func(x): \par
\hspace*{1em}return 4 * g(f(np.sin(2 * np.pi * x))) \par
def x\_func(y): \par
\hspace*{1em}return 4 * g(f(np.cos(3 * np.pi * y))) \par
def equation(x): \par
\hspace*{1em}y = y\_func(x) \par
\hspace*{1em}x\_prime = x\_func(y) \par
\hspace*{1em}return x\_prime - x \par
\par
\# Use root\_scalar to find roots in [0, 1] \par
\par
\# Initial guess \par
initial\_guess = 0.5 \par
\par
\# Solve using bisection method \par
\textcolor{red}{result = root\_scalar(equation, bracket=[0, 1], method=`bisection', tol=1e-6)} \par
\par
\# Print the root \par
print(f``Root found at x = \{result.root\:.6f\}") \par
\par
\# Check for other roots by evaluating the function at different points \par
x\_values = np.linspace(0, 1, 1000) \par
function\_values = [equation(x) for x in x\_values] \par
\par
\# Count the number of sign changes \par
sign\_changes = 0 \par
for i in range(len(function\_values) - 1): \par
\hspace*{1em}if function\_values[i] * function\_values[i+1] $<$ 0: \par
\hspace*{2em}sign\_changes += 1 \par
\par
print(f``Number of sign changes (potential roots): \{sign\_changes\}")
\end{tcolorbox}

\begin{tcolorbox}[colback=gray!5!white, colframe=orange!75!black, 
title=Execution Result, boxrule=0.3mm, width=0.9\textwidth, arc=3mm, auto outer arc=true, fontupper=\footnotesize]
Traceback (most recent call last):
  \hspace*{1em}File ``main.py", line 20, in \textless module\textgreater \par
    \hspace*{2em}\textcolor{red}{result = root\_scalar(equation, bracket=[0, 1], method=`bisection', tol=1e-6)} \par
            \hspace*{5.6em}\^{}\^{}\^{}\^{}\^{}\^{}\^{}\^{}\^{}\^{}\^{}\^{}\^{}\^{}\^{}\^{}\^{}\^{}\^{}\^{}\^{}\^{}\^{}\^{}\^{}\^{}\^{}\^{}\^{}\^{}\^{}\^{}\^{}\^{}\^{}\^{}\^{}\^{}\^{}\^{}\^{}\^{}\^{}\^{}\^{}\^{}\^{}\^{}\^{}\^{}\^{}\^{}\^{}\^{}\^{}\^{}\^{}\^{}\^{}\^{}\^{}\^{}\^{}\^{}\^{}\^{}\^{}\^{}\^{}\^{}\^{}\^{}\^{}\^{}\^{}\^{}\^{}\^{}\^{}\^{}\^{}\^{}\^{}\^{}\^{}\^{}\^{} \par
TypeError: root\_scalar() got an unexpected keyword argument `tol'
\end{tcolorbox}
\caption{The example of \texttt{TypeError} in model-generated code scripts. The reasoning model is instructed by the model-based error correction component to check the correctness of the type of variables.}
\label{fig: typeerror}
\end{figure}

\begin{figure}[h!]
    
\centering
\begin{tcolorbox}[colback=gray!5!white, colframe=blue!75!black, 
title=Python Code, boxrule=0.3mm, width=0.9\textwidth, arc=3mm, auto outer arc=true, fontupper=\footnotesize]
import numpy as np \par
\par
\# Define h(x) \par
def h(x): \par
\hspace*{1em}z = abs(np.sin(2 * np.pi * x)) \par
\hspace*{1em}\textcolor{red}{if z \textless= 0.5:} \par
\hspace*{2em}f\_z = 0.5 - z \par
\hspace*{1em}else: \par
\hspace*{2em}f\_z = z - 0.5 \par
\hspace*{1em}if f\_z \textless= 0.25: \par
\hspace*{2em}g\_f\_z = 0.25 - f\_z \par
\hspace*{1em}else: \par
\hspace*{2em}g\_f\_z = f\_z - 0.25 \par
\hspace*{1em}return 4 * g\_f\_z \par
\par
\# Define k(y) \par
def k(y): \par
\hspace*{1em}w = abs(np.cos(3 * np.pi * y)) \par
\hspace*{1em}if w \textless= 0.5: \par
\hspace*{2em}f\_w = 0.5 - w \par
\hspace*{1em}else: \par
\hspace*{2em}f\_w = w - 0.5 \par
\hspace*{1em}if f\_w \textless= 0.25: \par
\hspace*{2em}g\_f\_w = 0.25 - f\_w \par
\hspace*{1em}else: \par
\hspace*{2em}g\_f\_w = f\_w - 0.25 \par
\hspace*{1em}return 4 * g\_f\_w \par
\par
\# Define F(x) = x - k(h(x)) \par
def F(x): \par
\hspace*{1em}\textcolor{red}{return x - k(h(x))} \par
\par
\# Evaluate F(x) over a grid of x values \par
x\_values = np.linspace(0, 1, 100000) \par
\textcolor{red}{F\_values = F(x\_values)} \par
\par
\# Count the number of sign changes in F\_values \par
sign\_changes = 0 \par
for i in range(len(F\_values) - 1): \par
\hspace*{1em}if F\_values[i] * F\_values[i+1] \textless 0: \par
\hspace*{2em}sign\_changes += 1 \par
\par
\# Print the number of sign changes \par
print(f`Number of sign changes: \{sign\_changes\}') \par
\end{tcolorbox}

\begin{tcolorbox}[colback=gray!5!white, colframe=orange!75!black, 
title=Execution Result, boxrule=0.3mm, width=0.9\textwidth, arc=3mm, auto outer arc=true, fontupper=\footnotesize]
Traceback (most recent call last): \par
\hspace*{2em}File ``main.py'', line 35, in \textless module\textgreater \par
\hspace*{4em}\textcolor{red}{F\_values = F(x\_values)} \par
\hspace*{4em}\hspace*{5em}\^{}\^{}\^{}\^{}\^{}\^{}\^{}\^{}\^{}\^{}\^{}\^{}\^{}\^{} \par
\hspace*{2em}File ``main.py'', line 31, in F \par
\hspace*{4em}\textcolor{red}{return x - k(h(x))} \par
\hspace*{4em}\hspace*{5.6em}\^{}\^{}\^{}\^{} \par
\hspace*{2em}File ``main.py'', line 6, in h \par
\hspace*{4em}\textcolor{red}{if z \textless= 0.5}: \par
\hspace*{4em}\hspace*{1em}\^{}\^{}\^{}\^{}\^{}\^{}\^{}\^{}\^{}\^{}\^{} \par
ValueError: The truth value of an array with more than one element is ambiguous. Use a.any() or a.all() \par
\end{tcolorbox}
\caption{The example of \texttt{ValueError} in model-generated code scripts. The model-based error correction component instructs the reasoning model to verify the validity of variable values.}
\label{fig: valueerror}
\end{figure}

\begin{figure}[h!]
\centering
\begin{tcolorbox}[colback=gray!5!white, colframe=blue!75!black, 
title=Python Code, boxrule=0.3mm, width=0.9\textwidth, arc=3mm, auto outer arc=true, fontupper=\footnotesize]
\textcolor{red}{from sympy import symbols, Eq, solve, maximize} \par
\par
x, y, z = symbols(`x y z') \par
\par
\# Constraints \par
constraint1 = Eq(x*y + y*z + z*x, 27) \par
constraint2 = Eq(x*y*z, 23) \par
\par
\# Function to maximize \par
objective = x**2 + y**2 + z**2 \par
\par
\# Use Lagrange multipliers \par
lambda1, lambda2 = symbols(`lambda1 lambda2') \par
L = objective - lambda1*(constraint1.lhs - constraint1.rhs) \textbackslash{} \par \hspace*{2em}- lambda2*(constraint2.lhs - constraint2.rhs) \par
\par
\# Partial derivatives \par
partial\_x = L.diff(x) \par
partial\_y = L.diff(y) \par
partial\_z = L.diff(z) \par
partial\_lambda1 = L.diff(lambda1) \par
partial\_lambda2 = L.diff(lambda2) \par
\par
\# Solve the system of equations \par
solutions = solve([partial\_x, partial\_y, partial\_z, partial\_lambda1, partial\_lambda2], (x, y, z, lambda1, lambda2)) \par
solutions \par
\end{tcolorbox}

\begin{tcolorbox}[colback=gray!5!white, colframe=orange!75!black, 
title=Execution Result, boxrule=0.3mm, width=0.9\textwidth, arc=3mm, auto outer arc=true, fontupper=\footnotesize]
Traceback (most recent call last): \par
\hspace*{2em}File ``main.py'', line 1, in \textless module\textgreater \par
\hspace*{4em}\textcolor{red}{from sympy import symbols, Eq, solve, maximize} \par
ImportError: cannot import name `maximize' from `sympy' \par (/usr/local/lib/python3.11/dist-packages/sympy/\_\_init\_\_.py) \par
\end{tcolorbox}
\caption{The example of \texttt{ImportError} in model-generated code scripts. The model-based error correction component provides instructions for the reasoning model to verify the imported packages in the code script.}
\label{fig: importerror}
\end{figure}

\begin{figure}[h!]
\centering
\begin{tcolorbox}[colback=gray!5!white, colframe=blue!75!black, 
title=Python Code, boxrule=0.3mm, width=0.9\textwidth, arc=3mm, auto outer arc=true, fontupper=\footnotesize]
import numpy as np \par
from scipy.optimize import root\_scalar \par
\par
\# Define the functions \par
\par
def f(x): \par
\hspace*{1em}return abs(abs(x) - 1/2) \par
\par
def g(x): \par
\hspace*{1em}return abs(abs(x) - 1/4) \par
\par
def y\_func(x): \par
\hspace*{1em}return 4 * g(f(np.sin(2 * np.pi * x))) \par
\par
def x\_func(y): \par
\hspace*{1em}return 4 * g(f(np.cos(3 * np.pi * y))) \par
\par
def equation(x): \par
\hspace*{1em}y = y\_func(x) \par
\hspace*{1em}x\_prime = x\_func(y) \par
\hspace*{1em}return x\_prime - x \par
\par
\# Use root\_scalar to find roots in [0, 1] \par
\par
\# Initial guess \par
initial\_guess = 0.5 \par
\par
\# Solve using bisection method \par
\textcolor{red}{result = root\_scalar(equation, bracket=[0, 1], method=`bisection', rtol=1e-6)}\par
\par
\# Print the root \par
print(f``Root found at x = \{result.root\:.6f\}") \par
\par
\# Check for other roots by evaluating the function at different points \par
x\_values = np.linspace(0, 1, 1000) \par
function\_values = [equation(x) for x in x\_values] \par
\par
\# Count the number of sign changes \par
sign\_changes = 0 \par
for i in range(len(function\_values) - 1): \par
\hspace*{1em}if function\_values[i] * function\_values[i+1] \textless 0: \par
\hspace*{2em}sign\_changes += 1 \par
\par
print(f``Number of sign changes (potential roots): \{sign\_changes\}") \par
\end{tcolorbox}

\begin{tcolorbox}[colback=gray!5!white, colframe=orange!75!black, 
title=Execution Result, boxrule=0.3mm, width=0.9\textwidth, arc=3mm, auto outer arc=true, fontupper=\footnotesize]
Traceback (most recent call last): \par
\hspace*{2em}File ``/usr/local/lib/python3.11/dist-packages/scipy/optimize/\_root\_scalar.py'', line 276, in root\_scalar \par
\hspace*{4em}methodc = getattr(optzeros, map2underlying.get(meth, meth)) \par
\hspace*{4em}\hspace*{5em}\^{}\^{}\^{}\^{}\^{}\^{}\^{}\^{}\^{}\^{}\^{}\^{}\^{}\^{}\^{}\^{}\^{}\^{}\^{}\^{}\^{}\^{}\^{}\^{}\^{}\^{}\^{}\^{}\^{}\^{}\^{}\^{}\^{}\^{}\^{}\^{}\^{}\^{}\^{}\^{}\^{}\^{}\^{}\^{}\^{}\^{}\^{}\^{}\^{}\^{}\^{}\^{}\^{}\^{}\^{}\^{}\^{}\^{}\^{}\^{}\^{}\^{}\^{}\^{}\^{}\^{}\^{}  \par
AttributeError: module `scipy.optimize.\_zeros\_py' has no attribute `bisection'. Did you mean: `bisect'? \par
\par
The above exception was the direct cause of the following exception: \par
\par
Traceback (most recent call last): \par
\hspace*{2em}File ``main.py'', line 20, in \textless module\textgreater \par
\hspace*{4em}\textcolor{red}{result = root\_scalar(equation, bracket=[0, 1], method=`bisection', rtol=1e-6)} \par
\hspace*{4em}\hspace*{3.6em}\^{}\^{}\^{}\^{}\^{}\^{}\^{}\^{}\^{}\^{}\^{}\^{}\^{}\^{}\^{}\^{}\^{}\^{}\^{}\^{}\^{}\^{}\^{}\^{}\^{}\^{}\^{}\^{}\^{}\^{}\^{}\^{}\^{}\^{}\^{}\^{}\^{}\^{}\^{}\^{}\^{}\^{}\^{}\^{}\^{}\^{}\^{}\^{}\^{}\^{}\^{}\^{}\^{}\^{}\^{}\^{}\^{}\^{}\^{}\^{}\^{}\^{}\^{}\^{}\^{}\^{}\^{}\^{}\^{}\^{}\^{}\^{}\^{}\^{}\^{}\^{}\^{}\^{}\^{}\^{}\^{}\^{}\^{}\^{}\^{}\^{}\^{} \par
\hspace*{2em}File ``/usr/local/lib/python3.11/dist-packages/scipy/optimize/\_root\_scalar.py'', line 278, in root\_scalar \par
\hspace*{4em}raise ValueError(f`Unknown solver \{meth\}' ) from e \par
ValueError: Unknown solver bisection \par
\end{tcolorbox}
\caption{The example of \texttt{AttributeError} in model-generated code scripts. The model-based error correction component gives feedback to the reasoning model to verify the called attribute of using variables.}
\label{fig: attribute error}
\end{figure}

\begin{figure}[h!]
\centering
\begin{tcolorbox}[colback=gray!5!white, colframe=blue!75!black,
title=Python Code, boxrule=0.3mm, width=0.9\textwidth, arc=3mm, auto outer arc=true, fontupper=\footnotesize]
import sympy as sp\\
\\
x, y = sp.symbols(`x y')\\
\\
\# Equation of AB\\
ab\_eq = sp.Eq(y, -sp.sqrt(3)\*x + sp.sqrt(3)/2)\\
\\
\# Equation of the envelope\\
envelope\_eq = sp.Eq(x\*\*\,(2/3) + y\*\*\,(2/3), 1)\\
\\
\# Substitute y from ab\_eq into envelope\_eq\\
substituted\_eq = envelope\_eq.subs(y, ab\_eq.rhs)\\
\\
\# Solve the equation\\
\textcolor{red}{solution = sp.solve(substituted\_eq, x)}\\
\\
solution\\
\end{tcolorbox}

\clearpage
\begin{tcolorbox}[colback=gray!5!white, colframe=orange!75!black,
title=Execution Result, boxrule=0.3mm, width=0.9\textwidth, arc=3mm, auto outer arc=true, fontupper=\footnotesize]
Traceback (most recent call last):\\
\hspace*{2em}File ``main.py", line 15, in \textless module\textgreater\\
\textcolor{red}{\hspace*{4em}solution = sp.solve(substituted\_eq, x)}\\
\hspace*{8.6em}\^{}\^{}\^{}\^{}\^{}\^{}\^{}\^{}\^{}\^{}\^{}\^{}\^{}\^{}\^{}\^{}\^{}\^{}\^{}\^{}\^{}\^{}\^{}\^{}\^{}\^{}\^{}\^{}\^{}\^{}\^{}\^{}\^{}\^{}\\
\hspace*{2em}File ``/usr/local/lib/python3.11/dist-packages/sympy/solvers/solvers.py", line 1170, in solve\\
\hspace*{4em}solution = \_solve(f[0], \*symbols, \*\*flags)\\
\hspace*{8.6em}\^{}\^{}\^{}\^{}\^{}\^{}\^{}\^{}\^{}\^{}\^{}\^{}\^{}\^{}\^{}\^{}\^{}\^{}\^{}\^{}\^{}\^{}\^{}\^{}\^{}\^{}\^{}\^{}\^{}\^{}\^{}\^{}\^{}\^{}\^{}\\
\hspace*{2em}File ``/usr/local/lib/python3.11/dist-packages/sympy/solvers/solvers.py", line 1729, in \_solve\\
\hspace*{4em}raise NotImplementedError(`\textbackslash n'.join([msg, not\_impl\_msg \# fi]))\\
NotImplementedError: multiple generators \{(-2\texttt{*}sqrt(3)\texttt{*}x + sqrt(3))\texttt{*}\texttt{*}\{(1/3)\}, x\texttt{*}\texttt{*}\{(1/3)\}\}
No algorithms are implemented to solve equation x\texttt{*}\texttt{*}\{2/3\} + (-sqrt(3)\texttt{*}x + sqrt(3)/2)\texttt{*}\texttt{*}\{2/3\} - 1
\end{tcolorbox}

\caption{The example of \texttt{NotImplementedError} in model-generated code scripts.}
\label{fig: notimplementederror}
\end{figure}

\subsection{Cognitive Behaviors}
\label{app: case cognitive behaviors}
In this section, we provide the complete model responses in Figures~\ref{fig: full output for decomposition}, \ref{fig: full output for correction}, \ref{fig: full output for verification}, and \ref{fig: full output for backtracking} for the case studies in Sec.~\ref{sec: exp}. 
\clearpage

\begin{tcolorbox}[colback=cyan!5,colframe=cyan!40!black,title=Case study: Decomposition, breakable, enhanced]
\small
\begin{tcolorbox}[colback=cyan!5,colframe=cyan!60!black!60,boxrule=0.3pt,left=2pt,right=2pt,top=1pt,bottom=1pt,title=Problem]
Jen enters a lottery by picking $4$ distinct numbers from $S=\{1,2,3,\cdots,9,10\}.$ $4$ numbers are randomly chosen from $S.$ She wins a prize if at least two of her numbers were $2$ of the randomly chosen numbers, and wins the grand prize if all four of her numbers were the randomly chosen numbers. The probability of her winning the grand prize given that she won a prize is $\tfrac{m}{n}$ where $m$ and $n$ are relatively prime positive integers. Find $m+n$.
\end{tcolorbox}

To solve this problem, we need to break it down into several steps:

1. \textbf{Determine the total number of ways to pick 4 numbers out of 10}:
   This is a combination problem where we choose 4 numbers from a set of 10. The number of ways to do this is denoted as \( C(10, 4) \).
2. \textbf{Determine the number of favorable outcomes for winning a prize}:
   - \textbf{Winning a prize}: Jen's 4 chosen numbers must have at least 2 numbers in common with the 4 randomly chosen numbers.
   - \textbf{Winning the grand prize}: Jen's 4 chosen numbers must be exactly the same as the 4 randomly chosen numbers.

3. \textbf{Calculate the conditional probability}:
   - We need to find the probability of winning the grand prize given that she has won a prize.

Let's start by calculating the total number of ways to pick 4 numbers out of 10.

\begin{tcolorbox}[colback=cyan!5,colframe=cyan!60!black!60,boxrule=0.3pt,left=2pt,right=2pt,top=1pt,bottom=1pt,title=Python Interpreter Calling, breakable, enhanced]
from math import comb \\
\# Total number of ways to pick 4 numbers out of 10 \\
C\_10\_4 = comb(10, 4) \\
C\_10\_4
\end{tcolorbox}
The result of the code execution is: 210

Next, let's calculate the number of favorable outcomes for winning a prize. This involves counting the number of ways Jen's 4 chosen numbers can have at least 2 numbers in common with the 4 randomly chosen numbers.We will break this down further:\\
- \textbf{Exactly 2 numbers in common}: We choose 2 numbers from the 4 picked by Jen and 2 numbers from the remaining 6 numbers.\\
- \textbf{Exactly 3 numbers in common}: We choose 3 numbers from the 4 picked by Jen and 1 number from the remaining 6 numbers.\\
- \textbf{Exactly 4 numbers in common}: We choose all 4 numbers from the 4 picked by Jen.\\
Let's compute these values.

\begin{tcolorbox}[colback=cyan!5,colframe=cyan!60!black!60,boxrule=0.3pt,left=2pt,right=2pt,top=1pt,bottom=1pt,title=Python Interpreter Calling, breakable, enhanced]
from math import comb \\
\# Number of ways to have exactly 2 numbers in common\\
C\_4\_2 = comb(4, 2)\\
C\_6\_2 = comb(6, 2)\\
ways\_2\_common = C\_4\_2 * C\_6\_2\\

\# Number of ways to have exactly 3 numbers in common\\
C\_4\_3 = comb(4, 3)\\
C\_6\_1 = comb(6, 1)\\
ways\_3\_common = C\_4\_3 * C\_6\_1\\

\# Number of ways to have exactly 4 numbers in common\\
C\_4\_4 = comb(4, 4)\\
C\_6\_0 = comb(6, 0)\\
ways\_4\_common = C\_4\_4 * C\_6\_0\\

\# Total number of favorable outcomes for winning a prize\\
favorable\_winning\_prize = ways\_2\_common + ways\_3\_common + ways\_4\_common\\

favorable\_winning\_prize, ways\_2\_common, ways\_3\_common, ways\_4\_common\\

\end{tcolorbox}

The result of the code execution is: 115 90 24 1 \\

The calculations yield the following results:
- Number of ways to have exactly 2 numbers in common: \( 90 \)
- Number of ways to have exactly 3 numbers in common: \( 24 \)
- Number of ways to have exactly 4 numbers in common: \( 1 \)

Thus, the total number of favorable outcomes for winning a prize is:
\[ 90 + 24 + 1 = 115 \]

Now, we need to find the probability of winning the grand prize given that she has won a prize. This is the ratio of the number of ways to win the grand prize to the total number of ways to win a prize:
\[ P(\text{Grand Prize} \mid \text{Prize}) = \frac{\text{Number of ways to win the grand prize}}{\text{Total number of favorable outcomes for winning a prize}} = \frac{1}{115} \]

Since 1 and 115 are relatively prime, the fraction is already in its simplest form. Therefore, \( m = 1 \) and \( n = 115 \), and the sum \( m + n \) is:
\[ 1 + 115 = 116 \]

Thus, the final answer is: $\boxed{116}$.

\end{tcolorbox}
\captionsetup{type=figure}
\captionof{figure}{The complete model response for the Decomposition case study.}
\label{fig: full output for decomposition}

\newpage
\begin{tcolorbox}[colback=OliveGreen!5,colframe=OliveGreen!70!Black!70,title=Case study: Correction, breakable, enhanced]
\small
\begin{tcolorbox}[colback=OliveGreen!5,colframe=OliveGreen!50!Black!40,boxrule=0.3pt,left=2pt,right=2pt,top=1pt,bottom=1pt,title=Problem]
Let $N$ be the greatest four-digit positive integer with the property that whenever one of its digits is changed to $1$, the resulting number is divisible by $7$. Let $Q$ and $R$ be the quotient and remainder, respectively, when $N$ is divided by $1000$. Find $Q+R$.
\end{tcolorbox}
To find the greatest four-digit positive integer \( N \) with the property that changing any one of its digits to \( 1 \) results in a number divisible by \( 7 \), we need to follow these steps:

1. \textbf{Understand the Problem:}\\
   - We need \( N \) such that \( N \equiv 0 \pmod{7} \).\\
   - Changing any digit of \( N \) to \( 1 \) should also result in a number divisible by \( 7 \).

2. \textbf{Formulate the Constraints:}\\
   - Let \( N = 1000a + 100b + 10c + d \), where \( a, b, c, \) and \( d \) are the digits of \( N \).\\
   - For each digit \( x \in \{a, b, c, d\} \), the number formed by changing \( x \) to \( 1 \) must be divisible by \( 7 \).

3. \textbf{Check the Conditions:}\\
   - If \( a \) is changed to \( 1 \), the new number is \( 1000 + 100b + 10c + d \). This must be divisible by \( 7 \):
     \[
     1000 + 100b + 10c + d \equiv 0 \pmod{7}
     \]
   - Similarly, for \( b, c, \) and \( d \):
     \[
     1000a + 100 + 10c + d \equiv 0 \pmod{7}
     \]
     \[
     1000a + 100b + 10 + d \equiv 0 \pmod{7}
     \]
     \[
     1000a + 100b + 10c + 1 \equiv 0 \pmod{7}
     \]

4. \textbf{Simplify the Conditions:}\\
   - Since \( 1000 \equiv 6 \pmod{7} \), \( 100 \equiv 2 \pmod{7} \), and \( 10 \equiv 3 \pmod{7} \), we can rewrite the conditions as:
     \[
     6a + 2b + 3c + d + 1 \equiv 0 \pmod{7}
     \]
     \[
     6a + 2b + 3c + d + 2 \equiv 0 \pmod{7}
     \]
     \[
     6a + 2b + 3c + d + 3 \equiv 0 \pmod{7}
     \]
     \[
     6a + 2b + 3c + d + 6 \equiv 0 \pmod{7}
     \]

5. \textbf{Find the Greatest Four-Digit Number:}\\
   - We need to find the largest \( N \) such that all these conditions hold true. We will start from the largest four-digit number and check downwards.
   
Let's write a Python script to find the largest $N$ that satisfies these conditions.
\begin{tcolorbox}[colback=OliveGreen!5,colframe=OliveGreen!50!Black!40,boxrule=0.3pt,left=2pt,right=2pt,top=1pt,bottom=1pt,title=Python Interpreter Calling]
def find\_largest\_N():\\
\hspace*{2em}for N in range(9999, 999, -1):\\
\hspace*{4em}a, b, c, d = map(int, str(N))\\
\hspace*{4em}if (6*a + 2*b + 3*c + d + 1) \% 7 == 0 and \textbackslash\\
\hspace*{6em}(6*a + 2*b + 3*c + d + 2) \% 7 == 0 and \textbackslash\\
\hspace*{6em}(6*a + 2*b + 3*c + d + 3) \% 7 == 0 and \textbackslash\\
\hspace*{6em}(6*a + 2*b + 3*c + d + 6) \% 7 == 0:\\
\hspace*{6em}return N\\
\\
largest\_N = find\_largest\_N()\\
print(largest\_N)
\end{tcolorbox}

The result of the code execution is: None

It seems there was an issue with the initial approach. Let's re-evaluate the constraints and ensure the logic is correct.

We need to find the largest four-digit number \( N \) such that changing any one of its digits to \( 1 \) results in a number divisible by \( 7 \). Let's re-check the modular arithmetic conditions and ensure the constraints are correctly implemented.

We will re-run the script with a more detailed check to ensure the conditions are met.

\begin{tcolorbox}[colback=OliveGreen!5,colframe=OliveGreen!50!Black!40,boxrule=0.3pt,left=2pt,right=2pt,top=1pt,bottom=1pt,title=Python Interpreter Calling]
def find\_largest\_N():\\
\hspace*{2em}for N in range(9999, 999, -1):\\
\hspace*{4em}a, b, c, d = map(int, str(N))\\
\hspace*{4em}if (1000 + 100*b + 10*c + d) \% 7 == 0 and \textbackslash\\
\hspace*{6em}(1000*a + 100 + 10*c + d) \% 7 == 0 and \textbackslash\\
\hspace*{6em}(1000*a + 100*b + 10 + d) \% 7 == 0 and \textbackslash\\
\hspace*{6em}(1000*a + 100*b + 10*c + 1) \% 7 == 0:\\
\hspace*{6em}return N\\
\\
largest\_N = find\_largest\_N()\\
print(largest\_N)
\end{tcolorbox}

The result of the code execution is: 5694

The largest four-digit number \( N \) that meets the criteria is \( 5694 \).

Next, we need to find \( Q \) and \( R \) when \( N \) is divided by \( 1000 \):\\
- \( Q \) is the quotient, which is the integer part of \( \frac{5694}{1000} \).\\
- \( R \) is the remainder, which is \( 5694 \mod 1000 \).

Let's calculate \( Q \) and \( R \) and then find \( Q + R \).

\begin{tcolorbox}[colback=OliveGreen!5,colframe=OliveGreen!50!Black!40,boxrule=0.3pt,left=2pt,right=2pt,top=1pt,bottom=1pt,title=Python Interpreter Calling]
N = 5694\\
Q = N \/\/ 1000\\
R = N \% 1000\\
result = Q + R\\
print(result)]]
\end{tcolorbox}
The result of the code execution is: 699

The quotient \( Q \) when \( 5694 \) is divided by \( 1000 \) is \( 5 \), and the remainder \( R \) is \( 694 \). Therefore, \( Q + R = 5 + 694 = 699 \).

Thus, the final answer is: $\boxed{699}$.
\end{tcolorbox}
\captionsetup{type=figure}
\captionof{figure}{The complete model response for the Correction case study.}
\label{fig: full output for correction}

\newpage
\begin{tcolorbox}[colback=RoyalPurple!5,colframe=RoyalPurple!70,title=Case study: Verification, breakable, enhanced]
\small
\begin{tcolorbox}[colback=RoyalPurple!5,colframe=RoyalPurple!40,boxrule=0.3pt,left=2pt,right=2pt,top=1pt,bottom=1pt,title=Problem]
There exist real numbers $x$ and $y$, both greater than 1, such that $\log_x\left(y^x\right)=\log_y\left(x^{4y}\right)=10$. Find $xy$.
\end{tcolorbox}
To solve this problem, we will first make use of logarithmic properties to simplify the given equations. The given conditions are:

1. \(\log_x(y^x) = 10\)\\
2. \(\log_y(x^{4y}) = 10\)

We know from logarithm properties that \(\log_a(b^c) = c\log_a(b)\). Applying this property to both conditions, we have:

1. \(x\log_x(y) = 10\)\\
2. \(4y\log_y(x) = 10\)

Since \(\log_x(y) = \frac{\log(y)}{\log(x)}\) and \(\log_y(x) = \frac{\log(x)}{\log(y)}\), let's denote \(\log_x(y) = a\) and \(\log_y(x) = \frac{1}{a}\).

From the above, we can write:

1. \(xa = 10\)\\
2. \(4y\frac{1}{a} = 10\)

Solving these two equations will allow us to find \(x\) and \(y\), and subsequently, \(xy\). Let's start with the algebraic manipulation:

1. \(x = \frac{10}{a}\)\\
2. \(4y = 10a \Rightarrow y = \frac{10a}{4} = \frac{5a}{2}\)

Now, we need to compute \(xy\):

\[ xy = \left( \frac{10}{a} \right) \left( \frac{5a}{2} \right) = \frac{50a}{2a} = 25 \]

Let's verify the solution using the given conditions and ensure our steps are correct.

\begin{tcolorbox}[colback=RoyalPurple!5,colframe=RoyalPurple!40,boxrule=0.3pt,left=2pt,right=2pt,top=1pt,bottom=1pt,title=Python Interpreter Calling]
from sympy import symbols, Eq, solve, log \\
x, y, a = symbols('x y a') \\
eq1 = Eq(x*a, 10) \\
eq2 = Eq(4*y/a, 10) \\
sol = solve((eq1, eq2), (x, y)) \\
xy = sol[x] * sol[y] \\
xy.simplify()
\end{tcolorbox}

The result of the code execution is: 25

The value of $xy$ is $\boxed{25}$.

\end{tcolorbox}
\captionsetup{type=figure}
\captionof{figure}{The complete model response for the Verification case study.}
\label{fig: full output for verification}

\newpage
\begin{tcolorbox}[colback=Salmon!5,colframe=Salmon!90!Black,title=Case study: Backtracking, breakable, enhanced]
\small
\begin{tcolorbox}[colback=Salmon!5,colframe=Salmon!70!,boxrule=0.3pt,left=2pt,right=2pt,top=1pt,bottom=1pt,title=Problem]
Find the number of triples of nonnegative integers \((a,b,c)\) satisfying \(a + b + c = 300\) and
\begin{equation*}
a^2b + a^2c + b^2a + b^2c + c^2a + c^2b = 6,000,000.
\end{equation*}
\end{tcolorbox}
To solve this problem, we will follow these steps:

1. \textbf{Understanding the Equations}: We have two equations:\\
   - \(a + b + c = 300\)\\
   - \(a^2b + a^2c + b^2a + b^2c + c^2a + c^2b = 6,000,000\)

2. \textbf{Simplifying the Second Equation}: Notice that the second equation can be rewritten using symmetry:
   \[
   a^2(b+c) + b^2(a+c) + c^2(a+b) = 6,000,000
   \]

3. \textbf{Substitute \(b + c\), \(a + c\), and \(a + b\)}: Using \(a + b + c = 300\), we can express \(b + c\), \(a + c\), and \(a + b\) in terms of \(a\), \(b\), and \(c\):
   \[
   b + c = 300 - a, \quad a + c = 300 - b, \quad a + b = 300 - c
   \]

4. \textbf{Substitute into the Second Equation}:
   \[
   a^2(300 - a) + b^2(300 - b) + c^2(300 - c) = 6,000,000
   \]

5. \textbf{Expand and Simplify}:
   \[
   300a^2 - a^3 + 300b^2 - b^3 + 300c^2 - c^3 = 6,000,000
   \]
   \[
   300(a^2 + b^2 + c^2) - (a^3 + b^3 + c^3) = 6,000,000
   \]

6. \textbf{Express \(a^2 + b^2 + c^2\) in Terms of \(a + b + c\)}: Using the identity:
   \[
   (a + b + c)^2 = a^2 + b^2 + c^2 + 2(ab + bc + ca)
   \]
   \[
   300^2 = a^2 + b^2 + c^2 + 2(ab + bc + ca)
   \]
   \[
   90,000 = a^2 + b^2 + c^2 + 2(ab + bc + ca)
   \]

7. \textbf{Express \(ab + bc + ca\)}: From the first equation, we know:
   \[
   ab + bc + ca = \frac{(a + b + c)^2 - (a^2 + b^2 + c^2)}{2}
   \]
   Substituting \(a + b + c = 300\):
   \[
   ab + bc + ca = \frac{300^2 - (a^2 + b^2 + c^2)}{2}
   \]

8. \textbf{Combine Equations}: Substitute \(ab + bc + ca\) back into the simplified second equation:
   \[
   300(a^2 + b^2 + c^2) - (a^3 + b^3 + c^3) = 6,000,000
   \]
   \[
   300 \left( 90,000 - 2(ab + bc + ca) \right) - (a^3 + b^3 + c^3) = 6,000,000
   \]

9. \textbf{Solve for \(a, b, c\)}:
This step involves solving the resulting polynomial equation, which can be complex. However, we can use symmetry and the fact that \(a, b, c\) are nonnegative integers to find solutions.

Let's implement the above steps in Python to find the number of valid triples \((a, b, c)\).

\begin{tcolorbox}[colback=Salmon!5,colframe=Salmon!70!,boxrule=0.3pt,left=2pt,right=2pt,top=1pt,bottom=1pt,title=Python Interpreter Calling, breakable, enhanced]
from sympy import * \\
\# Define variables \\
a, b, c = symbols(`a b c', integer=True, nonnegative=True) \\
\# Define the equations \\
eq1 = Eq(a + b + c, 300) \\
eq2 = Eq(a**2 * b + a**2 * c + b**2 * a + b**2 * c + c**2 * a + c**2 * b, 6000000) \\

\# Solve the system of equations \\
solutions = solve([eq1, eq2], (a, b, c), dict=True) \\

\# Filter out only nonnegative integer solutions \\
valid\_solutions = [sol for sol in solutions if all(val.is\_integer and val >= 0 for val in sol.values())] \\

len(valid\_solutions)
\end{tcolorbox}
TypeError: cannot determine truth value of Relational

It appears that the direct symbolic solution approach using solve from SymPy did not work as expected due to the complexity of the equations. Let's try a different approach by iterating over possible values of a, b, and c to find all valid triples that satisfy both equations.

We will iterate over all possible values of \(a\) and \(b\) and compute \(c\) as \(300 - a - b\). Then, we will check if the computed \(c\) satisfies the second equation.

Let's implement this approach.
\begin{tcolorbox}[colback=Salmon!5,colframe=Salmon!70!,boxrule=0.3pt,left=2pt,right=2pt,top=1pt,bottom=1pt,title=Python Interpreter Calling]
count = 0\\
\# Iterate over possible values of a and b\\
for a in range(301):\\
\hspace*{2em}for b in range(301 - a):\\
\hspace*{4em}c = 300 - a - b\\
\hspace*{4em}if a**2 * b + a**2 * c + b**2 * a + b**2 * c + c**2 * a + c**2 * b == 6000000:\\
\hspace*{6em}count += 1\\
\\
print(count)
\end{tcolorbox}
The result of the code execution is: 601

Thus, the final answer is: $\boxed{601}$.
\end{tcolorbox}
\captionsetup{type=figure}
\captionof{figure}{The complete model response for the Backtracking case study.}
\label{fig: full output for backtracking}
\end{document}